\documentclass{article}

\usepackage{frExamplee}
\usepackage{apalike}
\usepackage{setspace}

\usepackage{xcolor}
\usepackage{graphicx} % Required for inserting images
\usepackage{bm}
\usepackage{amsmath}
\usepackage{times}
\usepackage{amsfonts}
\usepackage{amssymb}
\usepackage{amsthm}
\usepackage{xfrac}
\usepackage{balance}
\usepackage{pifont}
\usepackage{bm}
\usepackage{dsfont}
\usepackage{algorithm,algorithmic}
\usepackage{stfloats}
\usepackage{fancyhdr}

\usepackage[toc,page]{appendix} 
\usepackage[T1]{fontenc}
\usepackage{rotating}

\usepackage{acronym}[nolist,nohyperlinks]
\usepackage{subcaption}

\usepackage{hyperref}
\hypersetup{
    colorlinks=true,
    linkcolor=blue,
    citecolor=black,
    filecolor=magenta,      
    urlcolor=cyan,
    pdfpagemode=FullScreen,
    }

\usepackage[capitalize]{cleveref}
\crefname{section}{Section}{Sections}
\crefname{theorem}{Theorem}{Theorems}
\crefname{lemma}{Lemma}{Lemmas}
\crefname{table}{Table}{Tables}
\crefformat{equation}{(#2#1#3)}
\crefname{algocf}{Algorithm}{Algorithms}
\Crefname{algocf}{Algorithm}{Algorithms}
\crefname{ALC@unique}{Line}{Lines}

\newcommand{\ourbar}[1]{\overline{#1}}
\newtheorem{theorem}{Theorem}[section]

\fancypagestyle{firststyle}
{
   \fancyhf{}
   \fancyfoot[C]{\thepage}
   \fancyfoot[LE,LO]{\footnotesize{\quad Distribution Statement A. \\ \quad Approved for public release: distribution unlimited.}}
   
}

\title{Safe Autonomy for Uncrewed Surface Vehicles Using Adaptive Control and Reachability Analysis} % 

\author{Karan Mahesh\thanks{Indicates equal contribution. Authors are listed alphabetically.} \\
Aurora Flight Sciences, \\
a Boeing Company \\
Cambridge, MA, USA \\
{\tt \small mahesh.karan@aurora.aero}
\\\And
Tyler M. Paine$^{*}$\thanks{Corresponding author} \\
Marine Autonomy Laboratory \\
Massachusetts Institute of Tech.\\
Cambridge, MA, USA. \\
Woods Hole Oceanographic Institution \\
Woods Hole, MA, USA \\
{\tt \small tpaine@mit.edu}
\\\And
Max L. Greene \\
Aurora Flight Sciences, \\
a Boeing Company \\
Cambridge, MA, USA \\
{\tt \small greene.max@aurora.aero}
\\\AND
Nicholas Rober \\
Aerospace Controls Laboratory \\
Massachusetts Institute of Technology\\
Cambridge, MA, USA. \\
{\tt \small nrober@mit.edu}
\And
Steven Lee \\
Aurora Flight Sciences, \\
a Boeing Company \\
Cambridge, MA, USA \\
{\tt \small lee.steven@aurora.aero}
\\\And
Sildomar T. Monteiro \\
Aurora Flight Sciences, \\
a Boeing Company \\
Cambridge, MA, USA \\
{\tt \small monteiro.sildomar@aurora.aero}
\\\AND
Anuradha Annaswamy \\
Active-Adaptive Control Lab \\
Massachusetts Institute of Tech.\\
Cambridge, MA, USA. \\
{\tt \small  aanna@mit.edu}
\\\And
Michael R. Benjamin \\
Marine Autonomy Lab \\
Massachusetts Institute of Tech.\\
Cambridge, MA, USA. \\
{\tt \small mikerb@mit.edu}
\\\And
Jonathan P. How \\
Aerospace Controls Lab \\
Massachusetts Institute of Tech.\\
Cambridge, MA, USA. \\
{\tt \small jhow@mit.edu}
}

\begin{document}

%Define acronyms here
\acrodef{USV}{uncrewed surface vehicle}
\acrodef{MRAC}{model reference adaptive control}
\acrodef{LQR}{linear quadratic regulator}
\acrodef{PID}{proportional integral derivative}
\acrodef{PI}{proportional-integral}
\acrodef{RISE}{robust integral of the sign of the error}
\acrodef{NED}{north east down}
\acrodef{FRD}{front-right-down}
\acrodef{SO(2)}{special orthogonal group of order two}
\acrodef{UNREP}{underway replenishment}
\acrodef{MHE}{moving horizon estimator}
\acrodef{RSOA}{reachable set over-approximation}
\acrodef{CG}{computational graph}
\acrodef{RMSE}{root mean squared error}
\acrodef{RMSD}{root mean squared deviation}
\acrodef{RBF}{radial basis function}

\maketitle
\thispagestyle{firststyle}

\begin{abstract}

     Marine robots must maintain precise control and ensure safety during tasks like ocean monitoring, even when encountering unpredictable disturbances that affect performance. Designing algorithms for uncrewed surface vehicles (USVs) requires accounting for these disturbances to control the vehicle and ensure it avoids obstacles. While adaptive control has addressed USV control challenges, real-world applications are limited, and certifying USV safety amidst unexpected disturbances remains difficult.
     To tackle control issues, we employ a model reference adaptive controller (MRAC) to stabilize the USV along a desired trajectory. For safety certification, we developed a reachability module with a moving horizon estimator (MHE) to estimate disturbances affecting the USV. This estimate is propagated through a forward reachable set calculation, predicting future states and enabling real-time safety certification. We tested our safe autonomy pipeline on a Clearpath Heron USV in the Charles River, near MIT.
     Our experiments demonstrated that the USV's MRAC controller and reachability module could adapt to disturbances like thruster failures and drag forces. The MRAC controller outperformed a PID baseline, showing a 45\%-81\% reduction in RMSE position error. Additionally, the reachability module provided real-time safety certification, ensuring the USV's safety.
     We further validated our pipeline's effectiveness in underway replenishment and canal scenarios, simulating relevant marine tasks.

\end{abstract}

\section{Introduction}
Marine robotics are an essential component of how we observe remote ocean phenomena and defend national interests \cite{yuh2011applications,zereik2018challenges,pedrozo2023advent}. Due to the large spatiotemporal scale of ocean activity, there is a need for marine robots that can be deployed for long periods of time and cover increasingly large areas \cite{rynne2009unmanned}.  
In addition, these vehicles are periodically expected to precisely navigate waterways or maneuver in close proximity to another vehicle if, for example, refueling in the field is necessary. 
The difficulty of these types of high-risk operations presents a barrier to fielding fully autonomous \acp{USV}.  In the case of crewed vessels, the ship is usually carefully driven by a human operator when passing through a canal or close to another vessel, even though an autopilot capability is typically used in the open water.  
Furthermore, \acp{USV} that operate in marine environments, such as \acp{USV}, are often exposed to stochastic motion from wind, waves, and currents, and may be subject to actuator degradation due to biofouling or mechanical deterioration~\cite{haldeman2016lessening}.
The unknown nature of these disturbances creates challenges not only in how to effectively control a vehicle, but also in how to certify that the autonomous system satisfies safety constraints associated with the system and its environment.
The control and safety certification challenges must be addressed to ensure autonomous marine vehicles can successfully complete tasks that require precise tracking, possibly during long missions during which a wide range of possible disturbances may influence the vehicle over time.
Therefore, as we deploy autonomous vehicles on increasingly complex and long-range missions, it is critical to develop autonomy pipelines that consider the disturbances acting on the system to provide effective control and safety certification throughout all portions of a mission. 

Many approaches have been developed to control \acp{USV}, including various forms of robust~\cite{Fossen1994guidance} and adaptive~\cite{YoergerSlotine1991} control policies.
The uncertainty of the disturbances and the nonlinear nature of \ac{USV} dynamics make robust control difficult, especially when actuator dynamics are considered~\cite{YoergerSlotine1991,Fossen1993ActuatorDynamics,WhitcombYoerger1996ThrusterModel,McFarlandWhitcomb2014AMBC_Act_Dyn}.
In contrast, adaptive control can be used to quickly react to dynamic disturbances and model uncertainty by including additional terms parameterized by weights on generic function approximators such as Gaussian basis functions \cite{Shen2020ATTC_RBF} or neural networks \cite{Woo2018WAMV_LSTM}.
While adaptive control has been well studied in theory~\cite{Fossen1995AC_Comp,Lavretsky.Wise2012}, its application to \acp{USV} has been limited to only a small number of studies with vehicles in test tanks or pools~\cite{YoergerSlotine1991}, or to short field demonstrations that are limited in scope~\cite{Liu2021AC} and~\cite{Fischer2014RISE}.  In this work, we add to the literature on adaptive control for \acp{USV} by reporting field results from new and more complex missions that are more relevant to fielded applications. 

\begin{figure}
    \centering
    \includegraphics[width=0.7\columnwidth]{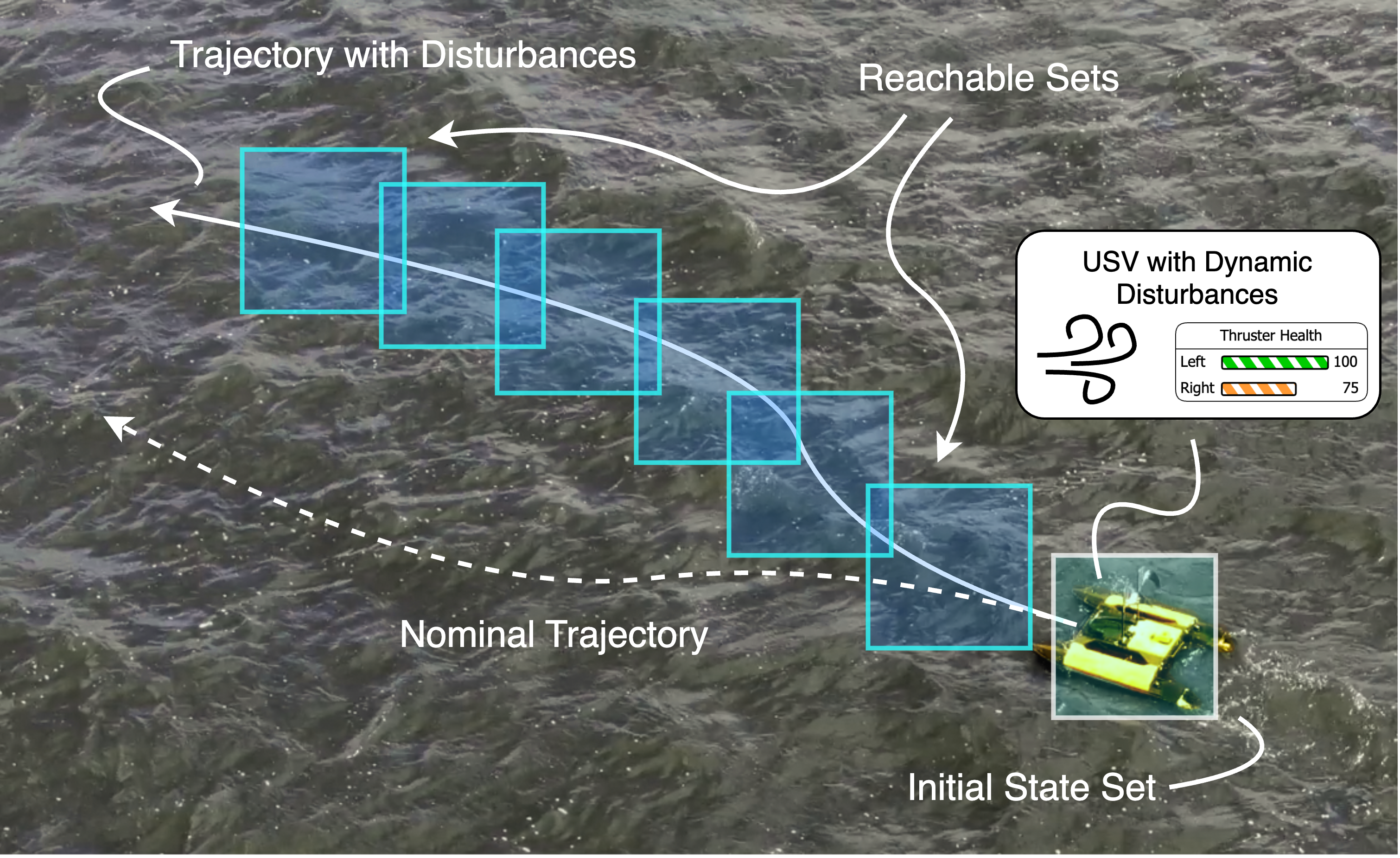}
    \caption{A diagram showing our approach to control and safety certification applied to a Clearpath Heron \ac{USV}.  The goal of the controller is to overcome disturbances and keep the trajectory with disturbances close to the nominal trajectory.
    The goal of reachability analysis is to generate reachable sets that bound the true (i.e., disturbed) trajectory.}
    \label{fig:intro_fig_heron}
\end{figure}

In addition to developing a controller, it is also important to ensure that the \ac{USV} remains safe in various scenarios. 
During its mission, a \ac{USV} may be tasked with navigating around obstacles~\cite{polvara2018obstacle} or performing complex maneuvers such as \ac{UNREP} ~\cite{miller1987development}. 
These situations are unavoidable since vessels often depart from 
harbors or pass through canals where they must navigate at least some of the way under their own power; the consequences of failure are enormous \cite{Lee2021SuezInpact,CRS2024BaltimoreBridge}. 
The task of \ac{UNREP} is particularly difficult since it requires precise coordination between vehicles, yet may be executed on missions that require sustained long deployments away from safe harbors. 
In such scenarios, it is critical to the safety of the system that collisions with other objects in the water are avoided, thus necessitating some method of certifying that the \ac{USV} is safe in the sense that there is a low probability that the vehicle moves outside of the prescribed safe region within a given time horizon. 

Given the uncertainty from the unknown nature of disturbances on the \ac{USV}, as well as uncertainty due to imperfect state estimation, this is a challenging problem. To this end, we formulate the general problem of safety as a \emph{reachability analysis} problem, and do so in a new way that is computationally tractable while capturing the uncertainty and complexity required for field deployments of \acp{USV}.

Reachability analysis~\cite{bansal2017hamilton,everett2021efficient,sidrane2022overt} provides safety certification given a known model and an imperfect state estimate by constructing \textit{reachable sets} that bound the possible future states of the system over some time horizon, as shown in~\Cref{fig:intro_fig_heron}.
By checking that the reachable sets do not overlap with an unsafe region of the state space, such as an area occupied by a ship or obstacle, reachability analysis can be used to certify safety.
While existing reachability approaches are capable of handling disturbances, they assume knowledge of a set of possible disturbances.
To account for all possible disturbances that could affect a system carrying out a long-range mission, this disturbance set would be very large, leading to overly conservative reachable sets that assume worst-case conditions at all times.
Thus, we develop a reachability analysis approach that uses real-time information to provide reasonable estimates of the reachable sets for safety certification of a \ac{USV} given the current understanding of disturbances acting on the system.

This work considers the challenges posed by control design and safety certification for \acp{USV}.
To address the challenge of controlling \acp{USV} in the presence of disturbances, this paper describes a \ac{MRAC} formulation similar to \cite{Lavretsky.Wise2012} wherein an adaptive term in the inner loop controller helps the system achieve the desired tracking response.
To address the problem of safety verification in the presence of disturbances, we use a formulation introduced in \cite{rober2023online} where a \ac{MHE} estimates the disturbances acting on the \ac{USV}, which can be incorporated into a real-time reachability calculation to predict the future motion of the vehicle.  
This paper extends \cite{rober2023online} by detailing the adaptive control law and showcasing extensive hardware experiments with a wide variety of disturbances that demonstrate how both the controller and reachabilty module are capable of adapting to changing disturbances that significantly affect the \ac{USV}'s dynamics. 
The presented methods are applicable to \acp{USV} in general, but are particularly helpful for \acp{USV} subject to widely varying disturbances or model uncertainty, such as those tasked for long-range missions where precise tracking and safety assurances are sometimes required.
With this in mind, we also demonstrate our algorithms in two scenarios relevant to long-range missions, including canal passage and \ac{UNREP}. 
The main contributions of our work can be summarized as follows:
\begin{itemize}
    \item We demonstrate the application of an \ac{MRAC} formulation, specifically \ac{MRAC} augmentation of optimal baseline controller, to a \ac{USV} operating in an environment with dynamic and unknown disturbances.
    Our controller shows improvement over a well-established PID control formulation, demonstrating better performance in the presence of real-world disturbances such as thruster faults, wind, currents, and drag forces.
    \item We develop a new reachability module that employs forward reachability analysis with real-time disturbance estimation to provide up-to-date certificates of the system's safety with respect to specified constraints.  Our method can display to a supervisor the effect of insurmountable disturbances on the \ac{USV}'s performance. 
\end{itemize}

We leverage a user-designed reference model in both our \ac{MRAC} and real-time reachability approach. In the \ac{MRAC} formulation, the objective of the adaptive controller is to drive the error between the reference model and the actual vehicle plant asymptotically to zero.  At the same time, the \ac{MHE} estimates the bias between the reference model and the plant using recently observed data. The estimated bias and bias uncertainty are then used in the forward reachability analysis.  Thus, the forward reachable sets are conservative in part because it is assumed that the bias could grow over the reachable set horizon, when in fact as time progresses the controller will be adapting with the goal to drive the reference model bias asymptotically to zero.  An overview of this system architecture is shown in Figure \ref{fig:intro_simple_block_dia}, and more detailed block diagrams are included in Appendix \ref{sec:appendix_a} and Appendix \ref{sec:appendix_b}.

\begin{figure}
    \centering
    \includegraphics[width=1.0\columnwidth]{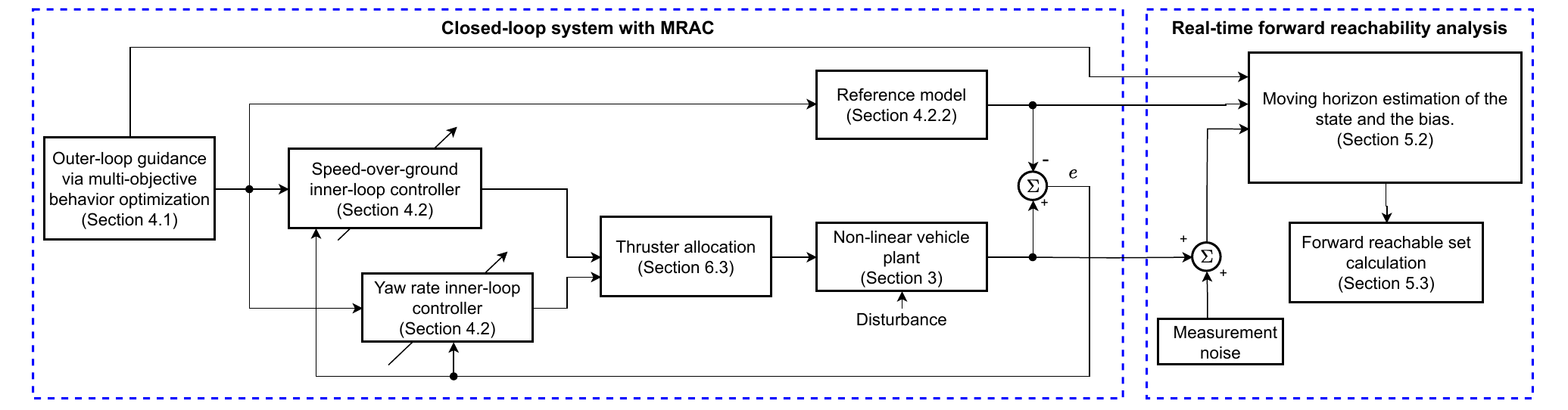}
    \caption{The block diagram of the control and reachability analysis architecture.  The reference model is used in both the \ac{MRAC} approach and the forward reachability analysis with the \ac{MHE}.}
    \label{fig:intro_simple_block_dia}
\end{figure}

\section{Related Work}
\subsection{Adaptive Control in Marine Vehicles.} 
The theoretical foundations of modern adaptive control were formed in the late 1970s and 1980s, and during this period researchers integrated similar but disconnected techniques for specific systems into a more unified framework for mathematical analysis of adaptive systems \cite{Narendra1985GeneralAdaptive,NarendraAnnaswamy2005SAS}. The first experiments using adaptive control for marine robotics followed thereafter in the 1990s and 2000s \cite{YoergerSlotine1991}, \cite{Antonelli2001AdaptiveODIN}, \cite{Skjetne2004AdaptControlShip}, \cite{Smallwood2004AMBC}.  These first experiments were confined to test tanks, and in the case of underwater vehicles, performance was limited by the lack of sensors to directly measure the vehicle-relative translational velocity in the horizontal plane \cite{Fossen1995AC_Comp}.  However, these pioneering studies demonstrated the superior performance of adaptive control when compared with PD control in situations where precise navigation is required for a system with unknown nonlinear dynamics.  

Adaptive control is well suited for the task of precise navigation in marine robotics, and previous work falls into two main categories: indirect and direct adaptive control \cite{NarendraAnnaswamy2005SAS}.  In the case of indirect adaptive control, estimation of the entire plant model is done simultaneously (or offline) with feedforward control \cite{McFarlandWhitcomb2014AMBC_Act_Dyn,Smallwood2004AMBC}. 
Learning the entire model of a marine vehicle online in real time can be challenging due to the presence of non-Gaussian sensor noise, unmodeled dynamics, and limited excitation \cite{McFarlandWhitcomb2014AMBC_Act_Dyn,Harris2023AID}.  This difficulty can be mitigated in part by only estimating modeling error relative to a nominal (known) non-linear plant \cite{Shen2020ATTC_RBF} or linearized plant \cite{Liu2021AC}.  In contrast, direct adaptive control approaches do not explicitly identify the plant parameters.  Instead, the controller is adjusted such that the closed-loop marine vehicle system tracks a desired signal (e.g., \ac{MRAC} \cite{Makavita2020MRAC}) or sliding surface \cite{YoergerSlotine1991,Shen2020ATTC_RBF}. Other types of direct adaptive controllers for marine vehicles include $\mathcal{L}_1$ control \cite{Rober2022L1} and adaptive \ac{RISE} \cite{Fischer2014RISE}.  
In this work we use a direct approach to adaptive control since it is difficult to accurately model the all different types of disturbances encountered in the field. More specifically, we implement an \ac{MRAC} formulation where a baseline optimal \ac{LQR} controller is augmented by an adaptive term to compensate for a range of disturbances with unknown structure.    

Despite the previous theoretical development and laboratory testing with marine vehicles, as well as similar development for robotic arms \cite{SlotineLi1987ATTC_Arm}, mobile ground robots \cite{Fukao2000ATTC_Mobile}, and aerial vehicle applications \cite{Dydek2013AC_UAVs,Bolender2015HIFire}, there are relatively few reports that describe the performance of adaptive controllers in the open water on deployed marine systems.  Of the previously mentioned studies, only \cite{Liu2021AC} and \cite{Fischer2014RISE} report results from limited field experiments.  Although simulation studies are useful in validating the theoretical robustness and stability of adaptive controllers, successful transitions of these controllers onto real marine systems remain difficult. As a related point, adaptive control has not been widely adopted for commercial marine vehicles. 
More effort is required to address the gap between academic research and industry practice. 

The work described herein begins to address this gap by describing the development and testing of direct adaptive control for a \ac{USV} deployed in the Charles River.  Although the Charles River does not meet the technical definition of open water, the portion of the river near the MIT sailing pavilion was an ideal location to perform the necessary iterations of the design-field-evaluate cycle in an outdoor environment for a small \ac{USV}.  
\vspace{-10pt}
\subsection{Forward Reachability Analysis and Safety Assurances} 
Safety assurances can generally be obtained using either testing/data collection~\cite{koopman2016challenges,huang2016autonomous}, or formal analysis~\cite{zhang2018efficient,weng2018towards,xu2020automatic,tjeng2017evaluating,katz2019marabou}.
Testing-based approaches rely on collecting large amounts of data through simulation or deployment of a given system and evaluating if/how the system may fail.
Testing-based approaches are incapable of providing guarantees about safety because it is generally impossible to cover all possible failure cases. 
Alternatively, formal methods can provide guarantees about a given system, but their practical application can be limited due to computational constraints or assumptions about the system that may not reflect its actual behavior.
In this paper, we incorporate online data collection into a formal reachability analysis framework to verify the safety of a system, thus striking a balance between data-driven and formal methods.

In this work we define a USV operation to be safe if the USV does not collide with the shore or other vehicles during the mission.   In the context of formal analysis, this problem is represented mathematically with a model of the dynamics of the vehicle, the action by the controller, and the environment.   To certify safety we must determine if the state of the system, which evolves per the mathematical model of the vehicle dynamics and the control action, can enter a region in the environment that is deemed unsafe.  The formal method for computing bounds for all possible future states of the vehicle, i.e. forward reachable sets, is reachability analysis. \cite{mitchell2007comparing}

Reachability analysis, including Hamilton-Jacobi methods~\cite{bansal2017hamilton,evans1998partial}, Lagrangian methods~\cite{gleason2017underapproximation}, and more recent approaches developed for systems with neural networks (NNs)~\cite{vincent2021reachable,dutta2019reachability,huang2019reachnn,ivanov2019verisig,fan2020reachnn,xiang2020reachable,hu2020reach,sidrane2022overt,everett2021reachability}, determines a set of possible future states, i.e., reachable sets), of a system given a set of possible initial states.
While all reachability analysis techniques handle uncertainty in the state (reflected in the initial state set) and many consider process/measurement noise~\cite{everett2021reachability,bansal2017hamilton}, they typically assume knowledge of the system's behavior under bounded uncertainty.
Moreover, with a few exceptions~\cite{lew2021sampling,althoff2014online,herbert2017fastrack}, reachability analysis is often too computationally expensive for online implementation, especially when the system is high-dimensional.

In this work, we perform online reachability analysis of a system subject to disturbances that are unknown \emph{a priori}. 
Reachable sets are calculated in real time by leveraging $\tt{jax\_verify}$~\cite{jaxverify}, a computational graph analysis tool
capable of efficiently providing output bounds of nonlinear functions.
Disturbances acting on the system are estimated online via \ac{MHE}~\cite{Allgower.Badgwell.ea1999}, and are incorporated into the reachability analysis to predict the behavior of the actual system.

\newpage
\section{Model of USV Dynamics and Kinematics}
\begin{figure*}[t!]
    \centering
    \begin{subfigure}{0.3\textwidth}
        \centering
        \includegraphics[height=2.1in]{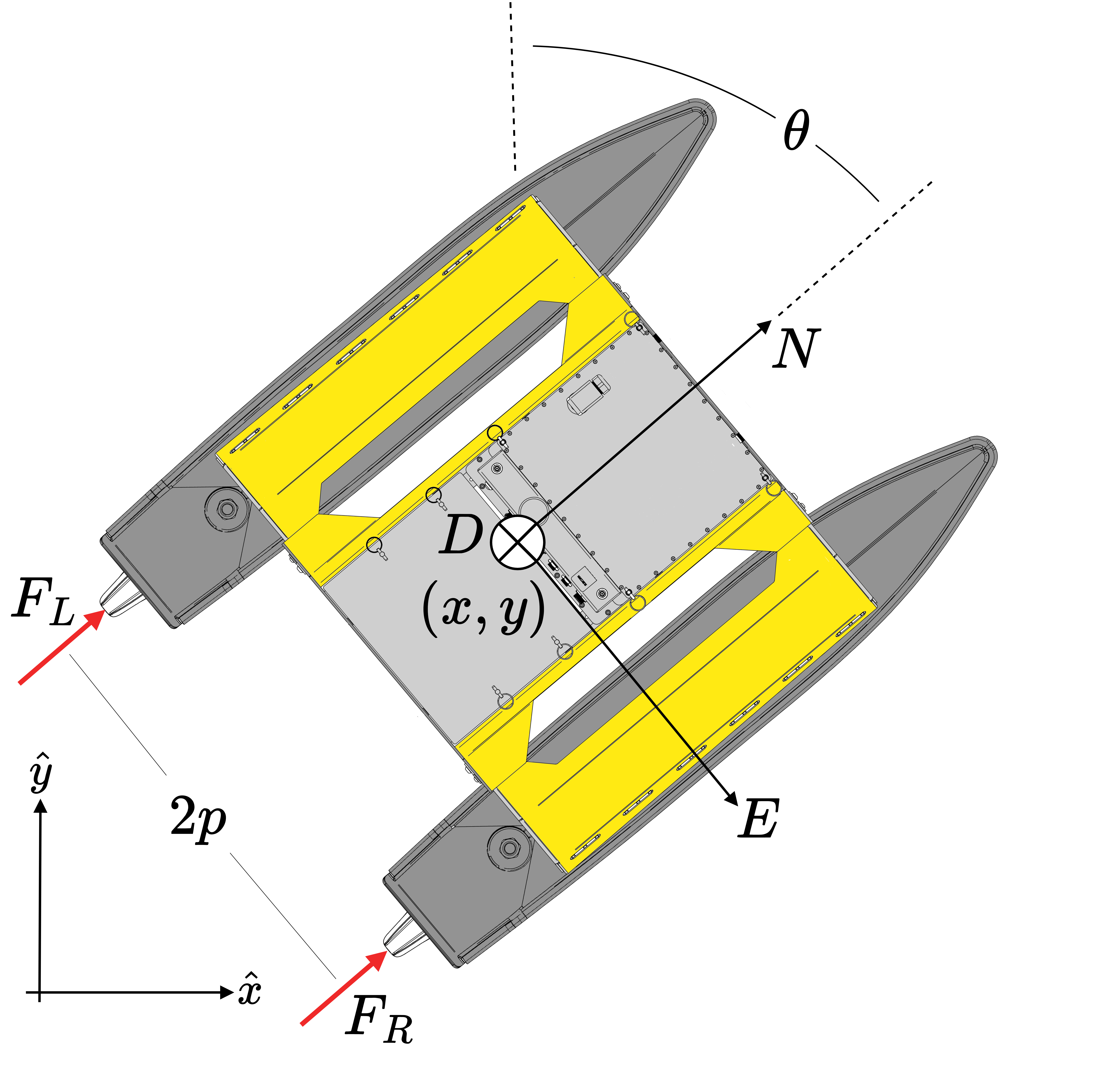}
        \caption{Definition of \ac{NED} and body-fixed coordinate frames}
         \label{fig:frame_def_heron}
    \end{subfigure}%
    ~ 
    \begin{subfigure}{0.35\textwidth}
        \centering
        \raisebox{7mm}{\includegraphics[height=1.2in]{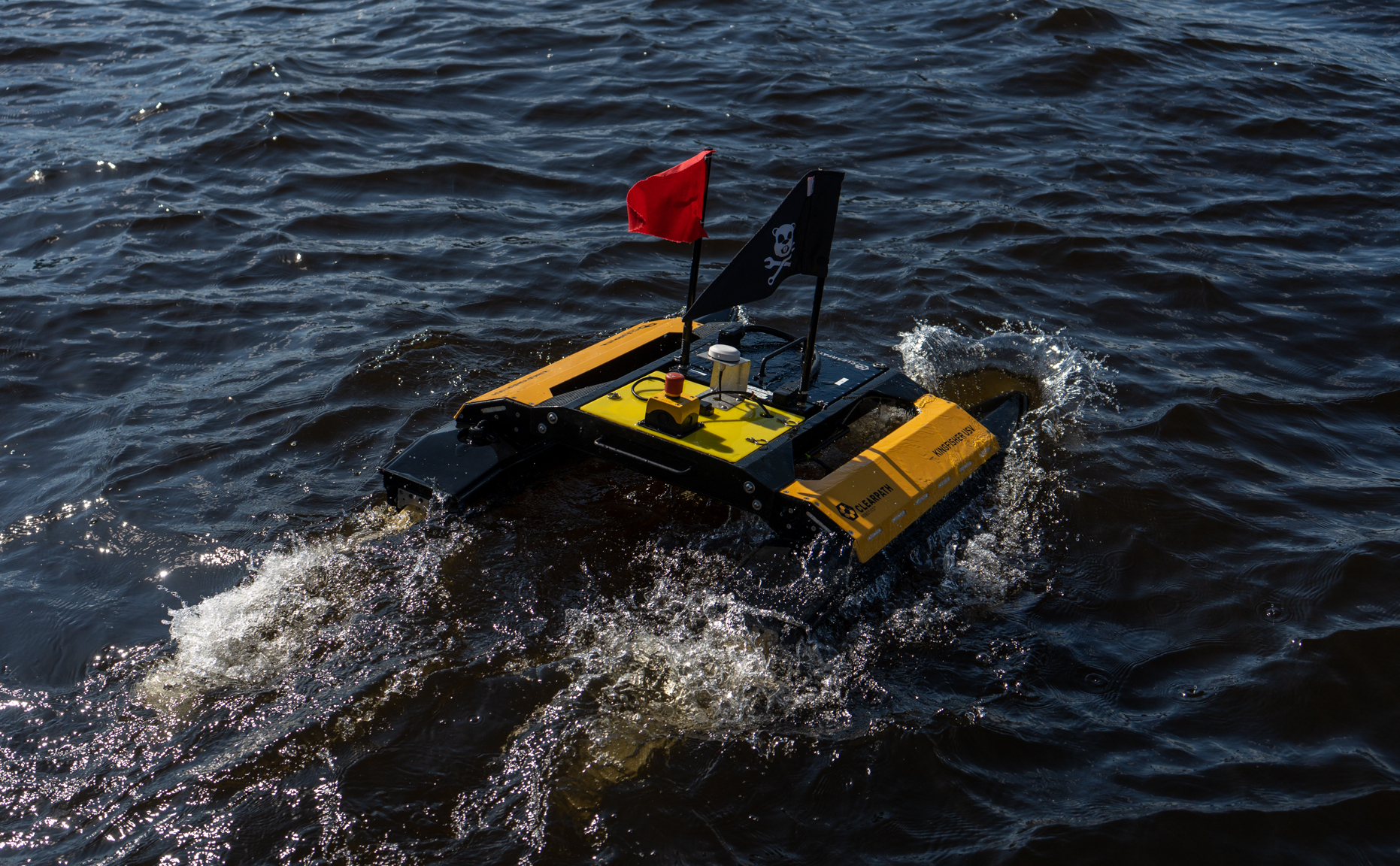}}
        \caption{Rear view}
    \end{subfigure}%
    ~
    \begin{subfigure}{0.3\textwidth}
        \centering
        \raisebox{7mm}{\includegraphics[height=1.2in]{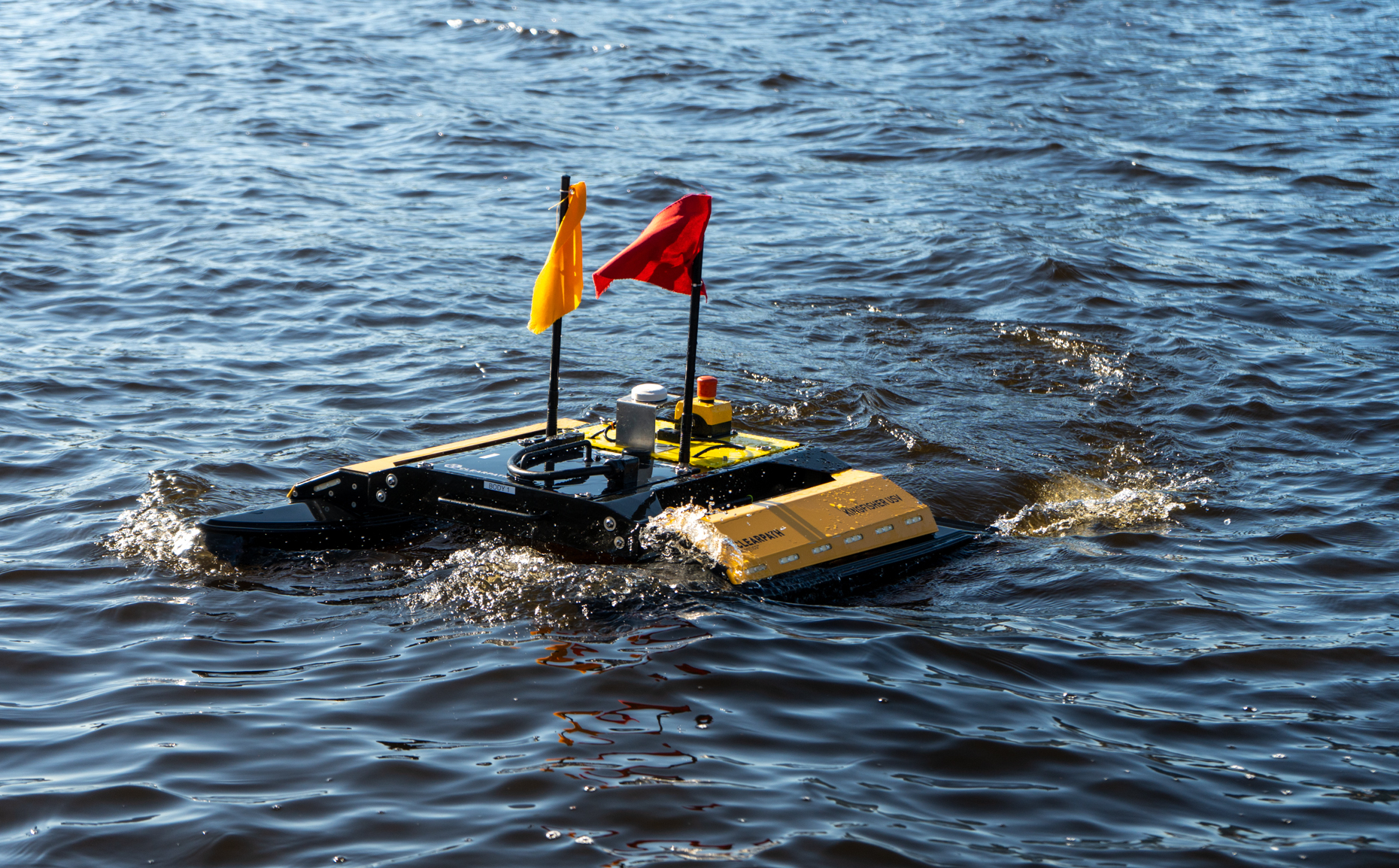}}
        \caption{Front view}
    \end{subfigure}
    \caption{Clearpath Robotics Heron \ac{USV} operated on the Charles River.}
\end{figure*}

The state of the \ac{USV} is represented by 
\begin{itemize}
    \item $\bm{\eta} = \begin{bmatrix} x & y & \theta \end{bmatrix}^T \in {\rm I\!R}^2 \times \mathbb{S}^1$ where $x$ and $y$ denote the position of the \ac{USV} in the \ac{NED} world frame, and $\theta$ is the \ac{USV}'s heading relative to true north as shown in Figure \ref{fig:frame_def_heron}
    \item $\bm{\nu} = \begin{bmatrix} u & v & r \end{bmatrix}^T \in {\rm I\!R}^3$ where $u$ and $v$ are the components of the translational velocity and $r$ is the rotational velocity of the body expressed in the body-fixed frame. 
\end{itemize}
For this class of catamaran \acp{USV}, the control inputs are the commanded rotational speed of the two thrusters.  We define $\bm{u} = \begin{bmatrix} u_L & u_R \end{bmatrix}^T$ where $u_L$ and $u_R$ are the commanded thruster speeds respectively. 

The standard model of kinematics and dynamics of \ac{USV} motion are  \cite{Fossen1994guidance}
\begin{align}
    \bm{\dot{\eta}} =& \begin{bmatrix}
        \bm{R}(\bm{\theta}) & 0_{2 \times 1} \\
        0_{1 \times 2} & 1
    \end{bmatrix}\bm{\nu}, \label{eq:Fossen_kin_model} \\
   \bm{\tau}(\bm{\nu}, \bm{u}) =& \bm{M}\dot{\bm{\nu}} + \bm{C}(\bm{M}, \bm{\nu})\bm{\nu} + \bm{D}(\bm{\nu})\bm{\nu} - \bm{\tau}_{dist}(\bm{\nu}), \label{eq:Fossen_dyn_model}
\end{align}
where $\bm{M}\in {\rm I\!R}^{3\times 3}$ is the inertia matrix with rigid-body and added mass terms, $\bm{C}(\bm{M}, \bm{\nu}): {\rm I\!R}^{3\times 3} \times {\rm I\!R}^{33} \rightarrow {\rm I\!R}^{3\times 3}$ is the Coriolis matrix, $\bm{D}(\bm{\nu}) \in {\rm I\!R}^{3\times 3}$ is the nonlinear drag matrix, and $\bm{\tau}(\bm{\nu}, \bm{u}) = \begin{bmatrix} F_L + F_R & 0 & p ( F_L - F_R) \end{bmatrix}^T \in {\rm I\!R}^3$ is the control induced forces and moments acting on the body.  All other unknown disturbances are captured by $\bm{\tau}_{dist}(\bm{\nu})$.  The rotation matrix $\bm{R}(\bm{\theta}) \in SO(2)$ is the kinematic relationship defined in Section 2.2.1 of \cite{Fossen1994guidance}. 

Herein the model (\ref{eq:Fossen_dyn_model}) is expressed in two ways: 1) a pair of linearized systems about a trim condition is used to develop the adaptive controller, and 2) a discrete model of the closed loop system that includes the controller with uncertainty in observations is used to develop the reachability analysis. Both models retain a term related to the disturbance of the system. 

\subsection{Linearized Models}\label{sec:linearized_model}
We define two linearized models of the dynamics of the Heron \ac{USV} (\ref{eq:Fossen_dyn_model}).  There are decoupled models for both the speed dynamics and sway and yaw-rate dynamics.   This separation is motivated by two factors: First, the Heron \ac{USV} is underactuated since there are only two actuators, but the model (\ref{eq:Fossen_dyn_model}) captures three degrees of freedom.   Second, the higher-level guidance algorithm, detailed in Section \ref{sec:controller_design}, only generates two reference signals: desired speed and trackline heading.

\subsubsection{Linearized Speed Model}
The speed over ground $u_{sog} = \sqrt{u^2 + v^2}$.  We linearize the model (\ref{eq:Fossen_dyn_model}) about a given non-zero trim condition of $u_{sog}$ %with corresponding control input $u_{trim}$%
to find 
\begin{equation}\label{eq:speed_lin_model}
\dot{u}_{sog} = a_{p_1} u_{sog} + b_{p_1} \Lambda_{1} \left( u_{thr} + g_1\left(u_{sog}\right) \right), 
\end{equation} 
where $a_{p_1}, \ b_{p_1} \in {\rm I\!R}$ are nominal model parameters identified a priori, $\Lambda_{p_1} \in {\rm I\!R}_+$ is the unknown thrust control effectiveness modeling uncertainty, and $g_1(u_{sog}):{\rm I\!R} \rightarrow {\rm I\!R}$ is the unknown parametric matched uncertainty.   

The control input $u_{thr} \in {\rm I\!R}$ is applied equally to each thruster. 

\subsubsection{Linearized Sway Yaw-Rate Model}
The sway and yaw-rate model (\ref{eq:Fossen_dyn_model}) are linearized about the trim condition of $v = 0$ and $r = 0$.  The resulting linearized model is
\begin{equation}\label{eq:sway_yawrate_lin_model}
    \begin{bmatrix}
        \dot{v} \\
        \dot{r}
    \end{bmatrix} = A_p 
    \begin{bmatrix}
        v \\
        r
    \end{bmatrix} + B_p \Lambda_{p_2} \big(u_{rud} + g_2(r) \big),
\end{equation}
where $A_p \in {\rm I\!R}^{2 \times 2}$ is an constant state transition matrix identified a priori, $B_p \in {\rm I\!R}^{2 \times 1}$ is the constant and known control effectiveness matrix, $\Lambda_{p_2} \in {\rm I\!R}_+$ is the unknown thrust control effectiveness modeling uncertainty, and $g_2(r):{\rm I\!R} \rightarrow {\rm I\!R}$ is the unknown parametric matched uncertainty.  

\subsubsection{Discussion}
The transformation of the nonlinear model (\ref{eq:Fossen_dyn_model}) into the two decoupled models of speed (\ref{eq:speed_lin_model}) and sway-heading (\ref{eq:sway_yawrate_lin_model}) allows  the control problem to be expressed in a 2D workspace (speed error and heading error).  This approach is considered common practice in the literature \cite{Fossen1994guidance}, but it has limitations.  For instance, the disturbances that act in the sway degree of freedom are not captured in either $g_1(u_{sog})$ nor $g_2(r)$. This limitation has no practical consequence since the Heron \ac{USV} is not directly actuated in the sway degree of freedom. Another limitation of this approach is that more complex disturbances, such as those that are nonlinear functions of both $u_{sog}$ and $r$, are not captured; we leave this issue for future work.  Finally, the formulation assumes the disturbances in the plant model (\ref{eq:Fossen_dyn_model}) can be represented in the linearized models by terms $g_1(u_{sog})$ or $g_2(r)$ which act in the control space of $u_{thr}$ and $u_{rud}$ respectively.

\subsection{Discrete Closed-Loop Model}
Consider a state-feedback controller $u_t = \pi(x_t)$, which is described later in Section \ref{sec:controller_design}. 
The closed loop dynamics of the system (\ref{eq:Fossen_kin_model}) and (\ref{eq:Fossen_dyn_model}) with the control policy $\pi(x_t)$ can be expressed as a discrete general nonlinear function with a disturbance term $\bm{w} \in {\rm I\!R}^{6}$.
\begin{align}
    \bm{x}_{t+1} = \begin{bmatrix}
        \bm{\eta}_{t+1} \\
        \bm{\nu}_{t+1}
    \end{bmatrix}=& f_{cl}(\bm{x}_{t}; \pi) + \bm{w}_{t} \\
    \bm{y}_{t} =& h(\bm{x}_{t}) + \bm{w}_{y_{t}},
    \label{eq:nonlinear_cl_dyn}
\end{align}
where $\bm{y}_{t} \in {\rm I\!R}^{n_y}$ is the measured system corrupted by noise $\bm{w}_{y_{t}}$. 

In this work we assume $\mathbf{w}_t \sim \mathcal{N}(\bm{\mu}_t, \bm{W}_t)$ and $\mathbf{w}_{y_t} \sim \mathcal{N}(\mathbf{0}, \bm{W}_{y_t})$.  Furthermore, 
we assume that $\bm{W}_t$ is diagonal.
Since the additive bias to the nominal system dynamics $\mathbf{w}_t$ is time-varying, we impose a condition that 
\begin{equation} \label{eqn:noise_assumption}
    \begin{split}
        & | (\bm{\mu}_{t+1})_i - (\bm{\mu}_t)_i | \leq \Delta\bm{\mu}_i \\
        & | (\bm{\sigma}_{t+1})_i - (\bm{\sigma}_t)_i | \leq \Delta\bm{\sigma}_i,
    \end{split}
\end{equation}
where $(\bm{\sigma}_{t})_i \in \mathbb{R}$ for $i =  1, \ldots , n_x$ is the standard deviation associated with the covariance matrix $\bm{W}_t$, i.e., $(\bm{W}_t)_{ii} = (\bm{\sigma}_t)_i^2$,
%$\bm{\sigma}_t = {\tt sqrt(diag}(\bm{W}_t))$,
and $\Delta\bm{\mu}_{i}, \Delta\bm{\sigma}_{i} \in \mathbb{R}^{n_x}$ represent the bounded possible variation in $\bm{\mu}_t$ and $\bm{\sigma}_t$, respectively, per time step.

%% LyX 2.3.6.1 created this file.  For more info, see http://www.lyx.org/.
%% Do not edit unless you really know what you are doing.

\section{\label{sec:controller_design}Control Architecture}

The design of the control system can be divided into two levels: 
\begin{itemize}
\item \textbf{Outer-loop guidance algorithms as behaviors} that generate a reference trajectory from mission objectives. 

\item \textbf{Inner-loop controllers} that are designed to drive the actual state of the vehicle to the reference trajectory.
\end{itemize}
Guidance algorithms are more specific to the mission requirements and we report three different algorithms used in the experiments. The implementation of \ac{LQR}-\ac{PI} and \ac{MRAC} inner-loop controllers are subsequently described. 

\subsection{\label{sec:guidanceBehaviorDesign} Guidance Behavior Design}
In this work the guidance algorithms were implemented as behaviors within the Interval Programming (IvP) multi-objective behavior optimization architecture MOOS-IvP \cite{Benjamin2010MOOSIvP}. 
The optimization is stated as
\begin{equation}
    u_{DES}, \theta_{DES} = \substack{\text{arg max} \\ u \in S_u, \  \theta \in S_\theta} \sum_{q=1}^{N_{active}} w_{q} f_{q}(\eta, \nu)
\end{equation}
where $u_{DES}$ is the desired speed selected from domain $S_u$, $\theta_{DES}$ is the desired heading selected from domain $S_\theta$.  The optimization is over the sum of $N_{active}$ utility functions $f_{q}(\eta, \nu)$ each generated by a behavior and also weighted by $w_{q}$.  Details regarding the specific behaviors used in this work are described in \cref{sec:guidanceBehaviorDesign}.

The following sections outline the guidance algorithms used in the experiments described in Sections \ref{sec:experimental_results_sys_eval} and \ref{sec:experimental_results_scenarios}. The baseline track-line following behavior is a PD controller that enables straight line tracking. $L_{1}$ Guidance \cite{Park.Dyest.ea2004} was used for more complex smooth paths. The method in \cite{Skejic.Brevik.ea2009} enabled the Underway Replenishment missions.

\subsubsection{\label{sec:baseline_trackline_guidance}Baseline Track-line Following }

The objective of the baseline track-line guidance algorithm is to have the vehicle track a user-selected path in the Earth-fixed frame using a proportional derivative (PD) error feedback approach. The baseline track-line following guidance algorithm is described in \cite[Section 9.7]{IvPManPages}. 

\subsubsection{L1 Guidance}\label{sec:L1_guidance}

The $L_{1}$ guidance algorithm \cite{Park.Dyest.ea2004} has the same objective as the baseline track-line following algorithm described in Section (\ref{sec:baseline_trackline_guidance}), but is better suited for missions with more complex (e.g., curved) paths \cite{Park.Dyest.ea2004}. The approach was modified to realize the improvement on track-line following for the underactuated Heron USV. Specifically, in \cite{Park.Dyest.ea2004} the term related to lateral (i.e., sway) acceleration command is
\begin{equation}
a_{cmd}=2\frac{u_{sog}^{2}}{L_{1}}\sin\eta,
\end{equation}
where $L_{1}\in\mathbb{R}_{\geq0}$ is the length of a line from ownship position to a reference point on the desired trajectory and $\eta$ is the angle between the course over ground and the line. However, the Heron USV is not actuated in the sway direction and cannot generate a lateral acceleration directly. The vehicle, however, may rotate  while moving at speed $u_{sog}$  to generate lateral acceleration. Using the coupling of angular and forward velocities, the desired yaw rate command is 
\begin{equation}
r_{cmd}=2\frac{u_{sog}}{L_{1}}\sin\eta.
\end{equation}
While this guidance algorithm is similar to approaches such as proportional guidance and line-of-sight guidance for certain paths, it includes a lookahead element that enables tight tracking on curved trajectories.

\subsubsection{Underway Replenishment (UNREP)}\label{sec:fossen_unrep_guidance}
As described in \Cref{sec:UNREP_results}, the UNREP mission requires a follower vessel to maneuver alongside a lead vehicle and maintain a constant relative position and heading to transfer goods between two vessels. The objective of the target tracking for UNREP guidance is to minimize errors in both the cross-track and inline degrees of freedom. In this study we use the algorithm presented in \cite{Skejic.Brevik.ea2009} to perform the UNREP mission. 

\subsubsection{Heading to Yaw Rate Controller} \label{sec:helmfilter}
The optimizer in the IvPHelm \cite{Benjamin2010MOOSIvP} combines the output from individual behaviors into a single decision of heading and speed. Optimization is performed over discrete domain intervals, and by definition the raw output from IvPHelm is discontinuous. To address this issue, the discrete output from the IvPHelm is filtered to generate a smooth reference trajectory for the inner-loop controller. Given a decision from the IvPHelm of $u_{DES}$ and $\theta_{DES}$, the reference trajectories for speed $u_{des}$ and yaw rate $r_{des}$ are defined as
\begin{align}
\begin{bmatrix}\dot{q}_{1}\\
\dot{q}_{2}
\end{bmatrix}= & \begin{bmatrix}-\frac{1}{\tau_{u}} & 0\\
0 & -\frac{1}{\tau_{r}}
\end{bmatrix}\begin{bmatrix}q_{1}\\
q_{2}
\end{bmatrix}+\begin{bmatrix}u_{DES}\\
\tau_{\theta}\Delta\theta
\end{bmatrix}\\
\begin{bmatrix}u_{des}\\
r_{des}
\end{bmatrix}= & \begin{bmatrix}\frac{1}{\tau_{u}} & 0\\
0 & \frac{1}{\tau_{r}}
\end{bmatrix}\begin{bmatrix}q_{1}\\
q_{2}
\end{bmatrix}
\end{align}
 where $\Delta\theta$ is computed using the following method: Let
\begin{align*}
\bm{e}_{\theta} & =\begin{bmatrix}\sin\left(\theta\right)\\
\cos\left(\theta\right)\\
0
\end{bmatrix}, \quad \text{and} \quad
\bm{e}_{\theta_{DES}} =\begin{bmatrix}\sin\left(\theta_{DES}\right)\\
\cos\left(\theta_{DES}\right)\\
0
\end{bmatrix},
\end{align*}
and $\bm{e}_{err}=\bm{e}_{\theta}\times\bm{e}_{\theta_{DES}}$ be the unit vectors in the direction of the current vehicle heading, desired heading, and cross product of the two vectors respectively. Then 
\begin{equation}
\Delta\theta=\arccos(\bm{e}_{\theta}^{T}\bm{e}_{\theta_{DES}})\cdot(-\mathrm{sgn}(\bm{e}_{err_{3}}))
\end{equation}

\subsection{Control Design}
The following section describes the structure of the inner-loop controllers used to facilitate the tracking of the reference trajectory from the guidance behaviors.  The synthesis of the speed controller and sway and yaw-rate controller is identical.   For brevity, this section only presents the development of the more complex sway and yaw-rate controller. Implementation details are provided for both controllers in Section \ref{sec:implementation}. 

We present the formulation for the yaw-rate controller, which is identical to the speed controller development.

Consider the linearized model of \ac{USV} motion in the sway and yaw-rate degrees of freedom (\ref{eq:sway_yawrate_lin_model}) where the uncertainty is parameterized as 
\begin{equation}
g_2(r) = \bm{\Theta}^{T}\bm{\Phi}\left(r\right)
\end{equation}
where $\bm{\Theta}\in\mathbb{R}^{N}$ is a vector of unknown constant parameters, and $\bm{\Phi}\left(r\right):\mathbb{R}\rightarrow\mathbb{R}^{N}$ is the known and locally Lipschitz $N$-dimensional regression vector. The linearized dynamics (\ref{eq:sway_yawrate_lin_model}) are expressed as 

\begin{equation}
    \begin{bmatrix}
        \dot{v} \\
        \dot{r}
    \end{bmatrix} = A_p 
    \begin{bmatrix}
        v \\
        r
    \end{bmatrix} + B_p \Lambda_{p_2} \big(u_{rud} + \bm{\Theta}^{T}\bm{\Phi}\left(r\right) \big)
\end{equation}

In the case of the linearized model of the Heron USV (\ref{eq:sway_yawrate_lin_model})  the pair $\left(A_{p},B_{p}\Lambda_{p_2}\right)$ is controllable. (See Section \ref{sec:implementation:yaw_rate_controller} for details.)

The desired control objective is to perform command tracking, meaning the control input $\bm{u}$ must be enable $y\in\mathbb{R}$ defined as
\begin{align}
y & \triangleq C_{p}\begin{bmatrix}
        v \\
        r
    \end{bmatrix},
\end{align}
to track the bounded and time-varying command $y_{cmd_r} = r_{des} \in \mathbb{R}$ as defined in Section \ref{sec:helmfilter}, with bounded errors in the presence of system uncertainties in $\left\{ A_{p},\Lambda,\Theta\right\} $. Furthermore, $C_{p} = \begin{bmatrix} 0 & 1 \end{bmatrix}$ is known and constant).  The integrated error that facilitates output tracking is
\begin{equation}
    e_{yI} = \int_{0}^{t} \left(y(\tau) - y_{cmd_r}(\tau) \right) d \tau. 
\end{equation}
%$\bm{y}_{cmd}:\mathbb{R}_{\geq0}\rightarrow\mathbb{R}^{m}$

To facilitate the tracking objective, the dynamics are expressed in the extended open-loop form  \cite[Chapter 10]{Lavretsky.Wise2012} as
\begin{align}
\begin{bmatrix}
    \dot{e}_{yI}  \\
    \dot{v} \\
    \dot{r}
\end{bmatrix} & =\underbrace{\begin{bmatrix}
    0 & C_p \\
    0_{2 \times 1} & A_p
\end{bmatrix}}_{A}
\begin{bmatrix}
    e_{yI}  \\
    v \\
    r
\end{bmatrix}+
\underbrace{\begin{bmatrix}
    0 \\
    B_p
\end{bmatrix}}_{B} \Lambda_{p_2} \left( u + \bm{\Theta}^{T}\bm{\Phi}\left(r\right)\right)+
\underbrace{\begin{bmatrix}
    - 1 \\
    0_{2 \times 1}
\end{bmatrix}}_{B_{ref}}y_{cmd},\label{eq:plant-1}
\end{align}

The controller $u_{rud}$ for the yaw-rate dynamics is the sum of the baseline controller $u_{bl}$
and an adaptive component $u_{ad}$ which are subsequently defined:
\begin{align}
u_{rud} & =u_{bl}+u_{ad}.\label{eq:utotal}
\end{align}

\subsubsection{Baseline Optimal LQR PI Controller}\label{sec:LQR_PI}

First, the baseline controller $\bm{u}_{bl}$ is constructed. Consider the dynamics in (\ref{eq:plant-1}) where $\Lambda_{p2} = 1$ and $\Theta=0_{N\times 1}$ as
\begin{align}
\begin{bmatrix}
    \dot{e}_{yI}  \\
    \dot{v} \\
    \dot{r}
\end{bmatrix} & =A
\begin{bmatrix}
    e_{yI}  \\
    v \\
    r
\end{bmatrix}+
B u +
B_{ref}y_{cmd},\label{eq:baselineExtendedModel}
\end{align}

Using the extended form of the dynamics in (\ref{eq:baselineExtendedModel}), a \ac{LQR} \ac{PI} feedback controller was designed using the robust \ac{LQR} \ac{PI} servomechanism design \cite[Eqs. 10.37-10.43]{Lavretsky.Wise2012}. The baseline component of the controller $u_{bl}\in\mathbb{R}$ is
\begin{align}
u_{bl} & \triangleq-\bm{K}_{LQR}^{T}\begin{bmatrix}
    e_{yI}  \\
    v \\
    r
\end{bmatrix}.
\end{align}
The baseline controller enables the system to perform reference command tracking without model uncertainty (e.g. control effectiveness and matched uncertainty). The aforementioned modeling errors are expected to degrade controller performance. 

\subsubsection{Adaptive Augmentation of Optimal Baseline Controller}
The baseline controller $u_{bl}$ is  designed to provide desired system responses (e.g., rise time, overshoot, and bandwidth). However, adding uncertainty degrades the desired response. Hence, the subsequent adaptive controller design will facilitate restoration of the degraded system response. 

Consider the reference model dynamics
\begin{align}
\begin{bmatrix}
    \dot{e}_{yI_{ref}}  \\
    \dot{v}_{ref} \\
    \dot{r}_{ref}
\end{bmatrix} & =A_{ref}
\begin{bmatrix}
    e_{yI_{ref}}  \\
    v_{ref} \\
    r_{ref}
\end{bmatrix}+
B_{ref}y_{cmd_r},
\end{align}
where $A_{ref}\triangleq A-BK_{LQR}^{T}$. The reference model dynamics represent the system response using only the baseline controller, i.e. $u = u_{bl}$, and without modeling errors. i.e. $\Lambda=1$ and $\bm{\Theta}=0_{N \times 1}$.

In the presence of model uncertainty, an adaptive control term $u_{ad}$ to augment the baseline controller is constructed. The adaptive augmentation component is designed as 
\begin{align}
u_{ad} & =-\bm{\hat{\overline{\Theta}}}^{T} \bm{\overline{\Phi}}\left(u_{bl},\ r\right),
\end{align}
where $\bm{\hat{\overline{\Theta}}}\in\mathbb{R}^{N+1}$ is an adaptive term that compensates for an extended matrix of unknown parameters, and $\bm{\overline{\Phi}}\left(u_{bl},\ r\right):\mathbb{R}\times\mathbb{R}^{N}\rightarrow\mathbb{R}^{N+1}$ is a regression matrix defined as
\begin{align}
\bm{\overline{\Phi}}\left( u_{bl},\,r \right) & \triangleq\left[u_{bl},\,\bm{\Phi}\left( r \right)^{T}\right]^{T}.
\end{align}

Following the development in \cite[Ch. 10.3]{Lavretsky.Wise2012}, the continuous-time adaptive update law is designed as
\begin{align}
\dot{\bm{\hat{\overline{\Theta}}}} & =-\Gamma_{\Theta}\bm{\overline{\Phi}}\left(u_{bl},\, r \right)\bm{e}^{T}PB,\label{eq:thetaHatDot}
\end{align}
where $\bm{e}\triangleq \begin{bmatrix}
    e_{yI}  \\
    v \\
    r 
\end{bmatrix}-\begin{bmatrix}
    e_{yI_{ref}}  \\
    v_{ref} \\
    r_{ref}
\end{bmatrix}$, the adaptation gain $\Gamma_{\Theta}\in\mathbb{R}^{\left(N+1\right)\times\left(N+1\right)}$, and $P\in\mathbb{R}^{3\times 3}$ is a positive definite solution to the Lyapunov equation
\begin{align}
PA_{ref}+A_{ref}^{T}P & =-Q\label{lyapalgequn},
\end{align}
for a user-selected positive definite $Q \in\mathbb{R}^{3\times 3}$.

\subsubsection{MRAC Implementation for Heron USV}\label{sec:MRAC_def}
The controller that is applied to the system is the sum of the baseline and adaptive controller
\begin{align}
u & = u_{bl} + u_{ad}.
\end{align}
The stability analysis for this controller and dynamics can be found in \cite[Ch. 10]{Lavretsky.Wise2012}.  This baseline adaptive controller with adaptive augmentation architecture was used for both speed and yaw rate tracking control objectives. In Section \ref{sec:implementation} we report the numerical values used for controller implementation. The baseline controller, adaptive controller, and adaptive updates laws were designed in continuous-time; however, they were implemented in discrete-time using forward Euler integration.

\section{Real-Time Forward Reachability Calculation}

In this section describes the formulation introduced in \cite{rober2023online} for discrete-time forward reachability analysis to determine the safety of our system in the face of \emph{a priori} unknown disturbances, which are estimated online. We later build on this work by demonstrating the effectiveness of this real-time reachability formulation on a diverse set of fielded experiments.

We define the forward reachable set at time $t+1$ of the system with initial state set $\mathcal{X}_0$ and closed loop dynamics model $f_{cl}(\bm{x}_{t}; \pi)$ in (\ref{eq:nonlinear_cl_dyn}) as 
\begin{equation}
    \mathcal{R}_{t+1}(\mathcal{X}_0) = f_{cl}(\mathcal{R}_t(\mathcal{X}_0); \pi).
\end{equation}
where $\mathcal{R}_0(\mathcal{X}_0) = \mathcal{X}_0$ and $f_{cl}(\mathcal{R}_t(\mathcal{X}_0); \pi)$ is shorthand for $\{f_{cl}(\bm{x}_t; \pi) \ \vert\ \bm{x}_t \in \mathcal{R}_t(\mathcal{X}_0)\}$.
The exact reachable set $\mathcal{R}_t$ is expensive to compute.
Instead, we compute \acp{RSOA}, i.e., $\ourbar{\mathcal{R}}_t \supseteq \mathcal{R}_t$, as is specified in \cref{sec:Comp_graphs_and_linear_relax}.
The \acp{RSOA} $\{\ourbar{\mathcal{R}}_{t}, \ourbar{\mathcal{R}}_{t+1}, \ldots, \ourbar{\mathcal{R}}_{t+\tau_r}\}$, denoted as $\ourbar{\mathcal{R}}_{t:t+\tau_r}$, are used to verify safety of the system over a horizon $\tau_r$ by checking if the future states can reach an unsafe region of the state space $\mathcal{C} \subset \mathbb{R}^{3}$.
If there is an $ i \in \mathcal{T} = \{t, \ldots, t+\tau_r\}$ such that $\ourbar{\mathcal{R}}_i \bigcap \mathcal{C} \neq \emptyset$, the system may enter the unsafe region and safety is not guaranteed.

Note that while RSOAs are capable of verifying safety, more conservative RSOAs make the verification conditions more difficult to satisfy, so tight RSOAs are preferred.

Our approach is designed to be executed online, regularly generating \acp{RSOA} over a finite time horizon at a fixed interval.
The proposed data-driven reachability approach is summarized as:
\begin{enumerate}
    \item First, before the deployment of the system, a \ac{CG} is constructed to represent the system dynamics $f_{cl}$.
    \item At each time step during runtime, we use the \ac{MHE} to obtain an estimate of the state $\hat{\bm{x}}_t$ and estimates of the most recent mean and covariance values of $\bm{w}_t$.
    \item Finally, the mean and covariance estimates from the \ac{MHE} are input into a \ac{CG} relaxation to conduct reachability analysis, thus predicting the behavior of the actual system.
\end{enumerate}

\subsection{Computational Graphs and Linear Relaxation}\label{sec:Comp_graphs_and_linear_relax}
A \acf{CG} $\bm{G}$ is a directed acyclic graph with nodes $\bm{V} = \{V_1, V_2, ..., V_n\}$ and edges $\bm{E}$. Each edge connects a pair of two nodes $(V_i, V_j)$ where the output of $V_i$ is the input to $V_j$. 
Each node is associated with a basic computation function $G_i(\cdot)$. 
We define the output of node $V_i$ as 
\begin{equation}
    g^{\bm{G}}_i = G_i(u(V_i))
\end{equation}
where $u(V_i)$ is the set of inputs to $V_i$ or equivalently, the outputs from nodes with edges directed toward $V_i$. 
Let the inputs to the graph $\bm{G}$ be $\mathbf{z} \in$ $\mathbb{R}^{n_i}$ and the graph has single output node $V_o$ with dim$(V_o) = \mathbb{R}^{n_o}$. 
Using this, the output of node $V_i$ as a function of $\mathbf{z}$, i.e. $g^{\bm{G}}_i = g^{\bm{G}}_i(\mathbf{z})$. 
Hence, the output of the computational graph is $g^{\bm{G}}_o$.
Given a set of computational graph inputs $\mathcal{I}$, we want to describe the set of possible outputs $\mathcal{O} = \{g^{\bm{G}}_o(\mathbf{z}_0) \ \vert \ \mathbf{z}_0 \in \mathcal{I} \}$. We construct $\mathcal{I}$ as a hyper-rectangular set, defined by 
\begin{equation}
    \mathcal{B}_{\infty}(\ourbar{\mathbf{z}}_0, \bm{\epsilon}) \triangleq \{\mathbf{z} \ \lvert\ \| (\mathbf{z} - \ourbar{\mathbf{z}}_0) \oslash \bm{\epsilon} \|_\infty \leq 1\},
\end{equation}
where $\ourbar{\mathbf{z}}_0\in\mathbb{R}^{n_i}$ is the center of the $l_{\infty}$-ball, $\bm{\epsilon}\in\mathbb{R}^{n_i}_{\geq 0}$ is a vector of radii for the corresponding elements of $\mathbf{z}$, and $\oslash$ denotes element-wise division.  Bound on the output set $\mathcal{O}$ can be found using the following theorem. 
\begin{theorem}[Linear Relaxation of CGs, Thm.~2~\cite{xu2020automatic}] \label{thm:lirpa}
Given a \ac{CG} $\bm{G}$ and a hyper-rectangular set of possible inputs $\mathcal{I}$, there exist two explicit functions 
\begin{equation*}
    g^{\bm{G}}_{L,o}(\bm{z}) = \bm{\Psi}\bm{z} + \bm{\alpha},\quad g^{\bm{G}}_{U,o}(\bm{z}) =  \bm{\Phi}\bm{z} + \bm{\beta}
\end{equation*}
such that the inequality $g^{\bm{G}}_{L,o}(\bm{z}) \leq g^{\bm{G}}_o(\bm{z}) \leq g^{\bm{G}}_{U,o}(\bm{z})$ holds element-wise for all $\bm{z} \in \mathcal{I}$,
with $\bm{\Psi}, \bm{\Phi} \in \mathbb{R}^{n_o \times n_i}$ and $\bm{\alpha}, \bm{\beta} \in \mathbb{R}^{n_o}$.
\end{theorem}
These bounds on $g^{\bm{G}}_o(\bm{z})$ allow us to define $\ourbar{\mathcal{O}}$, a hyper-rectangular over-approximation of the output set $\mathcal{O}$, as 
\begin{equation}
    \mathcal{O} \subseteq \ourbar{\mathcal{O}} = \{\bm{o} \ \vert\ g^{\bm{G}}_{L,o}(\mathbf{z}) \leq \bm{o} \leq g^{\bm{G}}_{U,o}(\mathbf{z}),\ \exists \mathbf{z} \in \mathcal{I}\}.
\end{equation}
To conduct reachability analysis, we generate \acp{RSOA} using the \ac{CG} analysis tool {\tt jax\_verify}~\cite{jaxverify}.
Other CG analysis tools, such as Auto-LiRPA~\cite{xu2020automatic} could also be used, but {\tt jax\_verify} supports just-in--time compilation and is thus fast enough to enable online computation.
By specifying the control policy $\pi$ and the discrete representation of the open-loop dynamics (\ref{eq:speed_lin_model}) and (\ref{eq:sway_yawrate_lin_model}) as functions in the {\tt jax\_verify} framework, we can generate a \ac{CG} representation of the discrete closed-loop model (\ref{eq:nonlinear_cl_dyn}).  Moreover, we can use results from Theorem \ref{thm:lirpa} for linear relaxation \cite{xu2020automatic} to obtain bounds on $\hat{\bm{x}}_{t+1}$ from a set of possible $\hat{\bm{x}}_t$ and thus find the \ac{RSOA} for the closed-loop discrete Heron \ac{USV} system model (\ref{eq:nonlinear_cl_dyn}). 

\subsection{Moving Horizon Bias Estimation} \label{sec:MHE_def}
To characterize disturbances to the system, we employ a \acf{MHE}~\cite{Allgower.Badgwell.ea1999} to estimate the difference between nominal system dynamics and the actual system behavior. This allows us to represent both internal changes to the system (i.e. thruster imbalances or inefficiencies) and external disturbances (i.e. wind or strong currents) as an additive bias to the nominal system dynamics. 
After collecting data measurements $\bm{y}_{t-\tau_e:t}$ in a receding horizon window from time $t-\tau_e$ to $t$, we perform the following optimization to find estimates of the state $\hat{\bm{x}}$ and uncertainty about the state dynamics $\hat{\mathbf{w}}$ 
\begin{equation}
    \label{eqn:mhe_optim}
    \min_{
        \hat{\bm{x}}_{t-\tau_e:t},
        \hat{\mathbf{w}}_{t-\tau_e:t}} J,
\end{equation}
where 
\begin{equation}
    \label{eqn:mhe_cost_fn}
    J = \|\hat{\bm{x}}_t - \ourbar{\bm{x}}_t\|^2_{\bm{Q}_{t|t-1}^{-1}} + \sum_{k=t - \tau_e}^{t} \| \bm{y}_k - h(\hat{\bm{x}}_k) \|^2_{\bm{W}_{y}^{-1}} + \|\hat{\bm{w}}_k\|^2_{\bm{W}_k^{-1}}.
\end{equation}
In (\ref{eqn:mhe_cost_fn}), $\ourbar{\bm{x}}_t$ is the state estimate prior, $\bm{Q}_{t|t-1} \in \mathbb{R}^{3 \times 3}$ is the state uncertainty prior,  and $\|\hat{\bm{x}}_t - \ourbar{\bm{x}}_t\|^2_{\bm{Q}_{t|t-1}^{-1}} \triangleq (\hat{\bm{x}}_t - \ourbar{\bm{x}}_t)^\top \bm{Q}_{t|t-1}^{-1} (\hat{\bm{x}}_t - \ourbar{\bm{x}}_t).$
The result of \cref{eqn:mhe_optim} are values of $\hat{\bm{x}}_{t-\tau_e:t}$ and $\hat{\bm{w}}_{t-\tau_e:t}$ that optimally estimate the state and disturbance terms over the given window.
Much like model predictive control~\cite{borrelli_bemporad_morari_2017}, the typical motivation for using a \ac{MHE} is to execute \cref{eqn:mhe_optim} at each discrete time step to determine $\hat{\bm{x}}$.
Each iteration of the process calculates a new estimate of $\hat{\bm{x}}$ with \cref{eqn:mhe_optim}, and the priors of the state estimate $\ourbar{\bm{x}}_{t+1}$ and covariance $\bm{Q}_{t+1|t}$ are generated for the next time step using the update law 
\begin{align}
    \ourbar{\bm{x}}_{t+1} =& f_{cl}(\hat{\bm{x}}_t; \pi) + \hat{\bm{w}}_t  \label{eq:posterior_state_est}\\
    \bm{Q}_{t|t} =& \left(\bm{Q}_{t|t-1}^{-1} + \bm{H}^\top \bm{R}^{-1} \bm{H}\right)^{-1} \\
    \bm{Q}_{t+1|t} =& \bm{A} \bm{Q}_{t|t} \bm{A}^\top + \bm{W}_t,
\end{align}
where $\bm{A} = \frac{\partial f_{cl}}{\partial \bm{x}}\big|_{\bm{x}=\hat{\bm{x}}_t}$, $\bm{H} = \frac{\partial h}{\partial \bm{x}}\big|_{\bm{x}=\hat{\bm{x}}_t} = \bm{I}_{3 \times 3}$ and $\bm{R}\in \mathbb{R}^{3 \times 3}$ is the noise covariance matrix for measurement sensing error.  
\subsection{Reachable Set Calculation}
At each time step $t$, the forward reachable sets are calculated using the \ac{CG} assembled before the mission and the results from the \ac{MHE} that are updated in real time.  
As described in \cref{sec:MHE_def}, we collect the state estimate $\hat{\bm{x}}_t$ and generate the posterior $\ourbar{\bm{x}}_{t + 1}$ per equation (\ref{eq:posterior_state_est}).  
We also make the approximations $\hat{\bm{\mu}}_t = {\tt mean} (\hat{\bm{w}}_{t-\tau_e:t})$ and $\hat{\bm{W}}_t = {\tt cov} (\hat{\bm{w}}_{t-\tau_e:t})$.
\subsubsection{Computational Graph with Augmented State}
Since in this formulation of forward reachabilty analysis we consider a (bounded) range of possible future disturbances given (\ref{eqn:noise_assumption}), we augment the state input to the \ac{CG} by the addition of terms for the disturbance and its assumed rate of change.  We also add into our augmented state the rate of change of the standard deviations associated with each state, which remain constant over the time horizon.
Thus, we introduce the augmented \ac{CG} $\bm{G_{cl}}$ with input 
\begin{equation}
    \bm{\ourbar{z}}_t = \begin{bmatrix} 
        \tilde{\bm{x}}_t \\
        \tilde{\bm{\mu}}_t \\
        \mathring{\bm{\mu}}_{t} \\ 
        \mathring{\bm{\sigma}}_{t} 
    \end{bmatrix} \quad \text{and output denoted} \quad g_{o}^{\bm{G_{cl}}}(\bm{\ourbar{z}}_t) = \begin{bmatrix}
        \tilde{\bm{x}}_{t+1} \\
        \tilde{\bm{\mu}}_{t+1} \\
        \mathring{\bm{\mu}}_{t+1} \\
        \mathring{\bm{\sigma}}_{t+1}
    \end{bmatrix},
\end{equation}
where $\tilde{\bm{x}}_t, \tilde{\bm{\mu}}_t, \mathring{\bm{\mu}}_{t}, \mathring{\bm{\sigma}}_{t} \in \mathbb{R}^{6}$.
Note that while the inputs $\tilde{\bm{x}}_t = \hat{\bm{x}}_t$ and $\tilde{\bm{\mu}}_t = \hat{\bm{w}}_{t}$ as computed via the \ac{MHE}, the inputs $\mathring{\bm{\mu}}_{t}$ and $\mathring{\bm{\sigma}}_{t}$ are internal to the reachability analysis and are used to account for the time variation of $\bm{\mu}_t$ and $\bm{W}_t$ as described by \cref{eqn:noise_assumption}.
$\bm{G_{cl}}$ is then manually constructed with the equations
\begin{align} \label{eqn:graph_update_law:f}
    \tilde{\bm{x}}_{t+1} &= f_{cl}(\tilde{\bm{x}}_t; \pi) + \tilde{\bm{\mu}}_{t} \\ \label{eqn:graph_update_law:mu}
    \tilde{\bm{\mu}}_{t+1} &= \tilde{\bm{\mu}}_{t} + \mathring{\bm{\mu}}_{t} + \gamma \mathring{\bm{\sigma}}_{t}\\
    \label{eqn:graph_update_law:delta_mu}
    \mathring{\bm{\mu}}_{t+1} &= \mathring{\bm{\mu}}_{t} \\
    \label{eqn:graph_update_law:delta_sigma}
    \mathring{\bm{\sigma}}_{t+1} &= \mathring{\bm{\sigma}}_{t},
\end{align}
where $\gamma > 0$ is a parameter used to \textit{concretize} an uncertainty bound for a selected confidence interval, e.g., $\gamma=3$ means we assume all samples fall within three standard deviations of the mean.
Concretization is done because {\tt jax\_verify} (and many other analysis tools, such as~\cite{xu2020automatic}) assumes concrete bounds on the possible input states.

\subsubsection{Algorithm Outline}
Having constructed $\bm{G_{cl}}$ and established how we use the \ac{MHE}, we can now explicitly specify our approach, which is summarized pictorially in Figure \ref{fig:approach:approach_diagram} and via pseudo-code in Algorithm \ref{alg:safety_cert}.
\noindent\begin{minipage}{0.5\textwidth}
    \begin{figure}[H]
        \centering
        \includegraphics[width=0.9\linewidth]{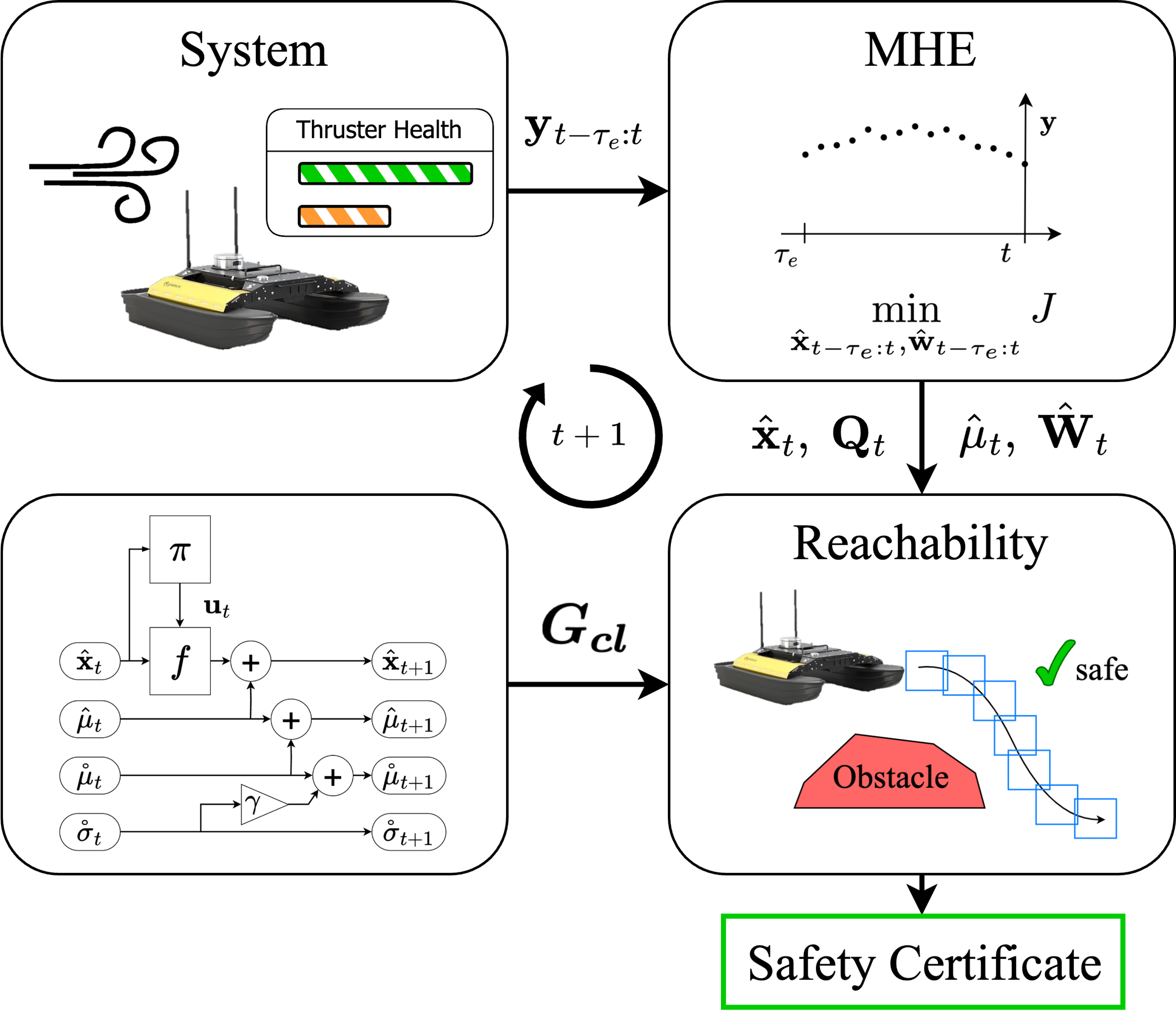}
        \caption{Block diagram depicting our approach \cite{rober2023online}. 
        Data is collected from the system and fed into the MHE, which estimates the state and disturbance terms. 
        The outputs of the MHE are used with a CG representation of the closed-loop dynamics to conduct reachability analysis and certify safety.}
        \label{fig:approach:approach_diagram}
        \vspace{-15pt}
    \end{figure}
\end{minipage}%
%\hfill
\hspace{1mm}
\begin{minipage}{0.48\textwidth}
    \begin{algorithm}[H]
    \caption{Online Safety Certification}
    \begin{algorithmic}[1]
    \setcounter{ALC@unique}{0}
    \renewcommand{\algorithmicrequire}{\textbf{Input:}}
    \renewcommand{\algorithmicensure}{\textbf{Output:}}
    \REQUIRE computational graph $\bm{G_{cl}}$, reachability horizon $\tau_r$, vehicle data $\bm{y}_{t-\tau_e:t}$, concretization parameter $\gamma$, unsafe region $\mathcal{C}$
    \ENSURE safety certificate $c$ over horizon $\tau_r$
        \STATE $c \leftarrow \mathrm{true}$
        \STATE $\hat{\bm{x}}_{t-\tau_e:t}, \hat{\bm{w}}_{t-\tau_e:t} \leftarrow \mathrm{MHE}(\bm{y}_{t-\tau_e:t})$ \label{alg:safety_cert:mhe}
        \STATE $\bm{\mu}_t \leftarrow {\tt mean}(\hat{\bm{x}}_{t-\tau_e:t})$ \label{alg:safety_cert:mean}
        \STATE $\bm{W}_t \leftarrow {\tt cov}(\hat{\bm{w}}_{t-\tau_e:t})$  \label{alg:safety_cert:cov}
        \STATE $\bm{\epsilon_x} \leftarrow {\tt concretize}(\mathrm{MHE}.\bm{Q}_t, \gamma)$  \label{alg:safety_cert:eps_x}
        \STATE $\bm{\epsilon_\mu} \leftarrow {\tt concretize}(\bm{W}_t, \gamma)$  \label{alg:safety_cert:eps_mu}
        \STATE $\bm{\epsilon} \leftarrow [\bm{\epsilon_x}^\top, \bm{\epsilon_\mu}^\top, \Delta\bm{\mu}^\top, \gamma \Delta\bm{\sigma}^\top]^\top$ \label{alg:safety_cert:eps}
        \STATE $\bm{z}_t \leftarrow [\hat{\bm{x}}_t^\top, \hat{\bm{\mu}}_t^\top, \bm{0}_6^\top, \bm{0}_6^\top]^\top$  \label{alg:safety_cert:z}
        \STATE $\ourbar{\mathcal{R}}_t' \leftarrow \mathcal{B}(\bm{z}_t, \bm{\epsilon})$  \label{alg:safety_cert:reach_init}
        \FOR{i in $\{t+1,\ \ldots t+\tau_r\}$} \label{alg:safety_cert:loop}
            \STATE $\ourbar{\mathcal{R}}_i' \leftarrow {\tt jax\_verify}(\bm{G_{cl}},\ \ourbar{\mathcal{R}}_{i-1}')$  \label{alg:safety_cert:reach_update}
            \STATE $\ourbar{\mathcal{R}}_i \leftarrow {\tt projection}(\ourbar{\mathcal{R}}_i')$ \label{alg:safety_cert:projection}
            \IF{$\ourbar{\mathcal{R}}_i \bigcap \mathcal{C} \neq \emptyset$}  \label{alg:safety_cert:safety_check}
                \STATE $c \leftarrow \mathrm{false}$ 
            \ENDIF
        \ENDFOR
    \RETURN $c$
    \end{algorithmic}\label{alg:safety_cert}
    \end{algorithm}
\end{minipage}%

At each time step, $\bm{y}_{t-\tau_e:t}$ is measured by the on-board sensors and passed to the \ac{MHE}, which determines optimal values for $\hat{\bm{x}}_{t-\tau_e:t}$ and $\hat{\bm{w}}_{t-\tau_e:t}$.
Using the output from the \ac{MHE}, we determine estimates for the state $\hat{\bm{x}}_t$ and its covariance $\bm{Q}_{t|t-1}$, as well as disturbance parameter estimates $\hat{\bm{\mu}}_t$ and $\hat{\bm{W}}_t$.
On lines \ref{alg:safety_cert:eps_x} and  \ref{alg:safety_cert:eps_mu} we then obtain concrete uncertainty bounds $\bm{\epsilon}$ by truncating the normal distribution according to concretization parameter $\gamma$.
Next we construct the set of possible inputs for $\bm{G_{cl}}$.
On lines \ref{alg:safety_cert:eps}, \ref{alg:safety_cert:z}, and \ref{alg:safety_cert:reach_init} the possible states $\mathcal{B}(\hat{\bm{x}}_t, \bm{\epsilon_x})$, noise values $\mathcal{B}(\hat{\bm{\mu}}_t, \bm{\epsilon_\mu})$, and reachability variables $\mathcal{B}({\bm{0}}_6, [\Delta\bm{\mu}^\top,\Delta\bm{\sigma}^\top]^\top)$, are concatenated to get the initial set $\ourbar{\mathcal{R}}_t'$ necessary for reachability analysis.
Next, on lines \ref{alg:safety_cert:loop} and \ref{alg:safety_cert:reach_update} we loop over the horizon $\tau_r$, calculating \acp{RSOA} for each time step.
Notice that the \ac{RSOA} is the set of possible states \textit{and} disturbance terms, $\ourbar{\mathcal{R}}_t' \subset \mathbb{R}^{4n_x}$.
Thus, on line \ref{alg:safety_cert:projection}, we project $\ourbar{\mathcal{R}}_t'$ onto $\mathbb{R}^{n_x}$, thereby enabling us to check the safety condition on \ref{alg:safety_cert:safety_check}, which is the desired result.

\subsubsection{Formal Analysis}
% As introduced in \cite{rober2023online},
\cref{thm:ours} formally states the results of our approach. The full proof of the theorem is outlined in \cite{rober2023online}. For continuity with the presentation in this paper we use a system state $\bm{x}_t \in \mathbb{R}^{6}$, but the same results can be extended to higher order systems. 
\begin{theorem}[Safety Verification for an Uncertain System] \label{thm:ours}
Consider a system \cref{eq:nonlinear_cl_dyn} subject to a disturbance $\bm{w}_t \sim \mathcal{N}(\bm{\mu}_t, \bm{W}_t)$ truncated at $\gamma$ standard deviations, where $\bm{\mu}_t$ and $\bm{W}_t$ satisfy the assumptions specified by \cref{eqn:noise_assumption} and where $\hat{\bm{\mu}}_t$ and $\hat{\bm{W}}_t$ are accurate estimates for their respective parameters at the current time step.
The iterative application of \cref{thm:lirpa} with $\bm{G_{cl}}$ defined by \cref{eqn:graph_update_law:f,eqn:graph_update_law:mu,eqn:graph_update_law:delta_mu,eqn:graph_update_law:delta_sigma} and where $\mathcal{I} = \mathcal{B_{\infty}}(\bm{z}_t, \bm{\epsilon})$, $\bm{\epsilon} = [\bm{\epsilon_x}^\top, \bm{\epsilon_\mu}^\top, \Delta\bm{\mu}^\top, \gamma\Delta\bm{\sigma}^\top]^\top$ and $\bm{z}_t = [\hat{\bm{x}}_t^\top, 
\hat{\bm{\mu}}_t^\top, \bm{0}_6^\top, \bm{0}_6^\top]^\top$ provides bounds on all possible $\hat{\bm{x}}_{t:t+\tau_r}$, i.e., $\ourbar{\mathcal{R}}_{t:t+\tau_r}$.

\end{theorem}

\section{Implementation Details}\label{sec:implementation}
\subsection{Speed Controller Details}
The linearized speed dynamics, which were experimentally obtained by offline system identification, are 
\begin{align}
\dot{u}_{sog} & =-24 u_{sog} + 0.618\Lambda_{1}\left(u_{thr}+ \Delta_{u_{sog}}\left(u_{sog}\right)\right),
\end{align}

where $\Delta_{u_{sog}}\left(u_{sog}\right):\mathbb{R}\rightarrow\mathbb{R}$ represents an unknown parametric matched uncertainty. The parameters for baseline controller design were
\begin{align}
    Q_{u_{sog}} = \begin{bmatrix}
        150 & 0\\
        0 & 1
        \end{bmatrix},  \quad \text{and} \quad 
        R_{u_{sog}} = 2
\end{align}
The controller was implemented at 10 Hz. The resulting PI gains were
\begin{align}
k_{u,p} =0.36556 \quad \text{and} \quad  k_{u,i} =8.6603.
\end{align}
The discrete time controller and integrator are
\begin{align}
u_{thr}\left(t\right) & =k_{u,p}u\left(t\right)+k_{u,i}e_{uI}\left(t\right),\label{eq:thrustPIControllerDisc}\\
e_{uI}\left(t+1\right) & =e_{uI}\left(t\right)+\left(u\left(t\right)-u_{d}\right) \Delta t.\label{eq:integralTermSpeed}
\end{align}
The integral component in (\ref{eq:integralTermSpeed}) is bounded using the anti-windup logic in \cite[Sec.3.5]{Astrom2006} with $k_{u,aw}=0.1407$. 
The regression vector $\bm{\Phi}(u_{sog})$ in the adaptive controller was implemented with a bias term and \acp{RBF} \cite[Eq. 12.6]{Lavretsky.Wise2012}. Specifically, 11 \acp{RBF} were placed in the speed dimension centered at speed values of $\left\{ 0.5,\,0.6,\,...,\,1.4,\,1.5\right\} $ meters per second. The adaptive update laws were implemented with linear deadzone operators \cite[Eq. 11.19]{Lavretsky.Wise2012} and a rectangular projection operator \cite[Pg. 352]{Lavretsky.Wise2012}.
\subsection{Yaw-Rate Controller Details} \label{sec:implementation:yaw_rate_controller}

The linearized sway-yaw rate dynamics, which were experimentally obtained by offline system identification, are 

\begin{align}
\left[\begin{array}{c}
\dot{v}\\
\dot{r}
\end{array}\right] & =\left[\begin{array}{cc}
-0.023 & -0.0075\\
0 & -61
\end{array}\right]\left[\begin{array}{c}
v\\
r
\end{array}\right]+\left[\begin{array}{c}
-0.0009\\
0.90
\end{array}\right]\Lambda_{p_2}\left(u_{rud}+\Delta_{r}\left(r\right)\right),
\end{align}

where $\Delta_{r}\left(r\right):\mathbb{R}\rightarrow\mathbb{R}$ represents an unknown parametric matched uncertainty. When estimating the system dynamics we found the entries in $A_p$ and $B_p$ the correspond to the sway dynamics to be small in magnitude but non-zero, even though the system is underactuated.  Their inclusion does not directly impact the yaw-rate dynamics, and are reported for completeness. For yaw-rate tracking the, the LQR parameters are
\begin{align}
Q_{r} =\left[\begin{array}{ccc}
100 & 0 & 0\\
0 & 10 & 0\\
0 & 0 & 10
\end{array}\right], \quad \text{and} \quad R_{r} =0.01.
\end{align}
The controller was implemented at 10 Hz. The resulting PI gains were
\begin{align}
k_{v,p} =-0.33, \quad 
k_{r,p} =8.44, \quad \text{and} \quad
k_{r,i} =100.
\end{align}
The discrete time controller and integrator are
\begin{align}
u_{rud}\left(t\right) & =k_{v,p}v\left(t\right)+k_{r,p}r\left(t\right)+k_{r,i}e_{rI}\left(t\right)\label{eq:thrustPIControllerDisc-2}\\
e_{rI}\left(t+1\right) & =e_{rI}\left(t\right)+\left({r\left(t\right)-r_{d}}\right) \Delta t.\label{eq:integralTermSpeed-1}
\end{align}
The integral component in (\ref{eq:integralTermSpeed-1}) is bounded using the anti-windup logic in \cite[Sec.3.5]{Astrom2006}  with $k_{r,aw}=1.1667$.
Similar to the speed controller, the regression vector $\Phi$ in the adaptive controller was implemented with a bias term and \acp{RBF} \cite[Eq. 12.6]{Lavretsky.Wise2012}. Specifically, 21 \acp{RBF} were placed in the yaw rate dimension centered at values of $\left\{ -20,-18,...,18,20\right\} $ degrees per second. 

Finally, we report the constants used for the heading to yaw-rate controller which filters the output from pHelmIvP as described in Section \ref{sec:helmfilter}. During experiments we used $\tau_u = 1s$, $\tau_r = 0.1$, and $\tau_\theta = 0.2$. During the \ac{UNREP} experiment we decreased the size of domain intervals to $0.01 \frac{m}{s}$ for speed and $0.1\frac{rad}{s}$ for heading.

\subsection{Thruster Allocation}
The output from the speed and heading controllers, $u_{thr}$ and $u_{rud}$, is allocated to the left and right thruster of the Heron \ac{USV} with the following thruster mixing equations: 
\begin{align}
u_{L} &= u_{thr} * \left(1 + \frac{u_{rud}}{R_{max}}\right) \\
u_{R} &= u_{thr} * \left(1 - \frac{u_{rud}}{R_{max}}\right)
\end{align}
$R_{max}$ is the maximum allowable rudder and was set at $R_{max} = 50$ for all field experiments.

\subsection{Payload Autonomy}
The system architecture follows the backseat driver design philosophy \cite{Benjamin2010MOOSIvP} with the following modification:  All the control processes ran on the payload autonomy computer which then sent the control inputs $u_L$ and $u_R$ to the front seat.  A system overview is provided in Figure \ref{fig:falcon_link_system}.
\noindent\begin{minipage}{0.52\textwidth}
    \begin{figure}[H]
        \centering
        \includegraphics[width=1.00\textwidth]{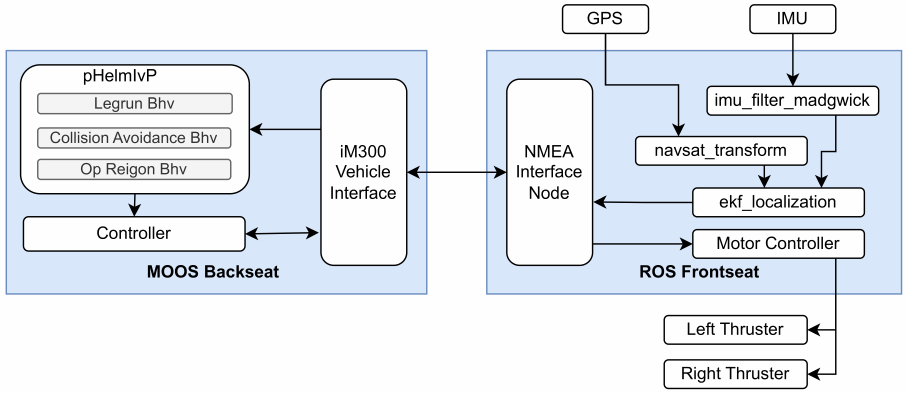}
        \caption{System architecture detailing backseat payload autonomy with commercial Heron frontseat.}
        \label{fig:falcon_link_system}
        \vspace{-12pt}
    \end{figure}
\end{minipage}%
%\hfill
\hspace{8mm}
\begin{minipage}{0.4\textwidth}
    \begin{table}[H]
        \centering
        \caption{Sensor and Process Rates .}
        \vspace{2mm}
        \begin{tabular}{c | c } \label{tab:sensor_rates}
            \textbf{Sensor or Process} & \textbf{Freq (Hz)}  \\ [0.5ex] 
            \hline
            IMU & 40  \\ 
            \hline
            GPS & 5 \\
            \hline
            robot\_localization ROS Package \\ \cite{MooreStouchKeneralizedEkf2014} & 20 \\
            \hline
            Controller & 20 \\
            \hline
            HelmIvP & 4 (10${}^*$)  \\
            \hline
            Reachability Calculation & 10 \\
            \hline
        \end{tabular}\\
        \vspace{2mm}
        \small{*Rate was increased during UNREP experiments}
    \end{table}
\end{minipage}%

As for the reachabilty calculation, the same architecture was simplified for speed of calculation.  More specifically, the guidance input was not generated from the pHelmIvP multi-objective behavior optimization process, but instead it was calculated based on a trackline of a single dominant waypoint behavior.  As a result, the reachability analysis is limited in the sense that it does not consider the multiple objectives of waypoint following, collision avoidance, and remaining inside the operational region.   However, we found this simplification to be acceptable given the limited time horizon.

\subsection{Timing}
There were no guarantees of message timing because the system operated on two separate computers with two different middlewares: ROS and MOOS.  Furthermore, the state estimation was shared over WiFi between vehicles during the UNREP experiment which resulted in additional time delays of signals used in the controller.  During preliminary testing we observed propagation of time delays through the system that resulted in poor performance.  These issues motivated a focus on increasing cycle frequency where possible and implementing processes that were robust to inconsistent message timing.  A summary of the cycle frequencies can be found in Table \ref{tab:sensor_rates}.

\section{Configuration of Field Experiments}\label{sec:experimental_setup}
\begin{figure}[ht]
    \centering
    \includegraphics[width=0.8\textwidth]{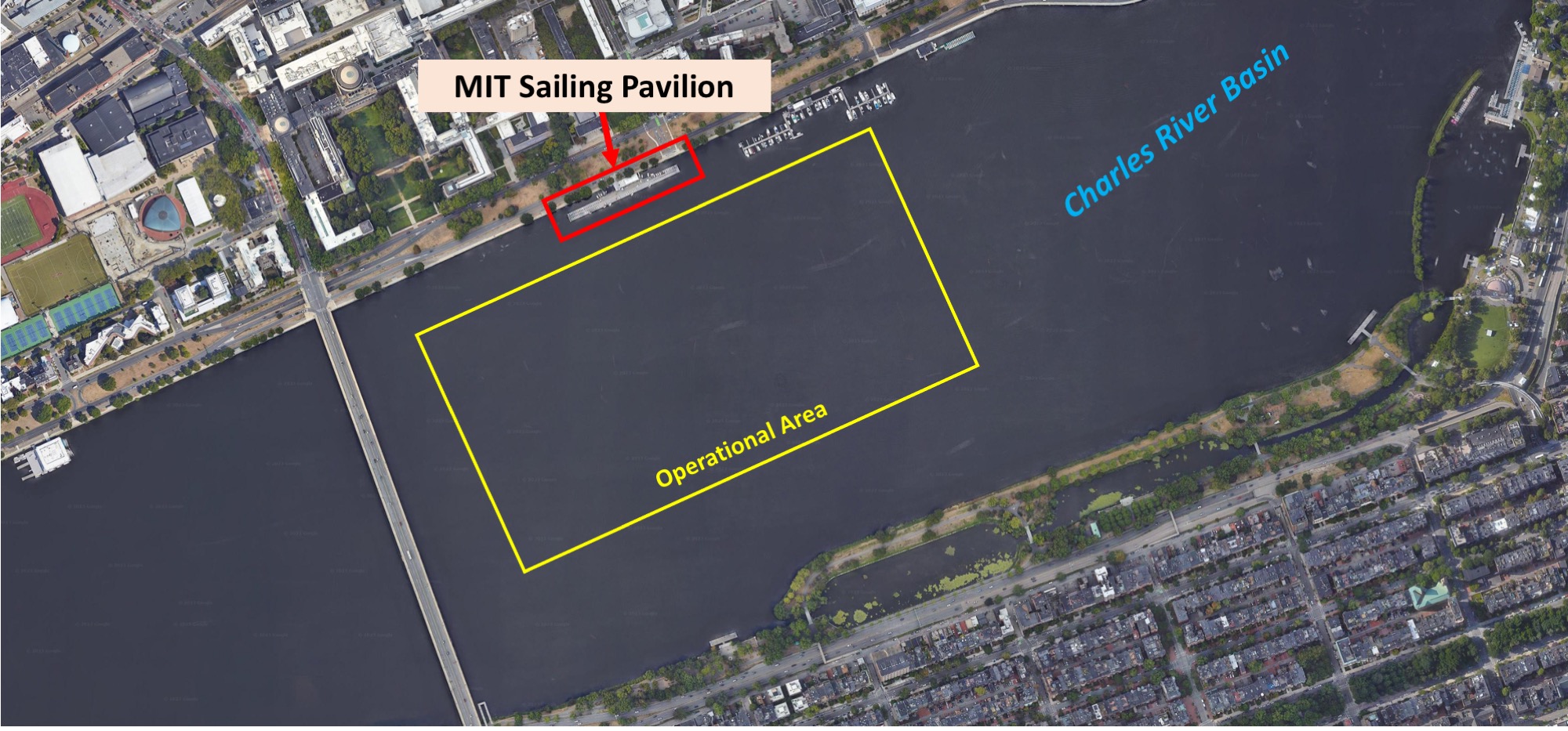}
    \caption{Overview of operational area near the MIT sailing pavilion}
    \label{fig:mit_sp_overview}
    \vspace{-12pt}
\end{figure}

Field experiments were conducted near the MIT Sailing Pavilion in Cambridge MA.  The sailing pavilion provided access to the Charles River Basin, a public water space near the MIT campus.  An overview of the operational area is shown in Figure \ref{fig:mit_sp_overview}.
Development began in Febuary 2023 and continued until a final demonstration on November 19, 2023.  Over 80 missions were conducted.  The control approaches were developed and tested in a variety of weather and surface conditions during the eight month season.  The experimental results reported herein were collected on the final day of testing, November 19th 2023.  The air temperature for the day ranged from -0.6 $^\circ$C to 3 $^\circ$C and the wind speed varied from 2.2 $\sfrac{m}{s}$ to 4.9 $\sfrac{m}{s}$ as recorded by the MIT Sailing pavilion weather station. 

\subsection{Definition of Control Approaches Tested in the LegRun Mission}\label{sec:three_controllers_tested}
In every test we compared the performance of the \ac{MRAC} controller to the existing \ac{PID} baseline controller.  Since the LegRun scenario offered a more structured and repeatable method of testing, we also compared performance of the \ac{MRAC} controller to an \ac{LQR} \ac{PI} controller that did not have the adaptive term.  Specifically the three controller used in testing were:
For each experiment, we deployed one of the following controllers:
\begin{itemize}
    \item A baseline \ac{PID} controller designed for the Heron USV by the manufacturer and members of the MIT PavLab.  This controller has been in service on the aforementioned platform for approximately seven years. 
    \item \ac{LQR} \ac{PI} as described in Section \ref{sec:LQR_PI}
    \item \ac{MRAC} as described in Section \ref{sec:MRAC_def}
\end{itemize}

\subsection{USV Disturbances}\label{sec:disturbances}
Our goal was to thoroughly test the performance of the controllers when the USV was subject to significant disturbances, or disturbances that are greater in magnitude than would be reasonable expected in operation. To that end we developed three disturbances: simulated thruster failure, a sail, and a drogue. 
\subsubsection{Simulated Thruster Failure}
Significant changes to thruster performance of the Heron USV can occur for many reasons including a blocked inlet duct, a line wrapped around the propeller, or a blocked outlet duct.  Since these real disturbances can cause permanent damage to the USV, the effect of thruster failure was simulated in software.  The simulated failures were parameterized by multiplicative and additive bias terms that transform the intended commanded input into a ``faulty'' input.  

Given the definition of these thruster failure modes:
\begin{itemize}
    \item $\digamma_{\!\!rf}$ is the rudder fault multiplicative factor where $\digamma_{\!\!rb} = 1$ is no disturbance
    \item $\digamma_{\!\!rb}$ is the rudder fault additive bias
    \item $\digamma_{\!\!tf}$ is the thruster fault multiplicative factor where $\digamma_{\!\!tf} = 1 $ is no disturbance
    \item $\digamma_{\!\!tb}$ is the thruster fault additive bias 
\end{itemize}
the calculation of the ``faulty'' thruster commands is expressed as 
\begin{align}
    \digamma_{\!\!{{rb}_L}} = & \frac{ \digamma_{\!\!rf} - 1}{2}  \cdot (u_{L} - u_{R})\\
    \digamma_{\!\!rb_R} = & \frac{ \digamma_{\!\!rf} - 1}{2} \cdot (u_{R} - u_{L})\\
    \Tilde{u}_{L} = &  u_{L} \cdot \digamma_{\!\!tf_L} + \digamma_{\!\!tb_L} + \digamma_{\!\!rb_L}   \label{eq:sim_fault_cmd_L}\\
    \Tilde{u}_{R} = &  u_{R} \cdot \digamma_{\!\!tf_R} + \digamma_{\!\!tb_R} + \digamma_{\!\!rb_R}, \label{eq:sim_fault_cmd_R}
\end{align}
where $u_{L}$ and $u_{R}$ are the intended inputs. The faulty commands $\Tilde{u}_{L}$ and $\Tilde{u}_{R}$ are sent to the front-seat motor controller instead of the intended commands.

\subsubsection{Sail}
\begin{figure*}[ht]
    \centering
    \begin{subfigure}{0.5\textwidth}
        \centering
        \includegraphics[height=1.7in]{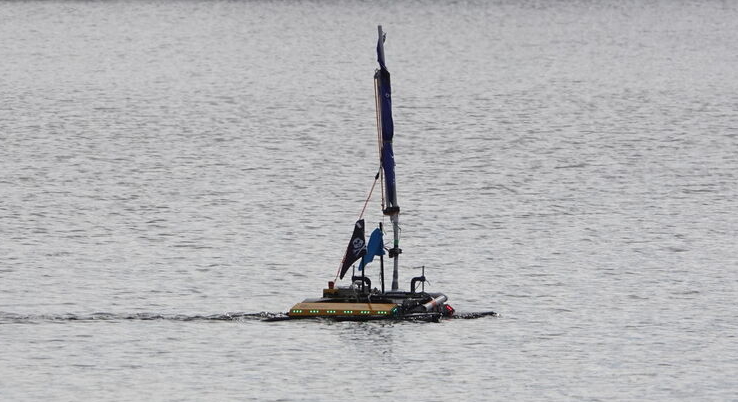}
        \caption{The sail secured in a stowed position at the start of the mission.}
         \label{fig:heron_sail_secured}
    \end{subfigure}%
    ~ 
    \begin{subfigure}{0.5\textwidth}
        \centering
        \includegraphics[height=1.7in]{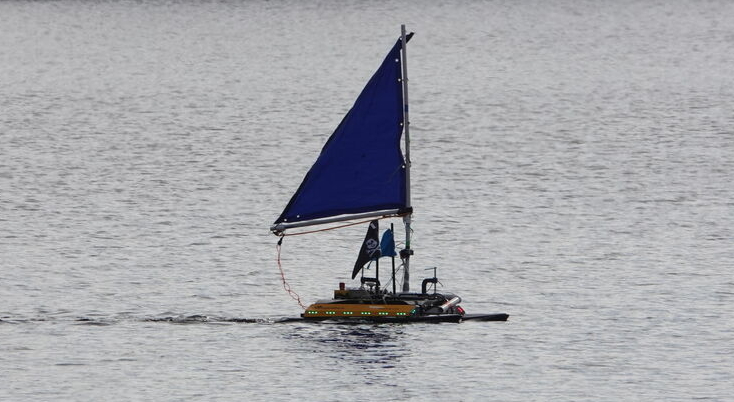}
        \caption{The sail was deployed remotely during the mission.}
        \label{fig:heron_sail_deployed}
    \end{subfigure}%
    \caption{Remotely deployed sail disturbance.}
    \vspace{-3mm}
\end{figure*}

The sail deployed remotely when given a command from either a shore side operator, or at a pre-programmed location along the mission.  The sail deployment is shown in Figure \ref{fig:heron_sail_secured} and Figure \ref{fig:heron_sail_deployed}.    The sheet attached to the free end of the boom was connected to the back end of the Heron and remained constant length.  As a result the boom traveled to a position that was almost perpendicular to the body center-line when the sail was engaged.  The sail was mounted approximately halfway back on the port side and was free to rotate to either side in response to wind.  As a result of the offset positioning, the sail could induce greater moments in the positive $Z$ direction. 
\subsubsection{Drogue}
\begin{figure*}[ht]
    \centering
    \begin{subfigure}{0.35\textwidth}
        \centering
        \includegraphics[height=1.8in]{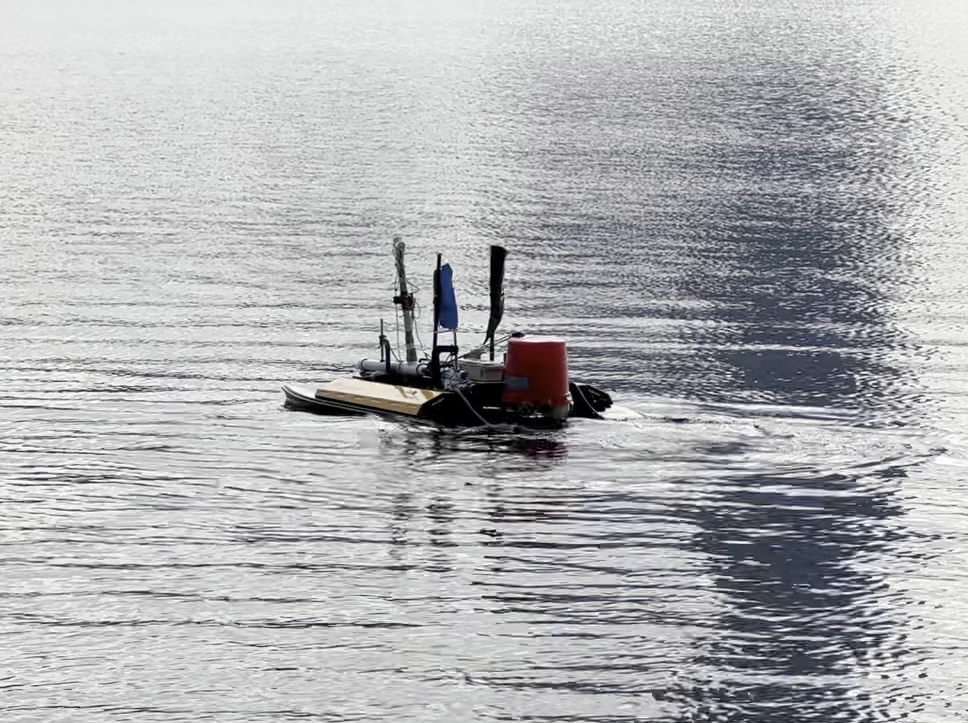}
        \caption{Drogue secured on the rear of the Heron at the start of the mission.}
         \label{fig:heron_drogue_secured}
    \end{subfigure}%
    ~  \hspace{5mm}
    \begin{subfigure}{0.5\textwidth}
        \centering
        \includegraphics[height=1.8in]{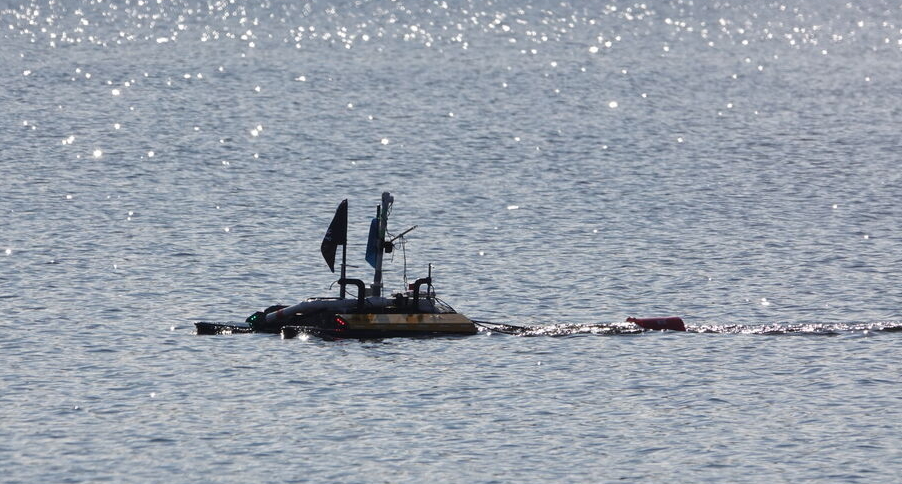}
        \caption{Drogue was deployed remotely during the mission and pulled through the water behind the USV.}
        \label{fig:heron_drogue_deployed}
    \end{subfigure}%
    \caption{Remotely deployed drogue (bucket) disturbance.}
    \vspace{-3mm}   
\end{figure*}

Similar to the sail, the drogue deployed remotely when given a command from either a shore side operator, or at a pre-programmed location along the mission.  A five gallon bucket served as the main component of the drogue and was attached to the body of the Heron with two ropes.   The bucket was mounted to the back of the Heron, and when the deployment command was received the bucket passively fell off the body into the water behind the vehicle.  The deployment process is shown in Figure \ref{fig:heron_drogue_secured} and Figure \ref{fig:heron_drogue_deployed}. 
\subsection{LegRun Evaluation Mission}
\begin{figure}
    \centering
    \includegraphics[width=0.8\textwidth]{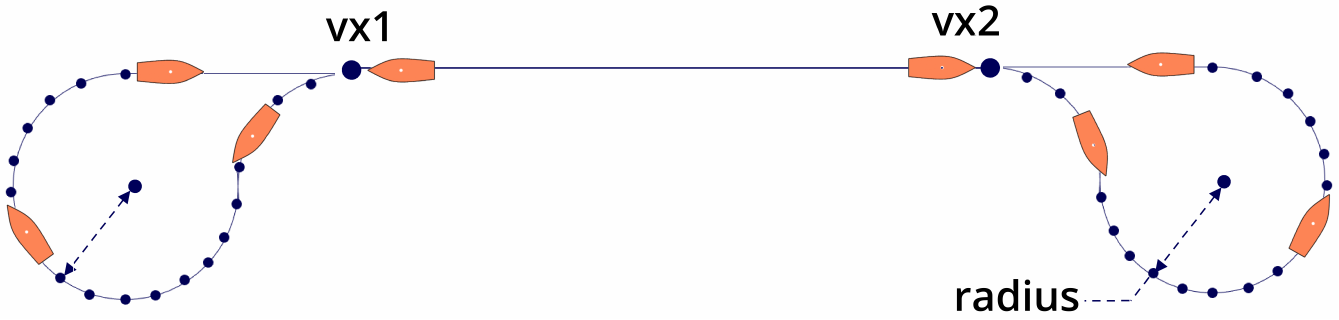}
    \caption{Overview of LegRun behavior used in experimental testing.  Each LegRun path is parameterized by a radius and two waypoints, vx1 and vx2.}
    \label{fig:legrun_overview}
    \vspace{-1mm}
\end{figure}

The LegRun behavior is designed to facilitate the testing of a marine robotic platform by enabling the vehicle to traverse back and forth over a ``legrun'' trackline prescribed by two vertices. After reaching the end of the trackline, the vehicle switches into a Williamson turn mode to ensure vehicle alignment with the trackline when starting in the other direction. An overview of the intended action is shown in Figure \ref{fig:legrun_overview}.  

The LegRun missions presented in this work starts with a sequence of two Williamson turns each of radius 20m followed by two tighter turns with radius 10m. 
In several ways the LegRun behavior is ideal for testing marine vehicles.  By alternating direction of travel along the middle leg sections we ensure that the vehicle is subject to wind and wave forces from opposite directions.  The behavior includes Williamson turns in both the left and right directions.  As a practical matter, the behavior publishes a status of progress along the track and that was used to consistently trigger actuation of disturbances.

The LegRun behavior generates a sequence of waypoints that parameterize the entire path. At each iteration of pHelmIvP the behavior produces an objective function that maps the set of possible combinations of actions for speed and heading to utility.   When multiple behaviors are active, the IvP optimizer in pHelmIvP determines the set of actions that maximize the sum of the utility functions of all active behaviors. More details can be found in \cite{Benjamin2010MOOSIvP}.

During this portion of the field testing the LegRun behavior was almost always the sole active behavior, since the main objective was trackline following and there were no other vehicles to consider.  Other behaviors usually remained idle, such as those related to safety, e.g. avoidColregs behavior, which implements collision avoidance in compliance with the naval COLREGS, and the opRegionBehavior which implements a containment region for the vehicle operations \cite{IvPManPages}. As a result the pHelmIvp generated a desired reference that reflected the desired action selection from the LegRun behavior. 

\subsection{Straight Trackline Mission}\label{sec:straight_trak_miss}
In these missions the desired behavior of the Heron USV was to follow a relatively long, straight trackline.   The intent of this mission was to validate the change in forward reachable sets as the USV transitions from a steady-state normal operating condition to a faulty condition.  The straight line test offers a method to clearly evaluate the performance of the thruster bias estimation because it eliminates some of the complications found in the more involved LegRun mission, such as integrator wind-up in the alternating turns.

\section{Experimental Results - Systematic Evaluation of Control and Forward Reachability Analysis}\label{sec:experimental_results_sys_eval}
This section reports the performance of the adaptive control and real-time forward reachability analysis module in a set of systematic field experiments.  The configuration details of these experiments is described in Section \ref{sec:experimental_setup}. 
Specifically, we evaluated the performance of the \ac{MRAC} controller against the \ac{LQR} \ac{PI} and \ac{PID} controllers in the LegRun mission scenario.  During several straight trackline missions the forward reachable sets were calculated in real-time throughout the mission, and this demonstrated the effectiveness of the bias estimation to capture the effect of significant disturbances on the reachable states. 
Discussion of the results is included at the end of the section.

\subsection{LegRun Missions for Evaluation of Control Performance}
In this section we report a sequence of tests that began with the USV completing the LegRun mission encumbered by only the naturally occurring disturbances from wind and waves.  After a baseline was established, we repeated the same mission three times, each with a different significant disturbance as described in Section \ref{sec:disturbances}.

\subsubsection{Baseline LegRun Mission}
\begin{figure}[H]
    \centering
    \begin{subfigure}[t]{0.5\textwidth}
        \centering
        \includegraphics[width=1.0\textwidth]{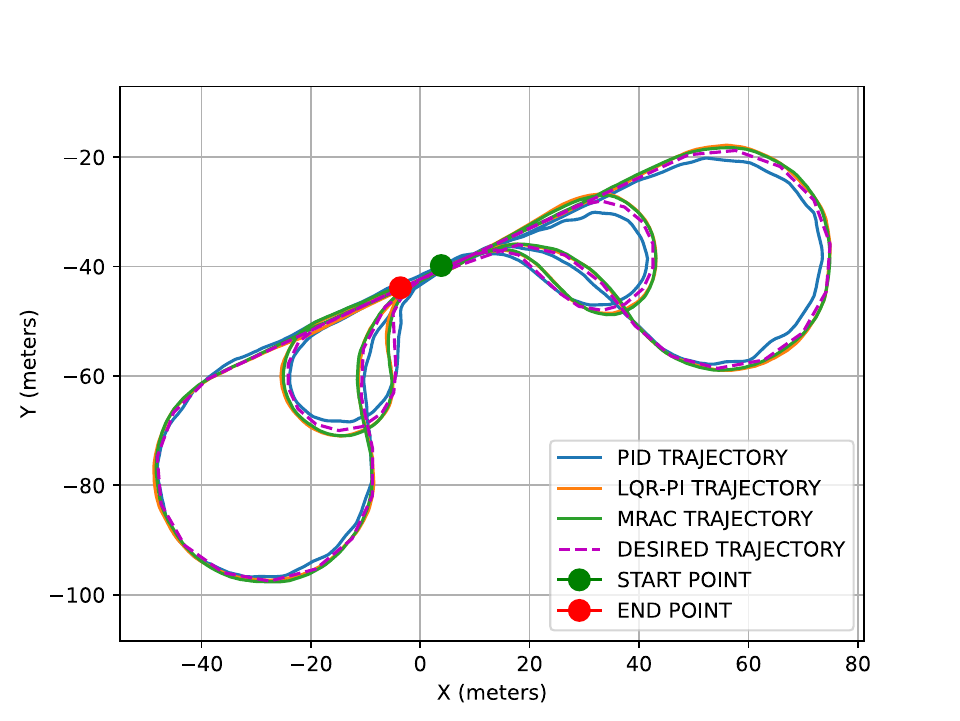}
        \caption{USV Trajectories during the baseline Legrun experiment.   }
        \label{fig:baseline_combined_traj}
    \end{subfigure}%
    ~  \hspace{0mm}
    \begin{subfigure}[t]{0.5\textwidth}
        \centering
        \includegraphics[width=1.0\textwidth]{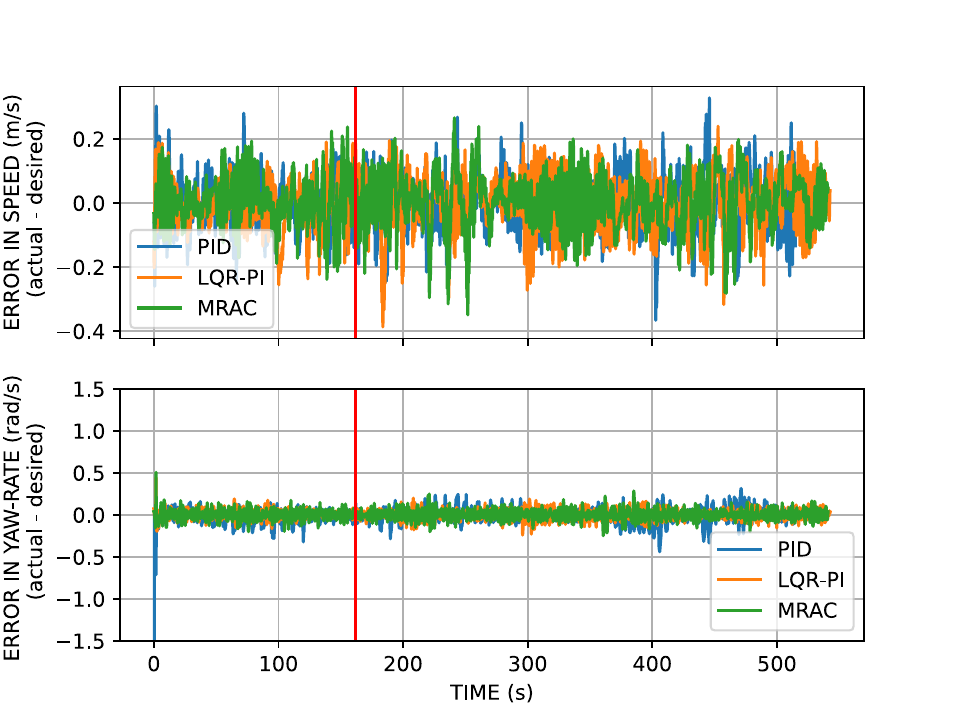}
        \caption{State Errors}
        \label{fig:baseline_errors}
    \end{subfigure}
    \caption{\textbf{Baseline} LegRun mission results.  The Heron USV used the baseline track-line following algorithm (Section \ref{sec:baseline_trackline_guidance}) with \ac{PID}, \ac{LQR} \ac{PI}, \ac{MRAC} inner loop controllers.}
    \vspace{-3mm}
    \label{fig:baseline_legrun}
\end{figure}

A series of experiments summarized in \cref{fig:baseline_legrun} show the performance of the \ac{PID}, \ac{LQR} \ac{PI}, and \ac{MRAC} controllers using the LegRun behavior with only naturally-occurring disturbances such as wind, waves, and river current.  The trajectory and state error plots are reported in Fig. \ref{fig:baseline_combined_traj} and Fig. \ref{fig:baseline_errors} respectively. Table \ref{table:baseline_stats} reports the speed, heading, yaw rate, and position error for the three controllers during the LegRun experiment.

\begin{table}[H]
    \centering
    \caption{Error Statistics (\ac{RMSE} $\pm$ \ac{RMSD}) for the \textbf{Baseline} LegRun mission.}
    \begin{tabular}[H]{lccc}
        \hline
        &PID&LQR-PI&MRAC\\
        \hline
        Speed Error (m/s)&0.09 $\pm$ 0.06&0.09 $\pm$ 0.06&\textbf{0.08 $\pm$ 0.06}\\
        Heading Error (deg)&\textbf{10.35 $\pm$ 7.34}&17.63 $\pm$ 11.4&17.47 $\pm$ 11.40\\
        Yaw Rate Error (rad/s)&0.09 $\pm$ 0.07&\textbf{0.06 $\pm$ 0.04}&\textbf{0.06 $\pm$ 0.04}\\
        Position Error (m)& 0.94 $\pm$ 0.57&0.47 $\pm$ 0.35&\textbf{0.45 $\pm$ 0.33}\\
        \hline
        \label{table:baseline_stats}
    \end{tabular}
\end{table}

\subsubsection{LegRun Mission with Simulated Fault}

\begin{figure}[htp]
    \centering
    \begin{subfigure}[t]{0.5\textwidth}
        \centering        
        \includegraphics[width=1.0\textwidth]{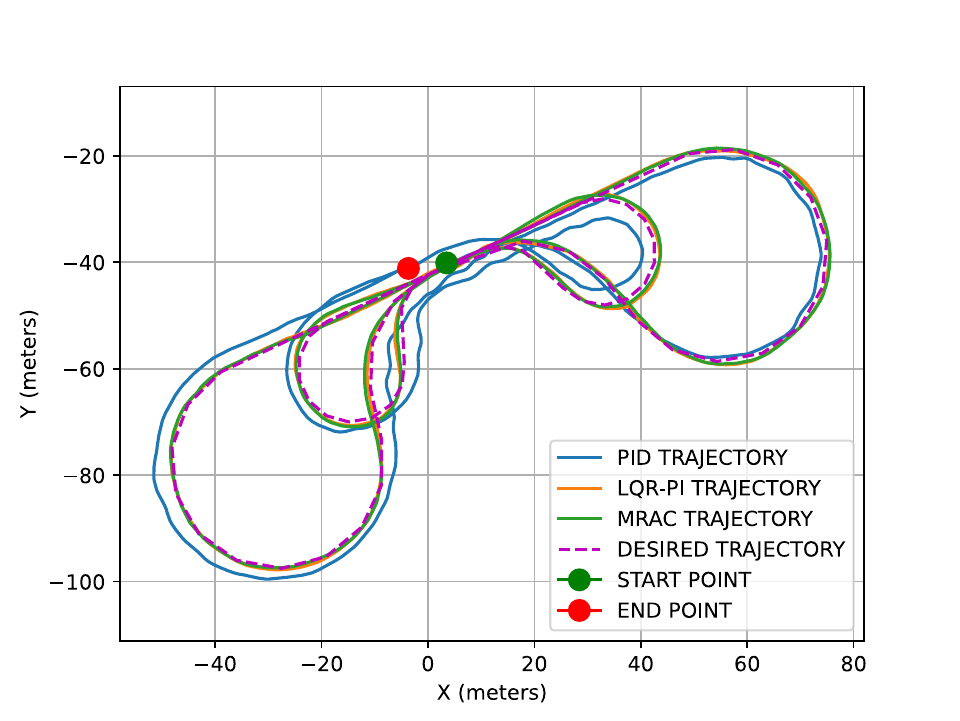}
        \caption{USV Trajectories during the LegRun experiment with simulated 50\% fault on the left thruster.}
        \label{fig:thrust_combined_traj}
    \end{subfigure}%
    ~  \hspace{0mm}
    \begin{subfigure}[t]{0.5\textwidth}
        \centering
        \includegraphics[width=1.0\textwidth]{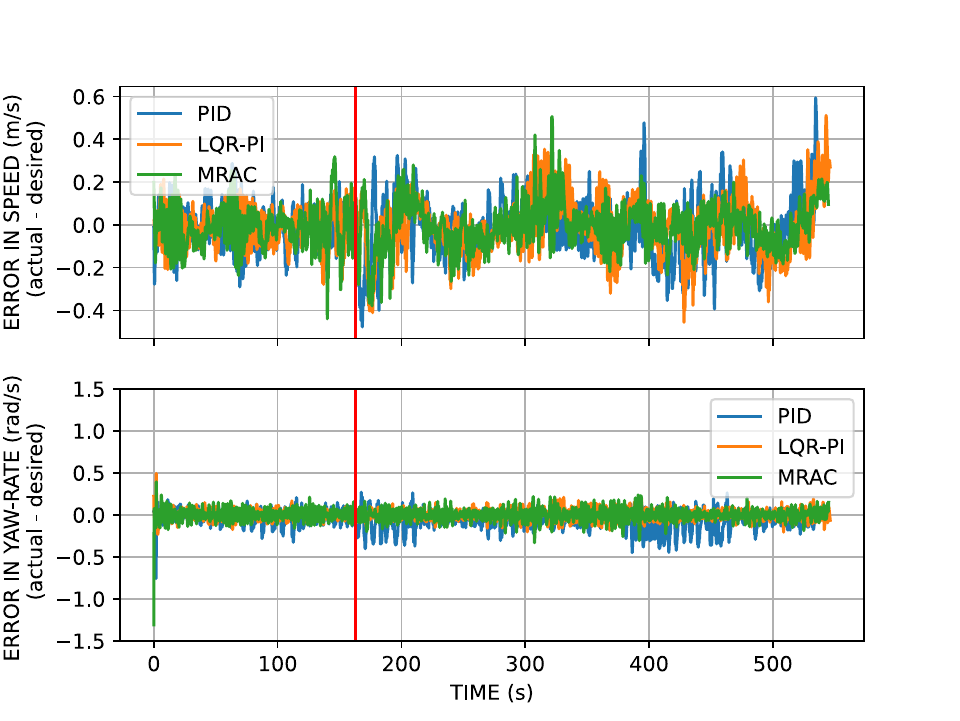}
        \caption{State Errors}
        \label{fig:sim_thr_fault_errors}
    \end{subfigure}%
    \caption{LegRun mission with \textbf{simulated thruster fault}. The Heron USV used the baseline track-line following algorithm (Section \ref{sec:baseline_trackline_guidance}) with \ac{PID}, \ac{LQR}, \ac{MRAC} inner loop controllers.  The simulated fault was induced after the Heron completed the first large right loop of the LegRun.}
    \vspace{-3mm}   
    \label{fig:legrun_thr_fault}
\end{figure}

In the next series of experiments, we tested the \ac{PID}, \ac{LQR} \ac{PI}, and \ac{MRAC} controllers on the LegRun behavior with the natural disturbances and a reduction in the output of the left thruster by 50\% of the commanded value, i.e. $\digamma_{\!\!tf_L} = 0.5$ in (\ref{eq:sim_fault_cmd_L}). The simulated fault was induced after the Heron completed the first large right loop of the LegRun. 
The results of these experiments are shown in \cref{fig:legrun_thr_fault} and \cref{table:sim_thr_stats}.  The trajectory and state error plots are reported in Fig. \ref{fig:thrust_combined_traj} and Fig. \ref{fig:sim_thr_fault_errors} respectively. Table \ref{table:sim_thr_stats} reports the speed, heading, yaw rate, and position error for the three controllers during the LegRun experiment with the simulated fault.

\vspace{5mm}
\begin{table}[H]
    \centering
    \caption{Error Statistics (\ac{RMSE} $\pm$ \ac{RMSD}) for the LegRun mission with \textbf{simulated thruster fault}}
    \begin{tabular}[H]{lccc}
        \hline
        &PID&LQR-PI&MRAC\\
        \hline
        Speed Error (m/s)&0.14 $\pm$ 0.10&0.13 $\pm$ 0.09&\textbf{0.11 $\pm$ 0.07}\\
        Heading Error (deg)&27.73 $\pm$ 11.33&\textbf{17.54 $\pm$ 10.99}&17.63 $\pm$ 11.00\\
        Yaw Rate Error (rad/s)&0.14 $\pm$ 0.10&\textbf{0.06 $\pm$ 0.04}&0.07 $\pm$ 0.05\\
        Position Error (m)&2.54 $\pm$ 0.72&\textbf{0.45 $\pm$ 0.30}&0.49 $\pm$ 0.31\\
        \hline
        \label{table:sim_thr_stats}
    \end{tabular}
\end{table}

\subsubsection{LegRun Mission with Deployed Drogue}
\begin{figure}[htp]
    \centering
    \begin{subfigure}[t]{0.5\textwidth}
        \centering
        \includegraphics[width=1.0\textwidth]{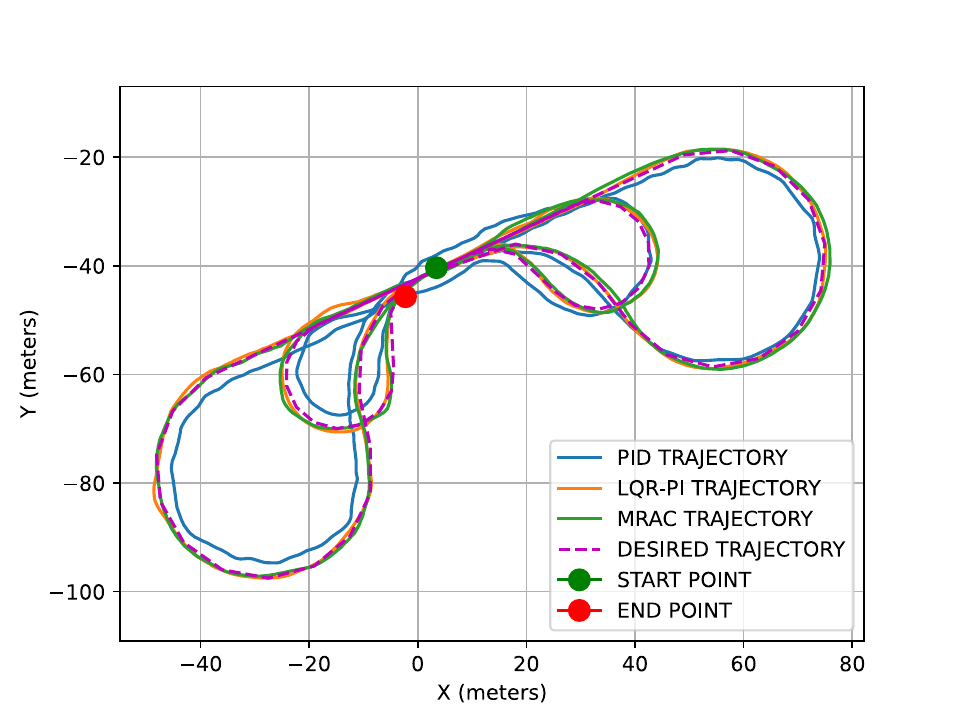}
        \caption{USV Trajectories during the LegRun experiment with drogue deployment.  }
        \label{fig:drogue_combined_traj}
    \end{subfigure}%
    ~  \hspace{0mm}
    \begin{subfigure}[t]{0.5\textwidth}
        \centering
        \includegraphics[width=1.0\textwidth]{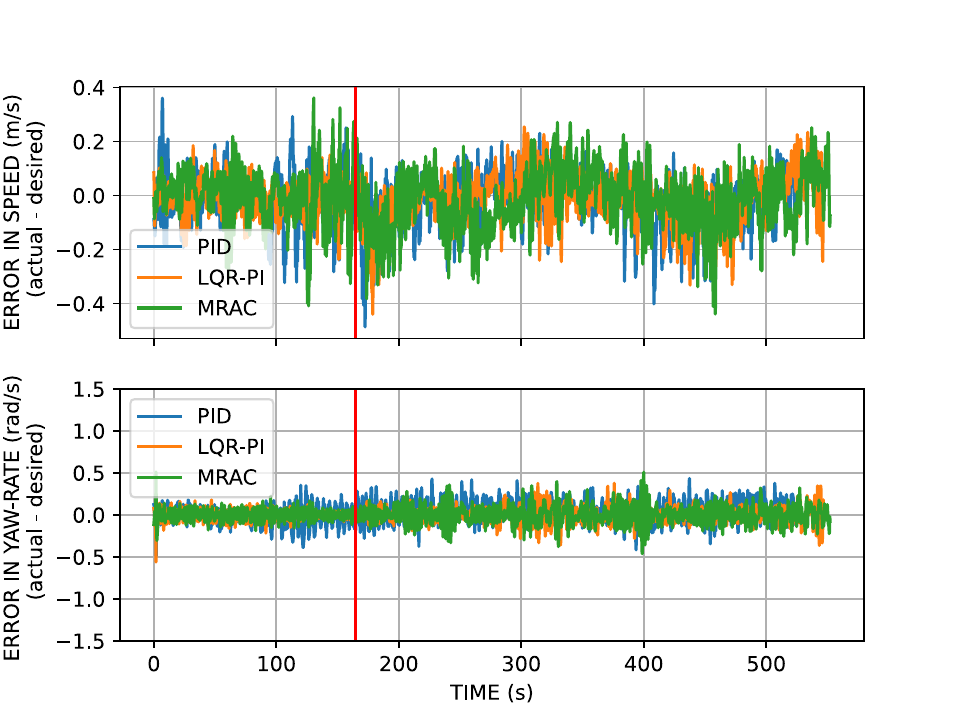}
        \caption{State Errors}
        \label{fig:drogue_errors}
    \end{subfigure}%
    \caption{LegRun mission with \textbf{drogue fault}. The Heron USV used the baseline track-line following algorithm (Section \ref{sec:baseline_trackline_guidance}) with \ac{PID}, \ac{LQR}, \ac{MRAC} inner loop controllers. The drogue was deployed after the Heron completed the first large right loop of the LegRun.}
    \vspace{-3mm}   
    \label{fig:drogue}
\end{figure}

In this series of experiments, we tested the \ac{PID}, \ac{LQR} \ac{PI}, and \ac{MRAC} controllers on the LegRun behavior with the drogue, which was deployed after the Heron completed the first large right loop of the LegRun.
The results of these experiments are shown in \cref{fig:drogue} and \cref{table:drogue_stats}.
The trajectory and state error plots are reported in Fig. \ref{fig:drogue_combined_traj} and Fig. \ref{fig:drogue_errors} respectively. Table \ref{table:drogue_stats} reports the speed, heading, yaw rate, and position error for the three controllers during the LegRun experiment with the drogue deployed.

\vspace{-5mm}
\begin{table}[H]
    \centering
    \caption{Error Statistics (\ac{RMSE} $\pm$ \ac{RMSD}) for the LegRun mission with \textbf{drogue fault}. }
    \begin{tabular}[H]{lccc}
        \hline
        &PID&LQR-PI&MRAC\\
        \hline
        Speed Error (m/s)&0.10 $\pm$ 0.07&\textbf{0.09 $\pm$ 0.07}&0.11 $\pm$ 0.07\\
        Heading Error (deg)&19.02 $\pm$ 9.70&18.01 $\pm$ 11.37&\textbf{17.76 $\pm$ 10.97}\\
        Yaw Rate Error (rad/s)&0.14 $\pm$ 0.08&\textbf{0.09 $\pm$ 0.07}&0.11$\pm$ 0.08\\
        Position Error (m)&1.89 $\pm$ 1.00&0.57$\pm$ 0.57&\textbf{0.43 $\pm$ 0.39}\\
        \hline
        \label{table:drogue_stats}
    \end{tabular}
\end{table}

\subsubsection{LegRun Mission with Deployed Sail}

\begin{figure}[H]
    \centering
    \begin{subfigure}[t]{0.5\textwidth}
        \centering
        \includegraphics[width=1.0\textwidth]{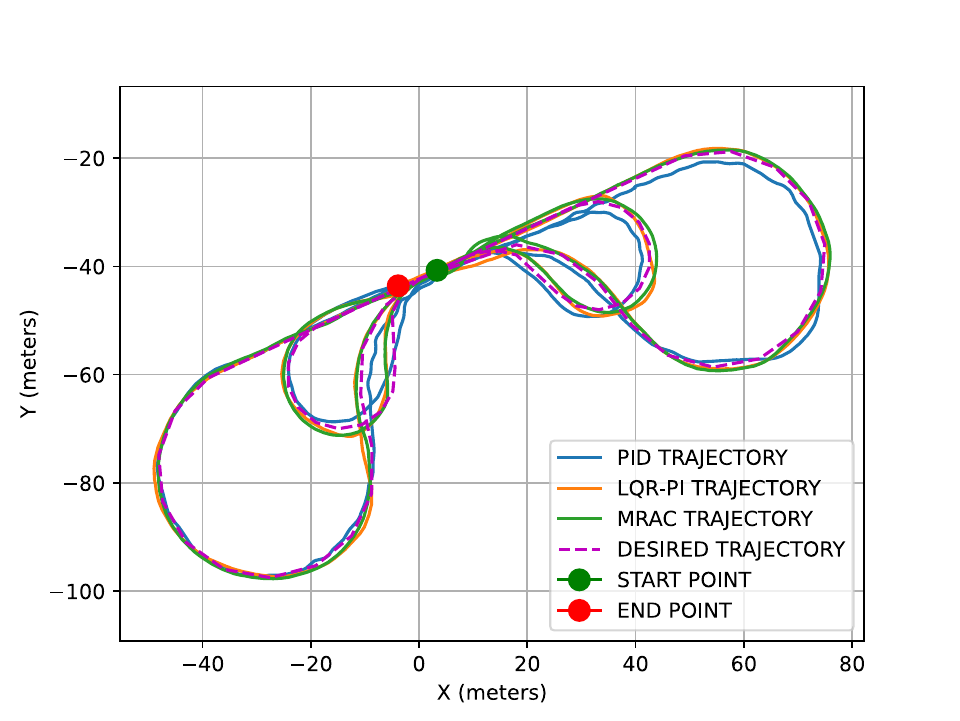}
        \caption{USV Trajectories during the LegRun experiment with deployed sail.  }
        \label{fig:sail_combined_traj}
    \end{subfigure}%
    ~  \hspace{0mm}
    \begin{subfigure}[t]{0.5\textwidth}
        \centering
        \includegraphics[width=1.0\textwidth]{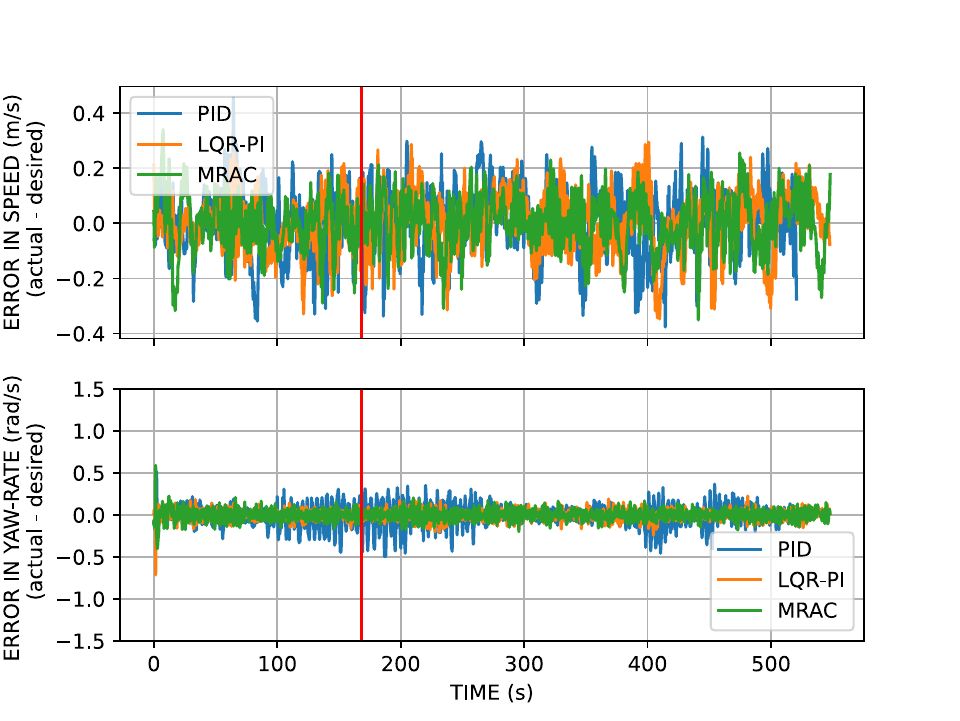}
        \caption{State Errors}
        \label{fig:sail_errors}
    \end{subfigure}%
    \caption{LegRun mission with \textbf{sail fault}. The Heron USV used the baseline track-line following algorithm (Section \ref{sec:baseline_trackline_guidance}) with \ac{PID}, \ac{LQR} \ac{PI}, \ac{MRAC} inner loop controllers. The sail was deployed after the Heron completed the first large right loop of the LegRun.}
    \vspace{-3mm}   
    \label{fig:sail}
\end{figure}

In this series of experiments, we tested the \ac{PID}, \ac{LQR} \ac{PI}, and \ac{MRAC} controllers on the LegRun behavior with the sail.
The results of these experiments are shown in \cref{fig:sail} and \cref{table:sail_stats}. The trajectory and state error plots are reported in Fig. \ref{fig:sail_combined_traj} and Fig. \ref{fig:sail_errors} respectively. Table \ref{table:sail_stats} reports the speed, heading, yaw rate, and position error for the three controllers during the LegRun experiment with the sail deployed.

\begin{table}[H]
    \centering
    \caption{Error Statistics (\ac{RMSE} $\pm$ \ac{RMSD}) for the LegRun mission with \textbf{sail fault}}
    \begin{tabular}[H]{lccc}
        \hline
        &PID&LQR-PI&MRAC\\
        \hline
        Speed Error (m) &0.12 $\pm$ 0.08&\textbf{0.06 $\pm$ 0.04}&0.09 $\pm$ 0.06\\
        Heading Error (deg)&\textbf{13.80 $\pm$ 9.22}&19.09 $\pm$ 12.67&18.04 $\pm$ 10.78\\
        Yaw Rate Error (rad/s)&0.13 $\pm$ 0.10&\textbf{0.06 $\pm$ 0.04}&\textbf{0.06 $\pm$ 0.04}\\
        Position Error (m) &1.07 $\pm$ 0.74&0.65 $\pm$ 0.47&\textbf{0.59 $\pm$ 0.41}\\
        \hline
        \label{table:sail_stats}
    \end{tabular}
\end{table}

\subsection{Straight Trackline Missions for Evaluation of Forward Reachability Analysis}

In this section we report the results from two experiments to validate the real-time calculation of forward reachable sets with bias estimation. We added unmodeled disturbances during the course of these experiments to highlight the change in forward reachable set. These significant disturbances were triggered approximately midway through the mission and after the \ac{USV} had approximately achieved steady tracking of the straight reference trajectory.   The configuration of the experiments is described in Section \ref{sec:straight_trak_miss}. 

\vspace{-20pt}

\subsubsection{Straight Trackline Mission with Simulated Fault}

\begin{figure}[H]
    \centering
    \includegraphics[width=1.0\textwidth]{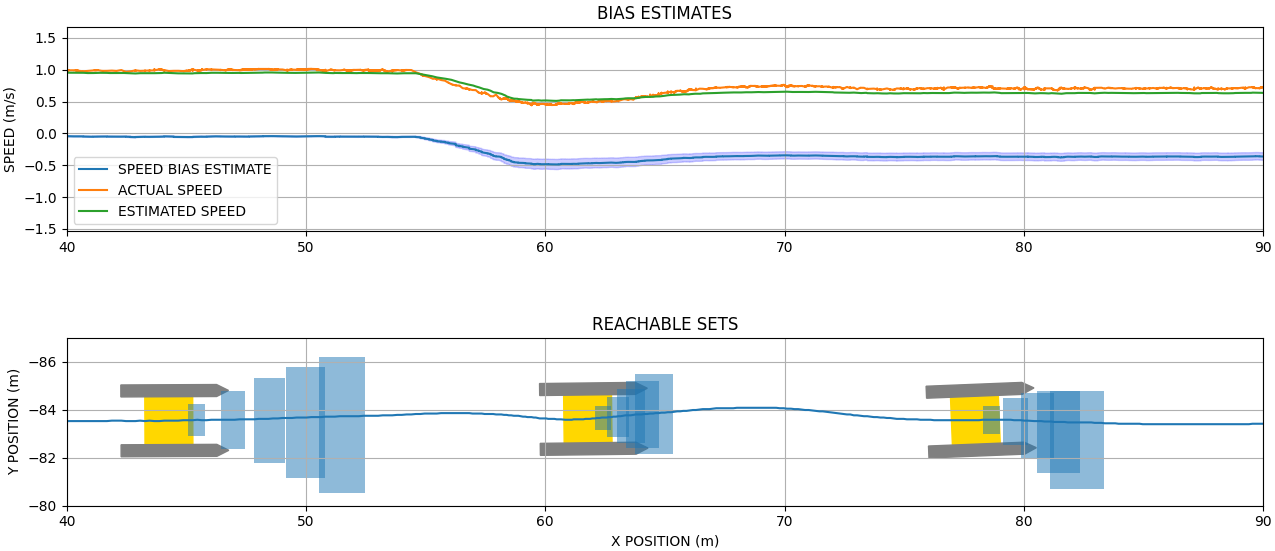}
    \caption{\ac{USV} real-time forward reachability computation during the straight trackline mission with simulated fault on both thrusters where the input was reduced to 30\% of the commanded value.  The simulated disturbance occurs when the vehicle is approximately 55 meters along the x axis. }
    \label{fig:thrust_reach}
\end{figure}

In the first reachability experiment, shown in \cref{fig:thrust_reach}, the \ac{USV} tracked along a straight line at $1.0 \sfrac{m}{s}$ until we induced a simulated thruster fault with $\digamma_{\!\!tf_L} = \digamma_{\!\!tf_L} = 0.3$ in (\ref{eq:sim_fault_cmd_L}).

\subsubsection{Straight Trackline Mission with Deployed Drogue}

\begin{figure}[H]
    \centering
    \includegraphics[width=1.0\textwidth]{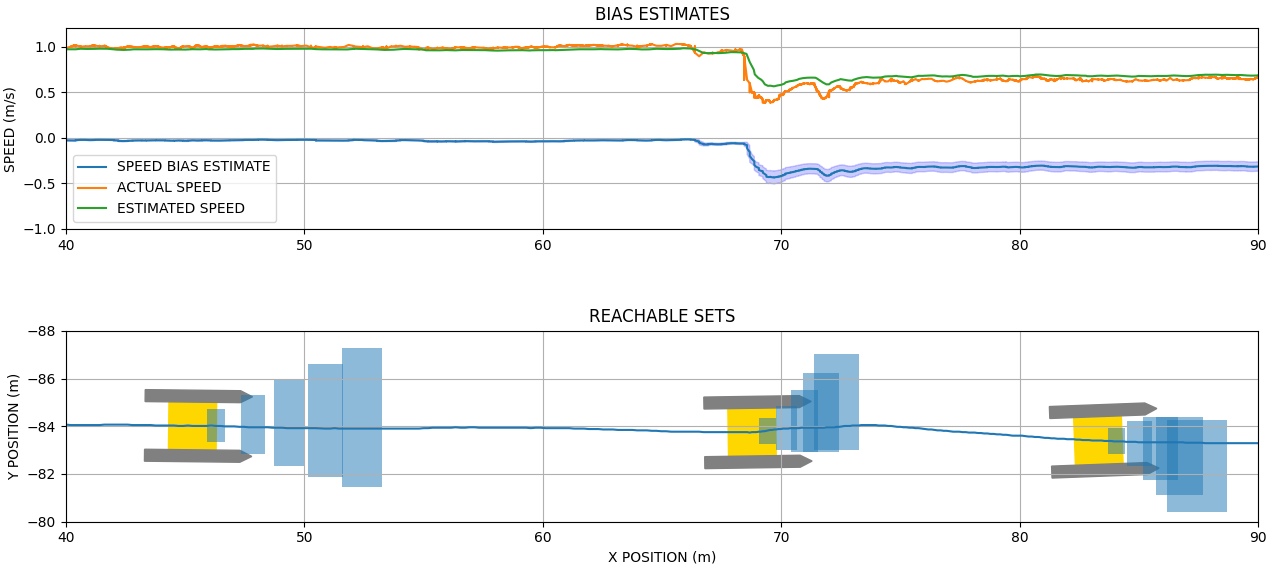}
    \caption{\ac{USV} real-time forward reachability computation during the straight trackline mission with deployed drogue. The simulated disturbance occurs when the vehicle is approximately 66 meters along the x axis.}
    \label{fig:drogue_reach}
\end{figure}

In the reachability experiment shown in \cref{fig:drogue_reach}, the \ac{USV} tracked along a straight line at $1.0 \sfrac{m}{s}$ until we deployed a drogue to induce drag and reduce the vehicle speed.

\subsection{Discussion}

We validated the performance of the controllers and forward reachability calculations on the \ac{USV} in systematic and isolated tests. With the exception of heading error, the \ac{LQR} \ac{PI} and \ac{MRAC} controllers performed better than the base-line PID.  The most notable improvement was with regard to position error in all scenarios, and the percentage reduction in \ac{RMSE} is listed in Table \ref{table:error_reduction}. 
\begin{table}[H] 
    \centering
    \caption{Percentage improvement in \ac{RMSE} for position}
    \label{table:error_reduction}
    \begin{tabular}[H]{lcc}
        \hline
         & Error reduction & Error reduction \\
        Scenario & using LQR-PI & using MRAC \\
        \hline
        Baseline                 & 50\%           & \textbf{52\%} \\
        Simulated Thruster Fault & \textbf{82\%}  & 81\%             \\
        Drogue Fault             & 70\%           & \textbf{77\%} \\
        Sail Fault               & 39\%           & \textbf{45\%} \\
        \hline
    \end{tabular}
\end{table}
We found more mixed results when comparing between the \ac{LQR} \ac{PI} and \ac{MRAC} controllers as the differences in performance where much smaller. 
As shown in \cref{table:error_reduction}, the \ac{MRAC} controller, which is an the adaptive augmentation of the \ac{LQR}-\ac{PI} controller, had slightly better performance on the basis of position error in three of the four scenarios, and similar performance in the simulated thruster fault scenario.    These results are evidence that including adaptive augmentation, such as in our \ac{MRAC} formulation, reduces error in position even when the underlying controller includes an integration term.  In many situations, the position error is of the utmost importance and in the following section we use two real-world scenarios that further illustrate this point.  

\section{Experimental Results - Safety-Critical Scenarios}\label{sec:experimental_results_scenarios}

This section describes the experimental results in two scenarios were precise control of a marine vehicle and safety assurances are necessary: an \ac{UNREP} scenario with two vehicles, and the transit through a narrow canal. There were two tests in each scenario: one where \ac{PID} control is used, and the other where \ac{MRAC} is used.  

As described previously, these types of scenarios are barriers to fielding fully autonomous \acp{USV} because they are difficult to complete and in many cases there are no good alternatives.   Even though they are hazardous, canals save significant time and energy.   Tight passages similar to canals are also found in many harbors and inter-coastal waterways.  Moreover, all vessels must depart from a pier or other structure which are usually located within a harbor, and while some crewed vessels use tugboats to carefully maneuver through a harbor, they often complete a portion of the passage out to open water under their own power.  Resupplying in the open water is also challenging, but it can significantly extend the operational use of vessels since they do not need to return to a harbor to do it.  The process of maneuvering close enough to another vessel to transfer supplies is complicated by the fact that most vessels are underactuated and lack a dynamic positioning system (DPS).  As a result, the standard \ac{UNREP} scenario in the open water involves both vehicles moving forward at a steady speed and then gradually closing the distance between them to connect cables.  The alternative of physically joining the vehicles can lead to significant damage.

\subsection{UNREP Scenario}\label{sec:UNREP_results}
\begin{figure}[ht]
    \centering
    \includegraphics[width=0.8\textwidth]{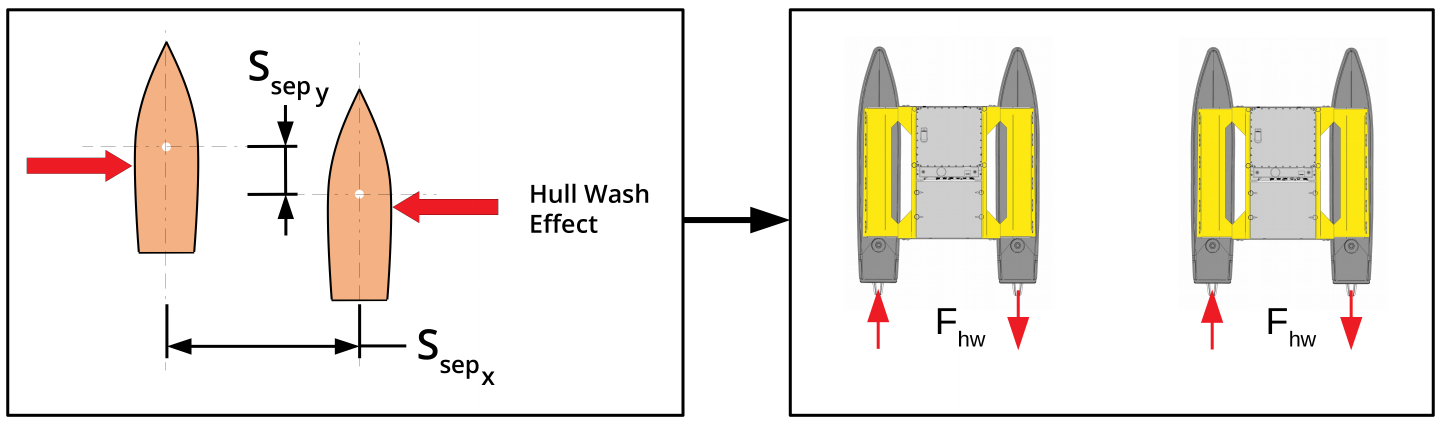}
    \caption{The simulated hull effect forces on each vehicle in the \ac{UNREP} scenario where implemented as a rotational bias from the two thrusters per (\ref{eq:hull_effect_thr_bias}).}
    \label{fig:hull_wash_effect_forces}
\end{figure}

In the \ac{UNREP} scenario, two Heron \acp{USV} are tasked to safely perform a maneuver where the two vehicles cooperatively move in close proximity to each other. One Heron \ac{USV} is a guide ship, and the other Heron \ac{USV} is the approach ship. The guide ship is tasked with a straight line path following task, and the approach ship must reach a position that is a user selected distance to the port or starboard side of the guide ship. In these experiments the approaching Heron \ac{USV} follows a behavior that uses the guidance law from \cite{Skejic.Brevik.ea2009} as reported in Section \ref{sec:fossen_unrep_guidance}. It is assumed that the vehicles can communicate state information freely between each other.  
\begin{figure}[ht]
    \centering    \includegraphics[width=0.7\textwidth]{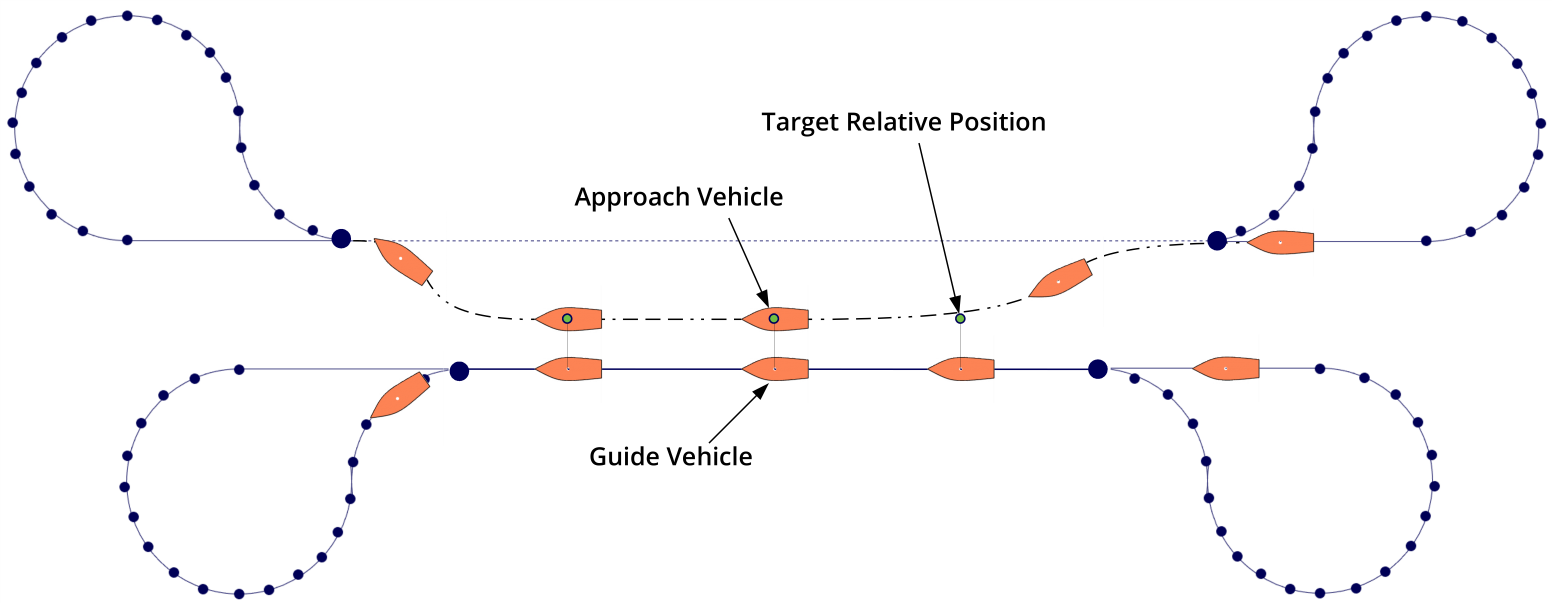}
    \caption{The \ac{UNREP} test scenario is built upon two LegRun behaviors.  When the guide vehicle enters the straight leg the \ac{UNREP} process begins, and the UNREP approach behavior (Section \ref{sec:fossen_unrep_guidance}) becomes active on the approaching vehicle.  The approaching vehicle tracks a target position relative to the guide vehicle until the end of the straight leg.  At the end of the leg both vehicles turn around and the process repeats in the opposite direction.}
    \label{fig:unrep_overview}
\end{figure}
A notable challenge when performing \ac{UNREP} operations with full-scale models is the presence of a hull wash effect that can cause a net force in the sway degree-of-freedom that pushes the two vessels together as shown in Figure \ref{fig:hull_wash_effect_forces}.  This effect is difficult to model, and typically occurs with slender hydrodynamic bodies near each other \cite{Reeve_Suction_1911,Zhou2023ShipShip}. The body of the Heron \ac{USV} is not large enough to produce a realistic hull wash effect; hence, a proximity-based thruster bias, which is not measured by the controller, was implemented to simulate hull wash effect for these experiments.  Since the Heron \ac{USV} is not directly actuated in the surge degree-of-freedom, the net effect was injected as a rotational bias on the two thrusters, i.e. in (\ref{eq:sim_fault_cmd_L}) and (\ref{eq:sim_fault_cmd_R}), $\digamma_{\!\!{tb}_L} = F_{hw}$ and $\digamma_{\!\!{tb}_R} = -F_{hw}$. The simulated hull wash disturbance $F_{hw}$ for each USV was a function of the x and y separation distances in meters, $s_{sep_x}$ and $s_{sep_y}$, with respect to the other USV's local coordinate frame. The hull wash disturbance was defined as
\begin{equation}\label{eq:hull_effect_thr_bias}
    F_{hw}=
        \begin{cases}
    		0, & \text{if $s_{sep_x} > 15$, $s_{sep_y} > 15$}\\
            \delta F_{hw_{max}}(1 - \dfrac{s_{sep_x}}{15})(1 - \dfrac{s_{sep_y}}{15}), & \text{if $s_{sep_x} <= 15$, $s_{sep_y} <= 15$}
        \end{cases}
\end{equation}

The maximum simulated hull wash disturbance $F_{hw_{max}}$ was set to 20 for the UNREP missions. Additionally, the hull wash disturbance was applied when $s_{sep_x}$ or $s_{sep_y}$ were within 15 meters. The $\delta$ term defines the direction of the rotational bias and is either 1 or -1 according to the heading of the other vehicle. This value is found by taking the sign of the outer product between the heading of the two vessels.

To add further uncertainty to the experiments, the drogue was deployed from the approach vessel shortly after the vehicle achieved relative position.  Furthermore, both the guide and approach vessels faced a sudden 50$\%$ reduction in left thruster control effectiveness, $\digamma_{\!\!tf_L} = 0.5$ in (\ref{eq:sim_fault_cmd_L}). 

\begin{figure*}[ht]
    \centering
    \begin{subfigure}[t]{0.47\textwidth}
        \centering
        \includegraphics[width=1.0\textwidth]{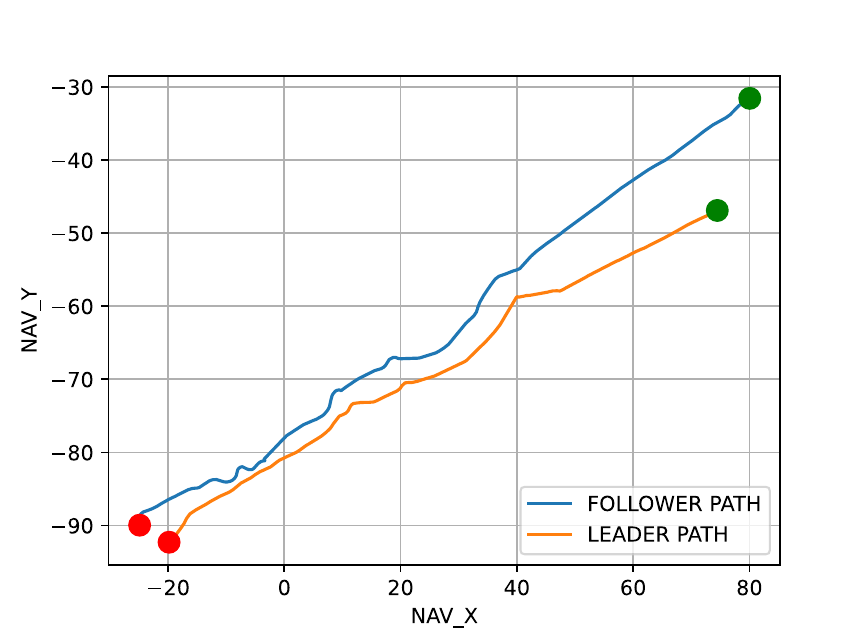}
        \caption{Trajectories with both vehicles under PID control.}
        \label{fig:pid_unrep_traj}
    \end{subfigure}%
    ~  \hspace{2mm}
    \begin{subfigure}[t]{0.47\textwidth}
        \centering    
        \includegraphics[width=1.0\textwidth]{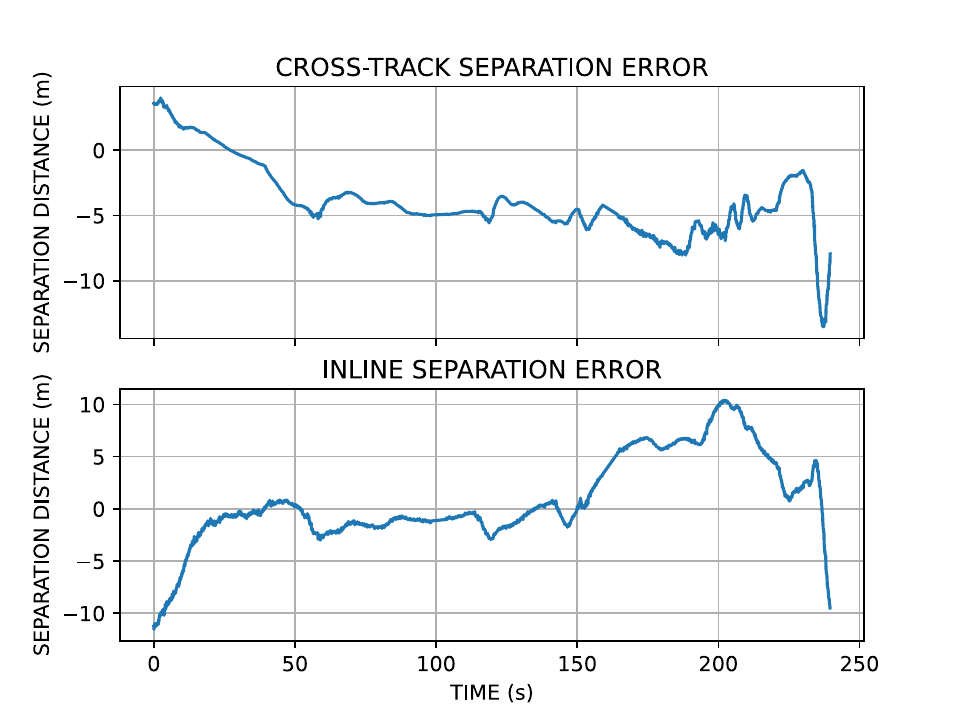}
        \caption{Cross-track and inline separation errors with PID controller}
        \label{fig:pid_urep_errors}
    \end{subfigure}%
    \caption{Experimental results for UNREP Scenario with \textbf{PID} controller}
    \label{fig:pid_urep_results}
    \vspace{-2mm}   
\end{figure*}
\begin{figure*}[ht]
    \centering
    \begin{subfigure}[t]{0.47\textwidth}
        \centering
        \includegraphics[width=1.0\textwidth]{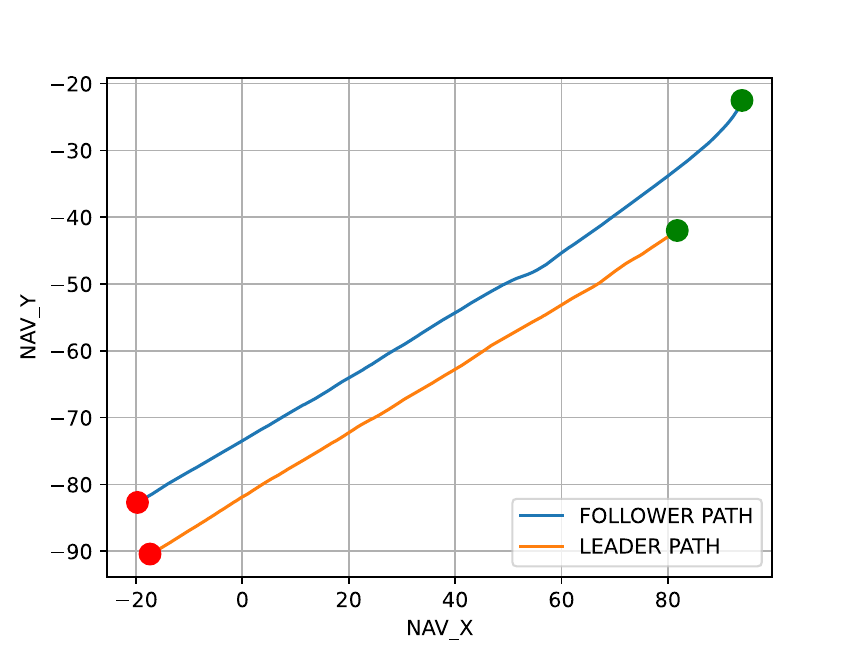}
        \caption{Trajectories with both vehicles under MRAC control.}
        \label{fig:mrac_unrep_traj}
    \end{subfigure}%
    ~  \hspace{2mm}
    \begin{subfigure}[t]{0.47\textwidth}
        \centering
        \includegraphics[width=1.0\textwidth]{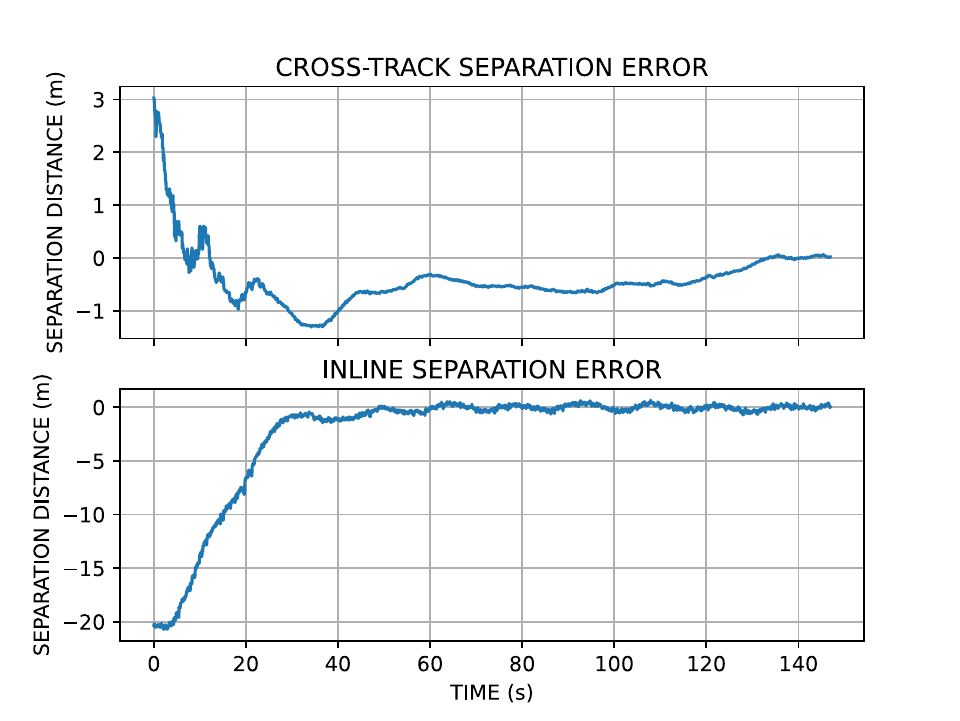}
        \caption{Cross-track and inline separation errors with MRAC controller}
        \label{fig:mrac_urep_errors}
    \end{subfigure}%
    \caption{Experimental results for UNREP Scenario with \textbf{MRAC} controller}
    \label{fig:mrac_urep_results}
    \vspace{-3mm}   
\end{figure*}
The results from two experimental trials are shown in Figure \ref{fig:pid_urep_results} and Figure \ref{fig:mrac_urep_results}.
Despite the simulated hull wash effect, thruster reduction in effectiveness, and drogue device, both vessels completed the \ac{UNREP} mission while using the developed \ac{MRAC} controllers. However, the \ac{PID} controller could not compensate for the uncertainty due to simulated hull wash effect alone; the \ac{PID} controller failed and triggered an emergency vessel collision avoidance behavior before the thurster and drogue effects could be deployed. As expected, the \ac{MRAC} controllers performed better than the \ac{PID} controllers did in the \ac{UNREP} scenario; \ac{MRAC} has better error tracking and ability to reconstitute control despite the additional effects from this experiment. 

\subsection{Canal Scenario}
\begin{figure}[ht]
    \centering
    \includegraphics[width=0.7\textwidth]{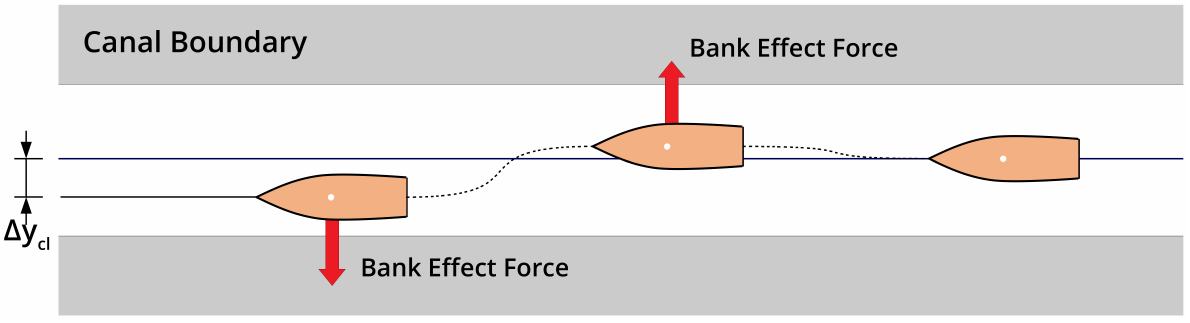}
    \caption{The canal transit scenario with simulated bank effect when the vehicle body approaches the edge of the canal.}
    \label{fig:canal_demo_bank_effect}
\end{figure}

The canal scenario was inspired by the 2021 Suez Canal obstruction incident where a ship ran aground due in part to additional wind loading and bank effect \cite{Fan2022SuezCanalObsAnalysis}. In this experiment, one Heron \ac{USV} was tasked to perform straight line-following through a virtual canal 8 meters wide where a bank effect was simulated when the vehicle deviates from the canal centerline. The virtual canal has a deadzone in the center where there is no simulated bank effect; however, outside of the deadzone but inside of the canal, the Heron \ac{USV} was subject to a thruster bias that pointed the Heron \ac{USV} away from the centerline and toward the ``shore'' of the virtual canal.  The details of this effect are shown in Figure \ref{fig:canal_demo_bank_effect}.  
The implementation of this bank effect disturbance was identical in implementation of the hull effect disturbance from Section \ref{sec:UNREP_results}.  Specifically, the value of the of simulated bank effect disturbance $F_{be}$ as a function of the deviation from centerline $\Delta y_{cl}$ was defined as
\begin{equation}
    F_{be}=
        \begin{cases}
    		0, & \text{if $\Delta y_{cl} < w_d/2$}\\
            \delta (F_{be_{max}} - F_{be_{min}})(1 - \dfrac{w_c/2 - \Delta y_{cl}}{(w_c - w_d)/2}) + F_{be_{min}}, & \text{if $w_d/2 <= \Delta y_{cl} < w_c/2$}\\
    		-99, & \text{if $\Delta y_{cl} >= w_c/2$}.\\
        \end{cases}
\end{equation}
The width of the canal $w_d$ for the mission was 8 meters. If the USV ever exits the canal boundary then a -99 bias is applied to both thrusters, effectively halting the USV. Note that the sign changes to $\digamma_{\!\!{tb}_L}$ and $\digamma_{\!\!{tb}_R}$ discussed for the hull wash disturbance are not applied when the ship exceeds the canal boundary. Additionally, the deadzone width for the canal $w_c$ is 1 meter. The deadzone of the canal is the area near the centerline of the canal where no simulated disturbance is induced. The disturbance increases linearly from $F_{be_{min}}$ to $F_{be_{max}}$ when $\Delta y_{cl}$ is between $w_d/2$ and $w_c/2$. $F_{be_{min}}$ and $F_{be_{max}}$ are 3 and 20 respectively for the canal mission. Similar to the UNREP case, the $\delta$ term is either 1 or -1 and defines the direction of the rotational bias by taking the sign of the outer product between the heading of the \ac{USV} and the canal boundaries.

Furthermore, the vessel was subject to a sudden 50$\%$ reduction in control effectiveness of the left thruster and the deployable sail to increase environmental wind loading. To follow the trackline the Heron \ac{USV} used the $L_1$ guidance behavior described in Section \ref{sec:L1_guidance}.

\begin{figure}[ht]
    \centering
    \includegraphics[width=1.0\textwidth]{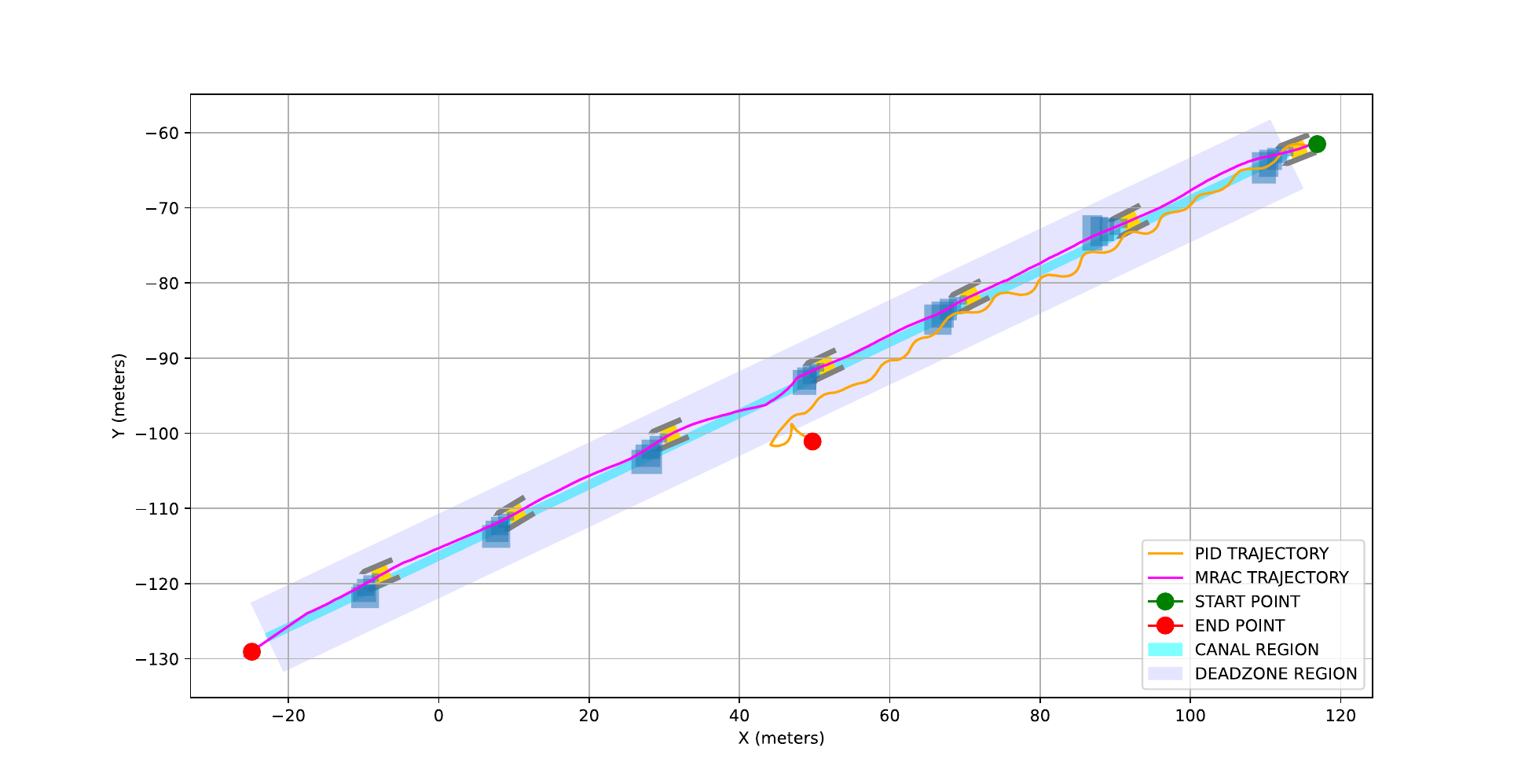}
    \caption{Trajectories of the Heron \ac{USV} in the simulated canal with both \ac{PID} and \ac{MRAC} controllers. Approximately half-way along the canal the sail was deployed (at $x \approx 70m$) and the simulated thruster failure was commanded.  The MRAC controller recovered and completed the mission, while the PID control did not.  The forward reachable sets for the closed-loop system with the \ac{MRAC} controller are shown in blue}
    \label{fig:canal_combined_traj}
\end{figure}
The experimental results from the canal scenario missions are shown in Figure \ref{fig:canal_combined_traj}.  Despite the presence of the additional disturbances, the Heron \ac{USV} with \ac{MRAC} controller successfully navigated through the virtual canal, while it could not do so with the \ac{PID} controller.  The results from the two tests are shown in Figure \ref{fig:canal_combined_traj}.
While the \ac{PID} controller attempted to recover, it was not able to correctly compensate for the additional uncertainty, causing the vehicle to ``run aground'' on the shore of the virtual canal. The MRAC controller quickly reconstituted control despite the effects of the three induced disturbances (simulated hull wash, thruster failure, and deployed sail) and naturally occurring disturbances (e.g., waves and river currents).   Furthermore, the real-time reachable sets (shown in blue) reflected the change in the system dynamics that occurs at $x \approx 70m$. Just after the sail is deployed the size of the reachable sets decreases and they do not extend as far ahead of the \ac{USV}.  As the \ac{USV} traveled further down the virtual canal we see the integration and adaptation in the \ac{MRAC} controller take effect as the size of the reachable sets increase and their location extend further ahead of the \ac{USV}, although neither fully recover to there original state.

\section{Conclusions}
This paper presented algorithms to provide control and safety verification to a \ac{USV} subject to strong environmental disturbances and actuator malfunctions.
We outlined an adaptive control formulation designed to handle varying disturbances and conducted extensive experiments showing its ability to maintain control despite induced drag forces and thruster malfunctions.
Moreover, to provide safety certification in the face of such disturbances we demonstrated our approach to computing forward reachable sets in real-time using a \ac{MHE} with a \ac{CG} representation of the closed loop system. 

Our experimental tests were conducted in two sets.  In the first set of experiments we investigated the performance of the \ac{MRAC} controller vs \ac{PID} and \ac{LQR}-\ac{PI} in a series of repeated missions. Both the \ac{LQR}-\ac{PI} and \ac{MRAC} controllers had better performance than the baseline \ac{PID}, and with one exception, the \ac{MRAC} controller had the lowest position error.
We found more mixed results when comparing between the \ac{LQR} \ac{PI} and \ac{MRAC} controllers as the differences in performance where much smaller. As shown in Table \ref{table:error_reduction}, the \ac{MRAC} controller, which is an the adaptive augmentation of the \ac{LQR}-\ac{PI} controller, had slightly better performance on the basis of position error in three of the four scenarios, and similar performance in the simulated thruster fault scenario.   These results are evidence that including adaptive augmentation, such as that in our \ac{MRAC} formulation, reduces error in position even when the underlying controller includes an integration term.  Finally, we also validated the performance of the real-time reachability calculation and showed it could capture the effect of significant disturbances when computing the forward reachable sets. 

In the second set of experiments, we demonstrated the performance of both adaptive control and forward reachability in two safety critical scenarios: \ac{UNREP} and a canal transit.  Both scenarios included significant disturbances in an effort to represent the worst case that could happen, and bolster confidence that these maneuvers can be completed autonomously, or with human supervision in the future. In both cases, we found that a Heron \ac{USV} using our \ac{MRAC} controller was able to complete the mission while the established baseline \ac{PID} controller failed. Overall, these results on both control performance and safety assurances via real-time reachability are encouraging. Both the systematic evaluation and scenario demonstration suggest that adaptive control and real-time reachability could help facilitate wider user of \acp{USV} in the field because of improved performance in the face of unanticipated disturbances and the ability to provide real-time assessment of safety.

\section{Acknowledgements}
This work was supported in part by the US Navy NIWC Atlantic Award N6523623C8011.
This research was developed with funding from the Defense Advanced Research Projects Agency (DARPA). The views, opinions and/or findings expressed are those of the author(s) and should not be interpreted as representing the official views or policies of the Department of Defense or the U.S. Government.

\bibliographystyle{apalike}
\bibliography{refs}

\begin{thebibliography}{}

\bibitem[Allg{\"o}wer et~al., 1999]{Allgower.Badgwell.ea1999}
Allg{\"o}wer, F., Badgwell, T.~A., Qin, J.~S., Rawlings, J.~B., and Wright, S.~J. (1999).
\newblock Nonlinear predictive control and moving horizon estimation—an introductory overview.
\newblock {\em Adv. Control}, pages 391--449.

\bibitem[Althoff and Dolan, 2014]{althoff2014online}
Althoff, M. and Dolan, J.~M. (2014).
\newblock Online verification of automated road vehicles using reachability analysis.
\newblock {\em IEEE Trans. Rob.}, 30(4):903--918.

\bibitem[Antonelli et~al., 2001]{Antonelli2001AdaptiveODIN}
Antonelli, G., Chiaverini, S., Sarkar, N., and West, M. (2001).
\newblock Adaptive control of an autonomous underwater vehicle: experimental results on {ODIN}.
\newblock {\em EEE Trans. Control Syst. Technol.}, 9(5):756--765.

\bibitem[{\AA}str{\"o}m and H{\"a}gglund, 2006]{Astrom2006}
{\AA}str{\"o}m, K.~J. and H{\"a}gglund, T. (2006).
\newblock {\em Advanced PID control}.
\newblock ISA-The Instrumentation, Systems and Automation Society.

\bibitem[Bansal et~al., 2017]{bansal2017hamilton}
Bansal, S., Chen, M., Herbert, S., and Tomlin, C.~J. (2017).
\newblock Hamilton-jacobi reachability: A brief overview and recent advances.
\newblock In {\em Proc. IEEE Conf. Decis. Control}, pages 2242--2253. IEEE.

\bibitem[Benjamin, 2024]{IvPManPages}
Benjamin, M.~R. (2024).
\newblock An overview of {MOOS-IvP} and a users guide to the {IvP} helm.
\newblock \url{https://oceanai.mit.edu/ivpman/pmwiki/pmwiki.php}.
\newblock Release 22.8.

\bibitem[Benjamin et~al., 2010]{Benjamin2010MOOSIvP}
Benjamin, M.~R., Schmidt, H., Newman, P.~M., and Leonard, J.~J. (2010).
\newblock Nested autonomy for unmanned marine vehicles with {MOOS-IvP}.
\newblock {\em J. Field Robotics}, 27(6):834--875.

\bibitem[Bolender et~al., 2015]{Bolender2015HIFire}
Bolender, M.~A., Dauby, B., Muse, J.~A., and Adamczak, D. (2015).
\newblock Hifire 6: Overview and status update 2015.
\newblock In {\em 20th AIAA International Space Planes and Hypersonic Systems and Technologies Conference}.

\bibitem[Borrelli et~al., 2017]{borrelli_bemporad_morari_2017}
Borrelli, F., Bemporad, A., and Morari, M. (2017).
\newblock {\em Predictive Control for Linear and Hybrid Systems}.
\newblock Cambridge University Press.

\bibitem[Deepmind, 2020]{jaxverify}
Deepmind (2020).
\newblock jax\_verify: Neural network verification in jax.

\bibitem[Dutta et~al., 2019]{dutta2019reachability}
Dutta, S., Chen, X., and Sankaranarayanan, S. (2019).
\newblock Reachability analysis for neural feedback systems using regressive polynomial rule inference.
\newblock In {\em Proc. Int. Conf. Hybrid Syst.: Comp. Control}, pages 157--168.

\bibitem[Dydek et~al., 2013]{Dydek2013AC_UAVs}
Dydek, Z.~T., Annaswamy, A.~M., and Lavretsky, E. (2013).
\newblock Adaptive control of quadrotor {UAV}s: A design trade study with flight evaluations.
\newblock {\em EEE Trans. Control Syst. Technol.}, 21(4):1400--1406.

\bibitem[Evans, 1998]{evans1998partial}
Evans, L.~C. (1998).
\newblock {\em Graduate studies in mathematics}, volume~19.
\newblock American Mathematical Society Providence, RI.

\bibitem[Everett et~al., 2021a]{everett2021efficient}
Everett, M., Habibi, G., and How, J.~P. (2021a).
\newblock Efficient reachability analysis of closed-loop systems with neural network controllers.
\newblock In {\em Proc. IEEE Int. Conf. Robot. Autom.}, pages 4384--4390. IEEE.

\bibitem[Everett et~al., 2021b]{everett2021reachability}
Everett, M., Habibi, G., Sun, C., and How, J.~P. (2021b).
\newblock Reachability analysis of neural feedback loops.
\newblock {\em IEEE Access}, 9:163938--163953.

\bibitem[Fan et~al., 2020]{fan2020reachnn}
Fan, J., Huang, C., Chen, X., Li, W., and Zhu, Q. (2020).
\newblock Reach{NN}*: A tool for reachability analysis of neural-network controlled systems.
\newblock In {\em Proc. Int. Symp. Autom. Tech. for Verif. Anal.}, pages 537--542. Springer.

\bibitem[Fan et~al., 2022]{Fan2022SuezCanalObsAnalysis}
Fan, S., Yang, Z., Wang, J., and Marsland, J. (2022).
\newblock Shipping accident analysis in restricted waters: {L}esson from the {S}uez {C}anal blockage in 2021.
\newblock {\em Ocean Eng.}, 266:113119.

\bibitem[Fischer et~al., 2014]{Fischer2014RISE}
Fischer, N., Hughes, D., Walters, P., Schwartz, E.~M., and Dixon, W.~E. (2014).
\newblock Nonlinear {RISE}-based control of an autonomous underwater vehicle.
\newblock {\em IEEE Trans. Rob.}, 30(4):845--852.

\bibitem[Fossen, 1994]{Fossen1994guidance}
Fossen, T. (1994).
\newblock {\em Guidance and Control of Ocean Vehicles}.
\newblock Wiley.

\bibitem[Fossen and Fjellstad, 1993]{Fossen1993ActuatorDynamics}
Fossen, T. and Fjellstad, O.-E. (1993).
\newblock Cascaded adaptive control of marine vehicles with significant actuator dynamics.
\newblock In {\em Proc. Int. Fed. Autom. Control}, volume 26, issue 2, part 4,, pages 1123--1128.

\bibitem[Fossen and Fjellstad, 1995]{Fossen1995AC_Comp}
Fossen, T.~I. and Fjellstad, O.-E. (1995).
\newblock Robust adaptive control of underwater vehicles: A comparative study.
\newblock {\em Int. Fed. Autom. Control Proc. Vol.}, 28, issue 2:66--74.
\newblock 3rd IFAC Workshop on Control Applications in Marine Systems, Trondheim, Norway, 10-12 May.

\bibitem[Frittelli et~al., 2024]{CRS2024BaltimoreBridge}
Frittelli, J., Goldman, B., and Lohman, A.~E. (2024).
\newblock Baltimore bridge collapse: Frequently asked questions.
\newblock Technical Report CRS Report No. R48028, Congressional Research Service, Washington DC, USA.

\bibitem[Fukao et~al., 2000]{Fukao2000ATTC_Mobile}
Fukao, T., Nakagawa, H., and Adachi, N. (2000).
\newblock Adaptive tracking control of a nonholonomic mobile robot.
\newblock {\em IEEE Trans. Robot. Autom.}, 16(5):609--615.

\bibitem[Gleason et~al., 2017]{gleason2017underapproximation}
Gleason, J.~D., Vinod, A.~P., and Oishi, M.~M. (2017).
\newblock Underapproximation of reach-avoid sets for discrete-time stochastic systems via {L}agrangian methods.
\newblock In {\em Proc. IEEE Conf. Decis. Control}, pages 4283--4290. IEEE.

\bibitem[Haldeman et~al., 2016]{haldeman2016lessening}
Haldeman, C.~D., Aragon, D.~K., Miles, T., Glenn, S.~M., and Ramos, A.~G. (2016).
\newblock Lessening biofouling on long-duration {AUV} flights: Behavior modifications and lessons learned.
\newblock In {\em Proc. IEEE OCEANS}, pages 1--8. IEEE.

\bibitem[Harris et~al., 2023]{Harris2023AID}
Harris, Z.~J., Mao, A.~M., Paine, T.~M., and Whitcomb, L.~L. (2023).
\newblock Stable nullspace adaptive parameter identification of 6 degree-of-freedom plant and actuator models for underactuated vehicles: Theory and experimental evaluation.
\newblock {\em Int. J. Robot. Res.}, 42(12):1070--1093.

\bibitem[Herbert et~al., 2017]{herbert2017fastrack}
Herbert, S.~L., Chen, M., Han, S., Bansal, S., Fisac, J.~F., and Tomlin, C.~J. (2017).
\newblock Fastrack: A modular framework for fast and guaranteed safe motion planning.
\newblock In {\em Proc. IEEE Conf. Decis. Control}, pages 1517--1522. IEEE.

\bibitem[Hu et~al., 2020]{hu2020reach}
Hu, H., Fazlyab, M., Morari, M., and Pappas, G.~J. (2020).
\newblock Reach-{SDP}: Reachability analysis of closed-loop systems with neural network controllers via semidefinite programming.
\newblock In {\em Proc. IEEE Conf. Decis. Control}, pages 5929--5934. IEEE.

\bibitem[Huang et~al., 2019]{huang2019reachnn}
Huang, C., Fan, J., Li, W., Chen, X., and Zhu, Q. (2019).
\newblock Reach{NN}: Reachability analysis of neural-network controlled systems.
\newblock {\em ACM Trans. Embed. Comp. Sys.}, 18(5s):1--22.

\bibitem[Huang et~al., 2016]{huang2016autonomous}
Huang, W., Wang, K., Lv, Y., and Zhu, F. (2016).
\newblock Autonomous vehicles testing methods review.
\newblock In {\em Proc. IEEE Int. Conf. Intell. Trans. Syst.}, pages 163--168. IEEE.

\bibitem[Ivanov et~al., 2019]{ivanov2019verisig}
Ivanov, R., Weimer, J., Alur, R., Pappas, G.~J., and Lee, I. (2019).
\newblock Verisig: verifying safety properties of hybrid systems with neural network controllers.
\newblock In {\em Proc. Int. Conf. Hybrid Syst.: Comp. Control}, pages 169--178.

\bibitem[Katz et~al., 2019]{katz2019marabou}
Katz, G., Huang, D.~A., Ibeling, D., Julian, K., Lazarus, C., Lim, R., Shah, P., Thakoor, S., Wu, H., Zelji{\'c}, A., et~al. (2019).
\newblock The {M}arabou framework for verification and analysis of deep neural networks.
\newblock In {\em Proc. Int. Conf. Comput. Aid. Verif.}, pages 443--452. Springer.

\bibitem[Koopman and Wagner, 2016]{koopman2016challenges}
Koopman, P. and Wagner, M. (2016).
\newblock Challenges in autonomous vehicle testing and validation.
\newblock {\em SAE Int. J. Transp. Saf.}, 4(1):15--24.

\bibitem[Lavretsky and Wise, 2012]{Lavretsky.Wise2012}
Lavretsky, E. and Wise, K.~A. (2012).
\newblock {\em Robust and adaptive control: With aerospace applications}.
\newblock Springer.

\bibitem[Lee and Wong, 2021]{Lee2021SuezInpact}
Lee, J. and Wong, E. (2021).
\newblock Suez canal blockage: an analysis of legal impact, risks and liabilities to the global supply chain.
\newblock {\em MATEC Web. Conf.}, 339:01019.

\bibitem[Lew and Pavone, 2021]{lew2021sampling}
Lew, T. and Pavone, M. (2021).
\newblock Sampling-based reachability analysis: A random set theory approach with adversarial sampling.
\newblock In {\em Proc. Conf. Robot Learn.}, pages 2055--2070. PMLR.

\bibitem[Liu et~al., 2021]{Liu2021AC}
Liu, Z., Zhang, Y., Yuan, C., and Luo, J. (2021).
\newblock Adaptive path following control of unmanned surface vehicles considering environmental disturbances and system constraints.
\newblock {\em IEEE Trans. Syst., Man, Cybern.: Syst.}, 51(1):339--353.

\bibitem[Makavita et~al., 2020]{Makavita2020MRAC}
Makavita, C.~D., Jayasinghe, S.~G., Nguyen, H.~D., and Ranmuthugala, D. (2020).
\newblock Experimental comparison of two composite {MRAC} methods for {UUV} operations with low adaptation gains.
\newblock {\em IEEE J. Ocean. Eng.}, 45(1):227--246.

\bibitem[McFarland and Whitcomb, 2014]{McFarlandWhitcomb2014AMBC_Act_Dyn}
McFarland, C.~J. and Whitcomb, L.~L. (2014).
\newblock Experimental evaluation of adaptive model-based control for underwater vehicles in the presence of unmodeled actuator dynamics.
\newblock In {\em Proc. IEEE Int. Conf. Robot. Autom.}, pages 2893--2900.

\bibitem[Miller et~al., 1987]{miller1987development}
Miller, M.~O., Hammett, J.~W., and Murphy, T.~P. (1987).
\newblock The development of the {US} {N}avy underway replenishment fleet.
\newblock {\em Soc. Naval Archit. Mar. Eng.-Trans.}, 95.

\bibitem[Mitchell, 2007]{mitchell2007comparing}
Mitchell, I.~M. (2007).
\newblock Comparing forward and backward reachability as tools for safety analysis.
\newblock In {\em Int. Workshop on Hybrid Sys.: Comp. Control}, pages 428--443. Springer.

\bibitem[Moore and Stouch, 2014]{MooreStouchKeneralizedEkf2014}
Moore, T. and Stouch, D. (2014).
\newblock A generalized extended {K}alman filter implementation for the robot operating system.
\newblock In {\em Proc. Int. Conf. Intel. Auto. Sys.} Springer.

\bibitem[Narendra and Annaswamy, 2005]{NarendraAnnaswamy2005SAS}
Narendra, K.~S. and Annaswamy, A. (2005).
\newblock {\em Stable Adaptive Systems}.
\newblock Dover Books on Electrical Engineering. Dover Publications.

\bibitem[Narendra et~al., 1985]{Narendra1985GeneralAdaptive}
Narendra, K.~S., Annaswamy, A.~M., and Singh, R.~P. (1985).
\newblock A general approach to the stability analysis of adaptive systems.
\newblock {\em Int. J. Control}, 41(1):193--216.

\bibitem[Park et~al., 2004]{Park.Dyest.ea2004}
Park, S., Deyst, J., and How, J. (2004).
\newblock A new nonlinear guidance logic for trajectory tracking.
\newblock In {\em Proc. AIAA Guid., Nav., Control}.

\bibitem[Pedrozo, 2023]{pedrozo2023advent}
Pedrozo, R. (2023).
\newblock Advent of a new era in naval warfare: Autonomous and unmanned systems.
\newblock In {\em Autonomous Vessels in Maritime Affairs: Law and Governance Implications}, pages 63--80. Springer.

\bibitem[Polvara et~al., 2018]{polvara2018obstacle}
Polvara, R., Sharma, S., Wan, J., Manning, A., and Sutton, R. (2018).
\newblock Obstacle avoidance approaches for autonomous navigation of unmanned surface vehicles.
\newblock {\em J. Nav.}, 71(1):241--256.

\bibitem[Reeve, 1911]{Reeve_Suction_1911}
Reeve, S.~A. (1911).
\newblock The hydraulic interaction between passing vessels, called ``suction'''.
\newblock In {\em Proc. United States Naval Inst.}, volume~37, pages 1347--1376.

\bibitem[Rober et~al., 2022]{Rober2022L1}
Rober, N., Hammond, M., Cichella, V., Martin, J.~E., and Carrica, P. (2022).
\newblock {3D} path following and $\mathcal{L}_1$ adaptive control for underwater vehicles.
\newblock {\em Ocean Eng.}, 253:110971.

\bibitem[Rober et~al., 2024]{rober2023online}
Rober, N., Mahesh, K., Paine, T.~M., Greene, M.~L., Lee, S., Monteiro, S.~T., Benjamin, M.~R., and How, J.~P. (2024).
\newblock Online data-driven safety certification for systems subject to unknown disturbances.
\newblock In {\em Proc. IEEE Int. Conf. Robot. Autom.} IEEE.

\bibitem[Rynne and von Ellenrieder, 2009]{rynne2009unmanned}
Rynne, P.~F. and von Ellenrieder, K.~D. (2009).
\newblock Unmanned autonomous sailing: Current status and future role in sustained ocean observations.
\newblock {\em Marine Tech. Soc. J.}, 43(1):21--30.

\bibitem[Shen et~al., 2020]{Shen2020ATTC_RBF}
Shen, Z., Wang, Y., Yu, H., and Guo, C. (2020).
\newblock Finite-time adaptive tracking control of marine vehicles with complex unknowns and input saturation.
\newblock {\em Ocean Eng.}, 198:106980.

\bibitem[Sidrane et~al., 2022]{sidrane2022overt}
Sidrane, C., Maleki, A., Irfan, A., and Kochenderfer, M.~J. (2022).
\newblock {OVERT}: An algorithm for safety verification of neural network control policies for nonlinear systems.
\newblock {\em J. Mach. Learn. Res.}, 23(117):1--45.

\bibitem[Skejic et~al., 2009]{Skejic.Brevik.ea2009}
Skejic, R., Breivik, M., Fossen, T.~I., and Faltinsen, O.~M. (2009).
\newblock Modeling and control of underway replenishment operations in calm water.
\newblock {\em Int. Fed. Autom. Control Proc. Vol.}, 42(18):78--85.

\bibitem[Skjetne et~al., 2004]{Skjetne2004AdaptControlShip}
Skjetne, R., Smogeli, {\O}.~N., and Fossen, T.~I. (2004).
\newblock {A Nonlinear Ship Manoeuvering Model: Identification and adaptive control with experiments for a model ship}.
\newblock {\em Model., Identif., Control}, 25(1):3--27.

\bibitem[Slotine and Li, 1987]{SlotineLi1987ATTC_Arm}
Slotine, J.-J.~E. and Li, W. (1987).
\newblock On the adaptive control of robot manipulators.
\newblock {\em Int. J. Robot. Res.}, 6(3):49--59.

\bibitem[Smallwood and Whitcomb, 2004]{Smallwood2004AMBC}
Smallwood, D. and Whitcomb, L. (2004).
\newblock Model-based dynamic positioning of underwater robotic vehicles: {T}heory and experiment.
\newblock {\em IEEE J. Ocean. Eng.}, 29(1):169--186.

\bibitem[Tjeng et~al., 2017]{tjeng2017evaluating}
Tjeng, V., Xiao, K.~Y., and Tedrake, R. (2017).
\newblock Evaluating robustness of neural networks with mixed integer programming.
\newblock In {\em Proc. Int. Conf. Learn. Represent.}

\bibitem[Vincent and Schwager, 2021]{vincent2021reachable}
Vincent, J.~A. and Schwager, M. (2021).
\newblock Reachable polyhedral marching ({RPM}): {A} safety verification algorithm for robotic systems with deep neural network components.
\newblock In {\em Proc. IEEE Int. Conf. Robot. Autom.}, pages 9029--9035. IEEE.

\bibitem[Weng et~al., 2018]{weng2018towards}
Weng, L., Zhang, H., Chen, H., Song, Z., Hsieh, C.-J., Daniel, L., Boning, D., and Dhillon, I. (2018).
\newblock Towards fast computation of certified robustness for {RELU} networks.
\newblock In {\em Proc. Int. Conf. Mach. Learn.}, pages 5276--5285. PMLR.

\bibitem[Whitcomb and Yoerger, 1996]{WhitcombYoerger1996ThrusterModel}
Whitcomb, L. and Yoerger, D. (1996).
\newblock Preliminary experiments in the model-based dynamic control of marine thrusters.
\newblock In {\em Proc. IEEE Int. Conf. Robot. Autom.}, volume~3, pages 2166--2173.

\bibitem[Woo et~al., 2018]{Woo2018WAMV_LSTM}
Woo, J., Park, J., Yu, C., and Kim, N. (2018).
\newblock Dynamic model identification of unmanned surface vehicles using deep learning network.
\newblock {\em Appl. Ocean Res.}, 78:123--133.

\bibitem[Xiang et~al., 2020]{xiang2020reachable}
Xiang, W., Tran, H.-D., Yang, X., and Johnson, T.~T. (2020).
\newblock Reachable set estimation for neural network control systems: A simulation-guided approach.
\newblock {\em IEEE Trans. Neural Netw. Learn. Syst.}, 32(5):1821--1830.

\bibitem[Xu et~al., 2020]{xu2020automatic}
Xu, K., Shi, Z., Zhang, H., Wang, Y., Chang, K.-W., Huang, M., Kailkhura, B., Lin, X., and Hsieh, C.-J. (2020).
\newblock Automatic perturbation analysis for scalable certified robustness and beyond.
\newblock {\em Adv. Neural Inf. Proc. Syst.}, 33:1129--1141.

\bibitem[Yoerger and Slotine, 1991]{YoergerSlotine1991}
Yoerger, D. and Slotine, J.-J. (1991).
\newblock Adaptive sliding control of an experimental underwater vehicle.
\newblock In {\em Proc. IEEE Int. Conf. Robot. Autom.}, volume~3, pages 2746--2751.

\bibitem[Yuh et~al., 2011]{yuh2011applications}
Yuh, J., Marani, G., and Blidberg, D.~R. (2011).
\newblock Applications of marine robotic vehicles.
\newblock {\em Intell. Serv. Robot.}, 4:221--231.

\bibitem[Zereik et~al., 2018]{zereik2018challenges}
Zereik, E., Bibuli, M., Mi{\v{s}}kovi{\'c}, N., Ridao, P., and Pascoal, A. (2018).
\newblock Challenges and future trends in marine robotics.
\newblock {\em Annu. Rev. Control}, 46:350--368.

\bibitem[Zhang et~al., 2008]{zhang2018efficient}
Zhang, H., Weng, T.-W., Chen, P.-Y., Hsieh, C.-J., and Daniel, L. (2008).
\newblock Efficient neural network robustness certification with general activation functions.
\newblock In {\em Adv. Neural Inf. Process. Syst.}, volume~31.

\bibitem[Zhou et~al., 2023]{Zhou2023ShipShip}
Zhou, J., Ren, J., and Bai, W. (2023).
\newblock Survey on hydrodynamic analysis of ship–ship interaction during the past decade.
\newblock {\em Ocean Eng.}, 278.

\end{thebibliography}

\newpage
\begin{appendices}
\vspace{-8mm}
\section{Block diagram of controller architecture}\label{sec:appendix_a}
\vspace{50mm}
\begin{figure}[htb]
  \centering
  \rotatebox{90}{%
    \begin{minipage}{0.95\linewidth}
        \includegraphics[height=0.5\textheight]{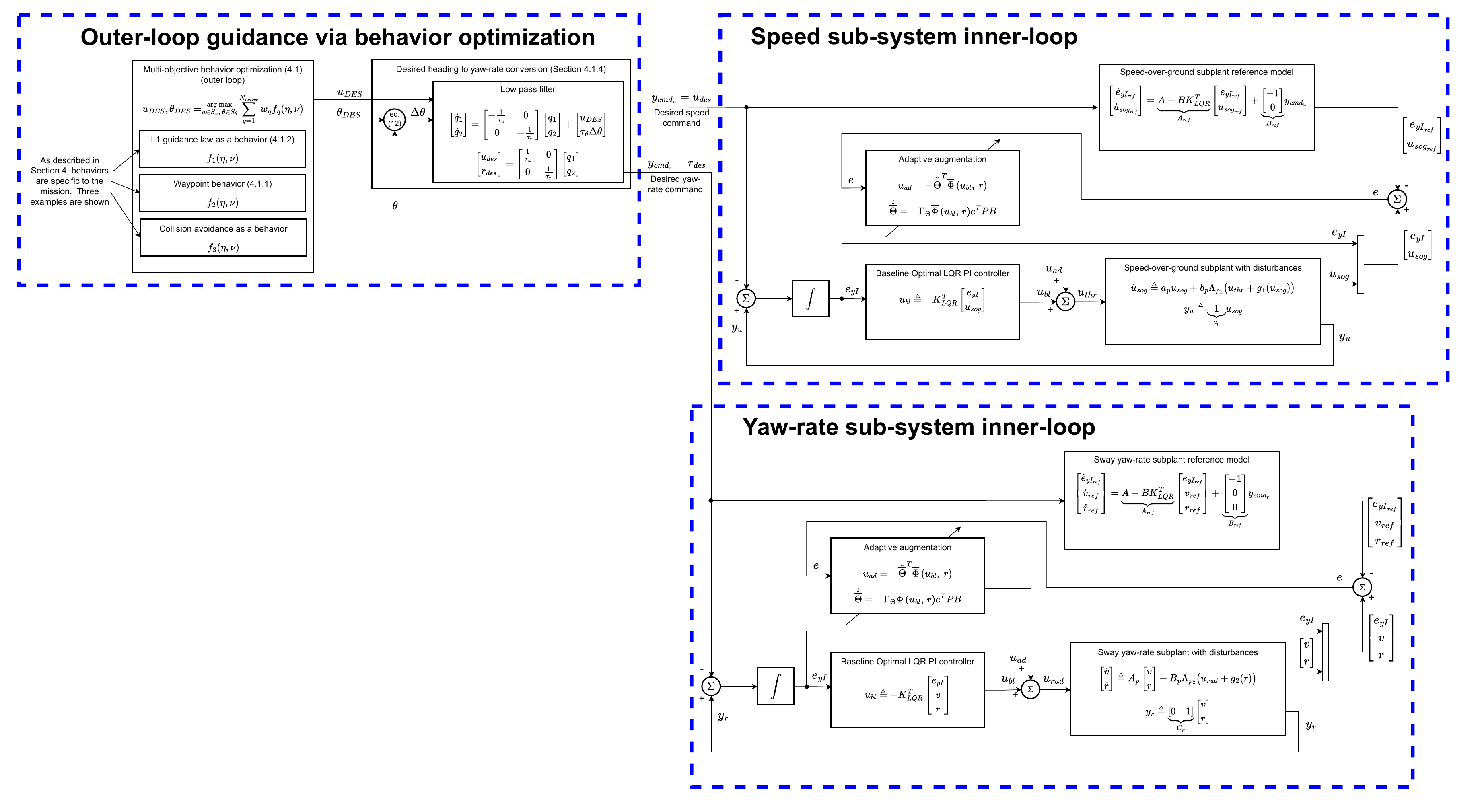}
    \caption{Block diagram of MRAC architecture. }
    \label{fig:MRAC_Block}
    \end{minipage}
  }
\end{figure}
\newpage

\section{Block diagram of forward reachability analysis method}\label{sec:appendix_b}
\vspace{60mm}
\begin{figure}[htb]
  \centering
  \rotatebox{90}{%
    \begin{minipage}{0.9\linewidth}
        \centering
        \includegraphics[height=0.45\textheight]{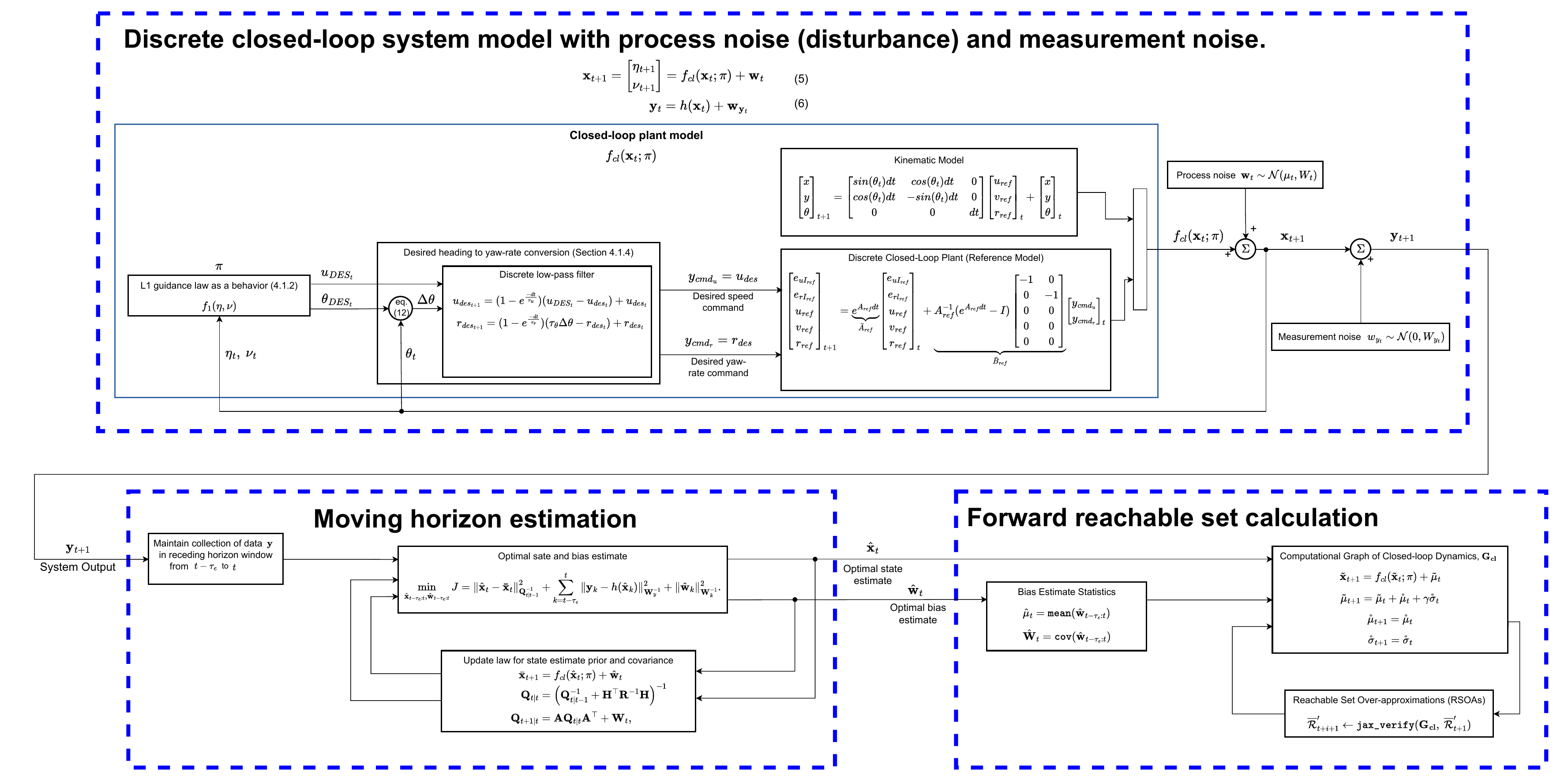}
    \caption{Block diagram of real-time forward reachability analysis with moving horizon estimation. }
    \label{fig:Reach_Block}
    \end{minipage}
  } 
\end{figure}

\end{appendices}

\end{document}